\documentclass[article]{siamonline190516}

\usepackage{lipsum}
\usepackage{amsfonts}
\usepackage{graphicx}
\usepackage{epstopdf}
\usepackage{algorithmic}
\ifpdf
  \DeclareGraphicsExtensions{.eps,.pdf,.png,.jpg}
\else
  \DeclareGraphicsExtensions{.eps}
\fi

\usepackage{wrapfig}
\usepackage{url}
\usepackage{mathtools}

\usepackage{booktabs} 
\usepackage{nicefrac}       
\usepackage{microtype}      
\usepackage{todonotes}
\usepackage{amsmath}
\usepackage{amssymb}
\usepackage{placeins}
\usepackage{array} 
\usepackage{enumitem}
\usepackage{subcaption}
\usepackage{xcolor}
\usepackage{times}

\usepackage{enumitem}
\setlist[enumerate]{leftmargin=.5in}
\setlist[itemize]{leftmargin=.5in}

\newsiamremark{remark}{Remark}
\newsiamremark{hypothesis}{Hypothesis}
\crefname{hypothesis}{Hypothesis}{Hypotheses}
\newsiamthm{claim}{Claim}

\headers{Simulation based denoising}{Mohan et. al.}

\title{Deep Denoising for Scientific Discovery: A Case Study in Electron Microscopy}

\author{Sreyas Mohan\thanks{Center for Data Science, New York University}
\and Ramon Manzorro\thanks{School for Engineering of Matter, Transport, and Energy, ASU}
\and Joshua L. Vincent\footnotemark[3]
\and Binh Tang\thanks{Department of Statistics and Data Science, Cornell University}
\and Dev Y. Sheth\thanks{Indian Institute of Technology Madras, India}
\and Eero P. Simoncelli\thanks{Howard Hughes Medical Institute, Center for Neural Science, and Courant Institute of Mathematical Sciences, NYU}
\and David S. Matteson \footnotemark[4]
\and Peter A. Crozier\footnotemark[3]
\and Carlos Fernandez-Granda \thanks{Courant Institute of Mathematical Sciences and Center for Data Science, NYU}
}

\usepackage{amsopn}

\usepackage{xr-hyper}
\makeatletter
\newcommand*{\addFileDependency}[1]{
  \typeout{(#1)}
  \@addtofilelist{#1}
  \IfFileExists{#1}{}{\typeout{No file #1.}}
}
\makeatother

\usepackage{graphicx,tikz,pgfplots}
\usepgfplotslibrary{groupplots}
\usepgfplotslibrary{statistics}
\usepackage[eulergreek]{sansmath}
\pgfplotsset{compat=1.16}

\definecolor{knred}{RGB}{175,57,55}
\definecolor{kngreen}{RGB}{106,148,81}
\definecolor{kngrey}{RGB}{125,125,125}

\ifpdf
\hypersetup{
  pdftitle={An Example Article},
  pdfauthor={D. Doe, P. T. Frank, and J. E. Smith}
}
\fi

\begin{document}

\maketitle

\begin{abstract}
Denoising is a fundamental challenge in scientific imaging. Deep convolutional neural networks (CNNs) provide the current state of the art in denoising natural images, where they produce impressive results. However, their potential has been inadequately explored in the context of scientific imaging. Denoising CNNs are typically trained on real natural images artificially corrupted with simulated noise. In contrast, in scientific applications, noiseless ground-truth images are usually not available. To address this issue, we propose a simulation-based denoising (SBD) framework, in which CNNs are trained on simulated images. We test the framework on data obtained from transmission electron microscopy (TEM), an imaging technique with widespread applications in material science, biology, and medicine. SBD outperforms existing techniques by a wide margin on a simulated benchmark dataset, as well as on real data.
We analyze the generalization capability of SBD, demonstrating that the trained networks are robust to variations of imaging parameters and of the underlying signal structure.
Our results reveal that state-of-the-art architectures for denoising photographic images may not be well adapted to scientific-imaging data. For instance, substantially increasing their field-of-view dramatically improves their performance on TEM images acquired at low signal-to-noise ratios.
We also demonstrate that standard performance metrics for photographs (such as PSNR and SSIM) may fail to produce scientifically meaningful evaluation. We propose several metrics to remedy this issue for the case of atomic resolution electron microscope images. In addition, we propose a technique, based on likelihood computations, to visualize the agreement between the structure of the denoised images and the observed data. Finally, we release a publicly available benchmark dataset of TEM images, containing 18,000 examples.
\end{abstract}



\section{Introduction}

Imaging technology is an essential tool in many scientific domains. Electron microscopy enables the visualization of atomic structures~\cite{book}, fluorescence microscopy makes it possible to study cellular processes~\cite{lichtman2005fluorescence}, and telescopes reveal galaxies and other astronomical objects that are light years away~\cite{mclean2008electronic}. In all these modalities, images are corrupted by noise associated with stochastic processes occurring during signal generation and detection, degrading the information content of the image data. The general goal of denoising is to estimate and restore the information missing from these noisy observations, thus facilitating the extraction of useful scientific information. 

In the past decade, convolutional neural networks (CNNs) \cite{lecun2015deep} have achieved state-of-the-art performance in image denoising~\cite{zhang2017beyond, chen2016trainable}. However, the potential of this methodology has barely been explored in the context of scientific imaging. In the vast majority of the existing work, noisy data are generated by adding Gaussian noise to clean photographs. The CNNs are then trained to approximate the ground-truth images from these measurements, usually by minimizing mean squared error~\cite{zhang2017beyond}.  Unfortunately, \emph{this paradigm is not adequate for most scientific domains}, where large, labeled datasets of ground-truth clean data are typically not available. 
To address this issue, we propose a simulation-based denoising (SBD) framework, in which CNNs are trained on simulated images. We validate our methodology through a case study in transmission electron microscopy. 

\begin{figure*}
    \centering
    \begin{tabular}{c@{\hskip 0.01in}c@{\hskip 0.01in}c@{\hskip 0.01in}c@{\hskip 0.01in}c@{\hskip 0.01in}c}
    \footnotesize{(a) Data} &  \footnotesize{(b) Spot Filter} & \footnotesize{(c) PURE-LET} & \footnotesize{(d) SBD } & \footnotesize{(e) Likelihood Map}& \\
    \includegraphics[width=1.05in]{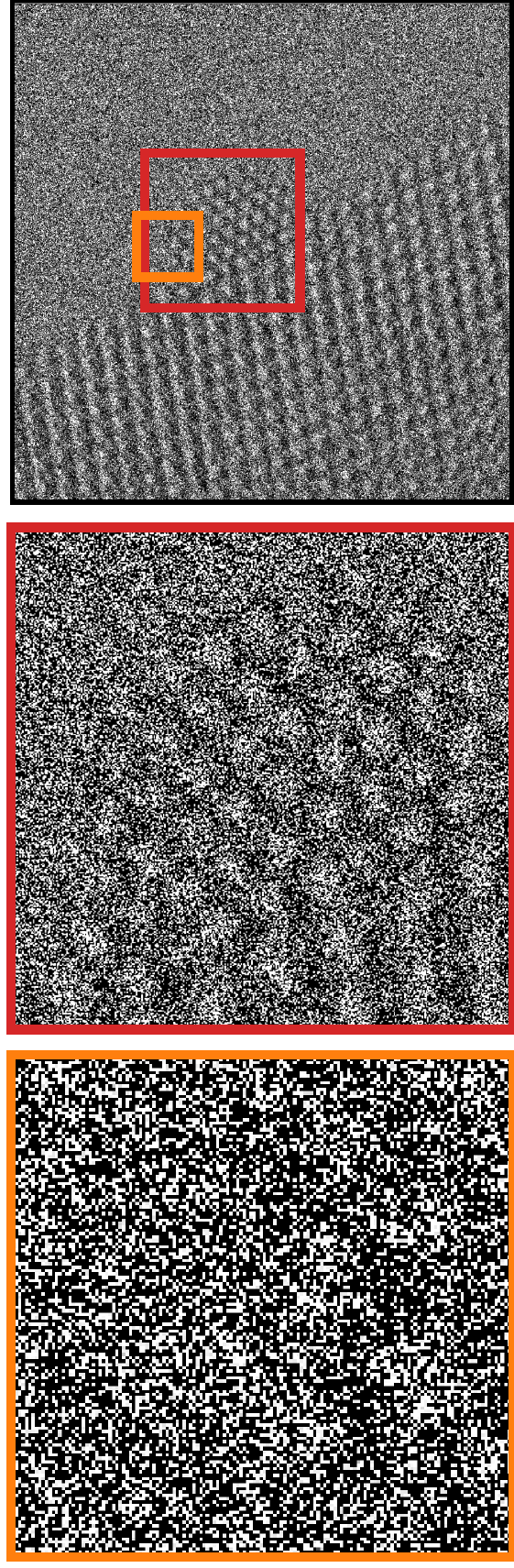}&
    \includegraphics[width=1.05in]{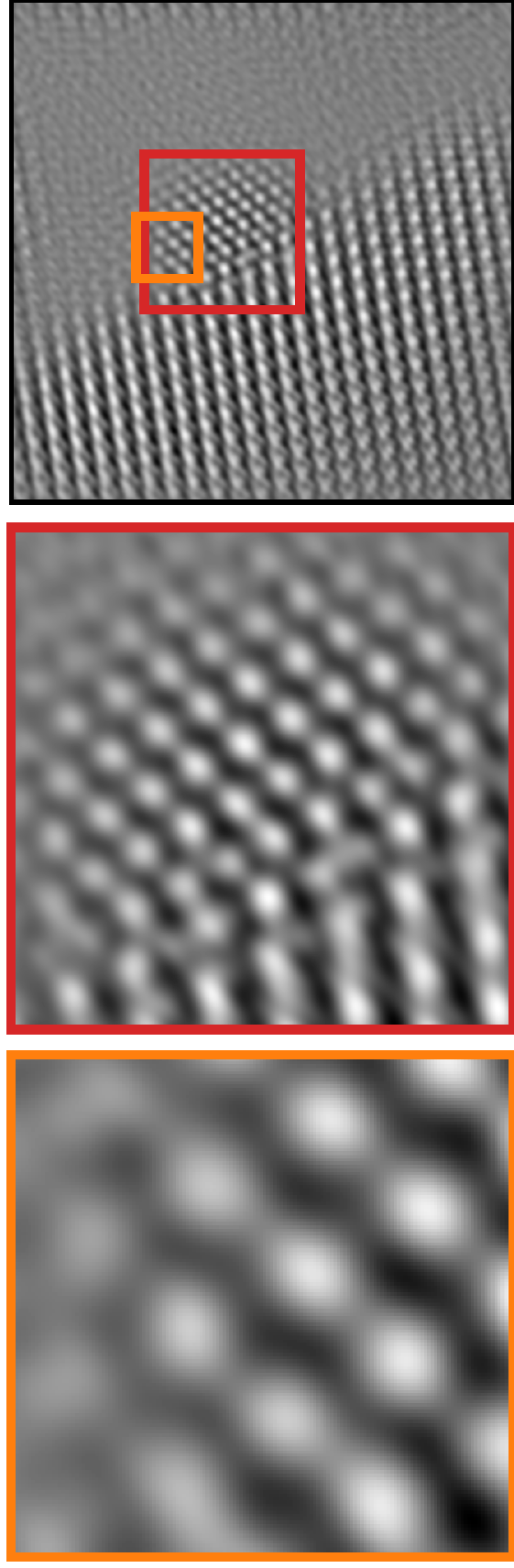}&
    \includegraphics[width=1.05in]{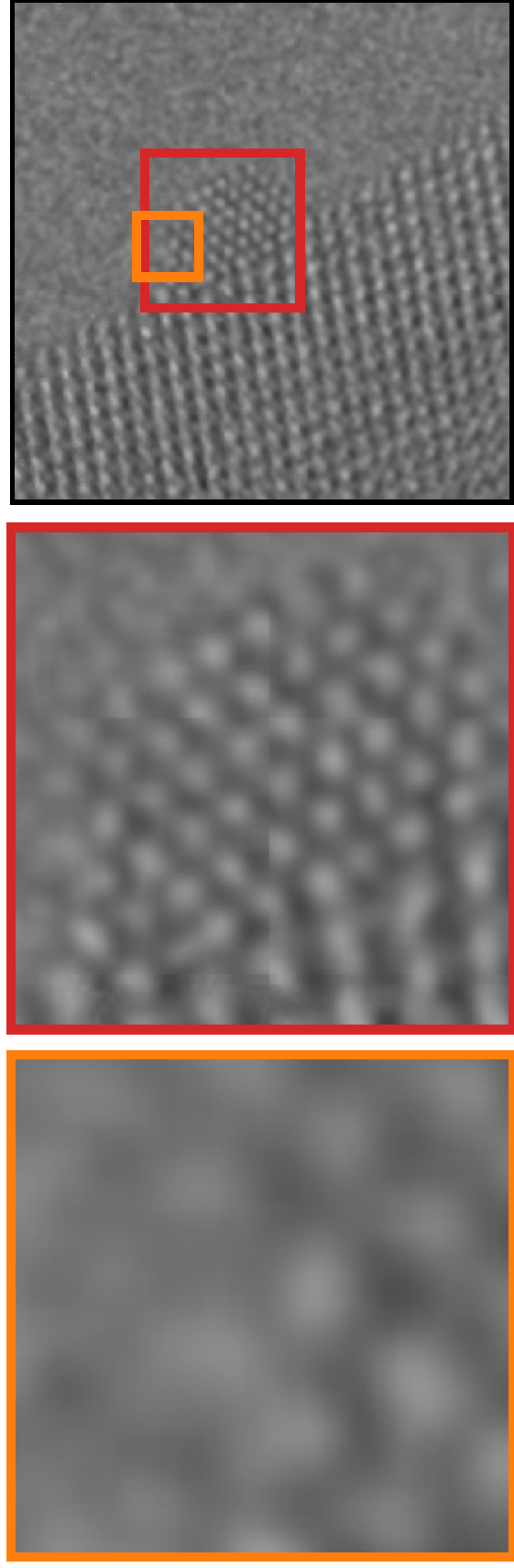}&
    \includegraphics[width=1.05in]{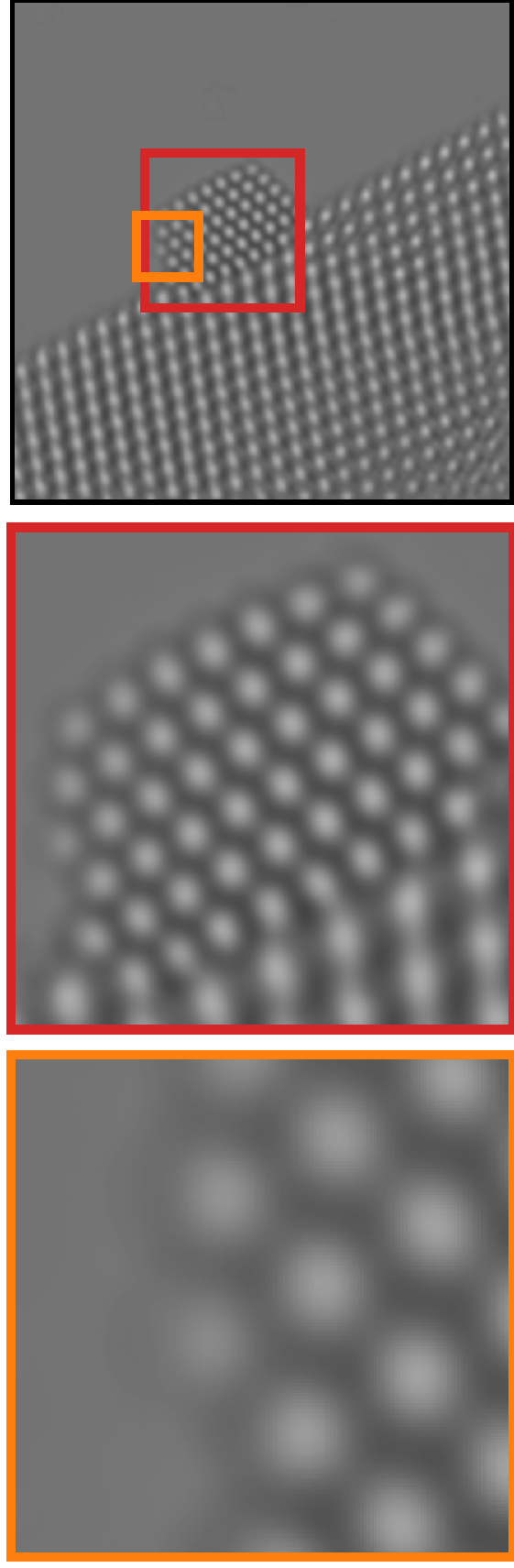}&
    \includegraphics[width=1.05in]{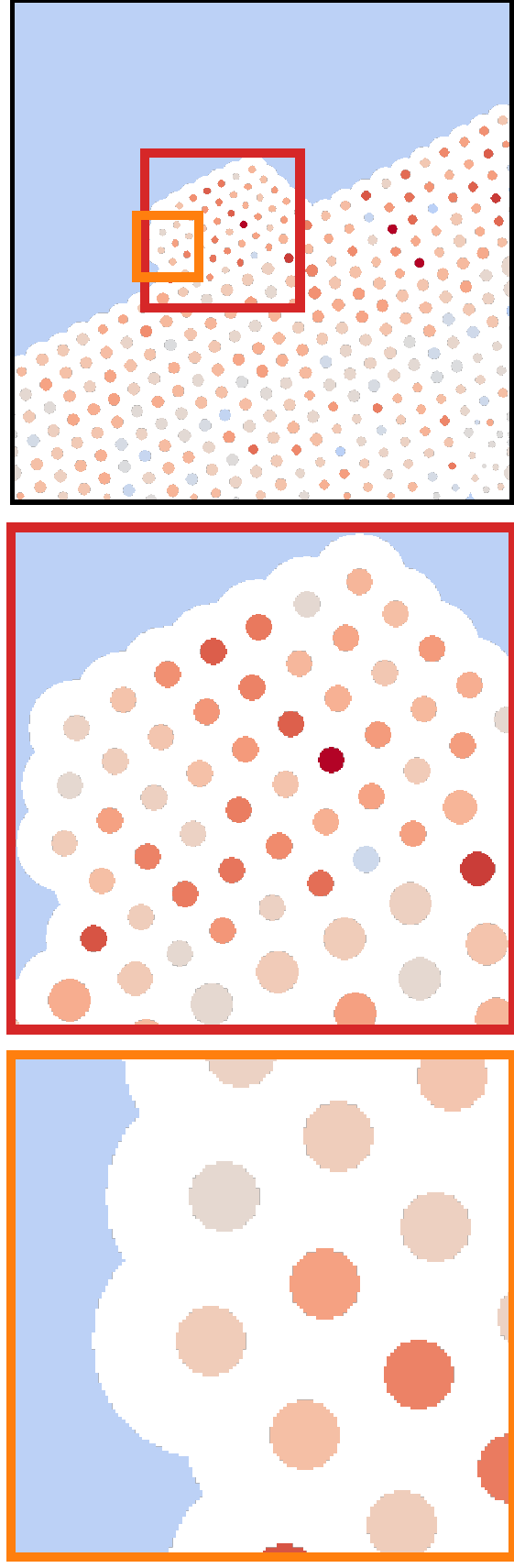}&
    \includegraphics[width=0.48in]{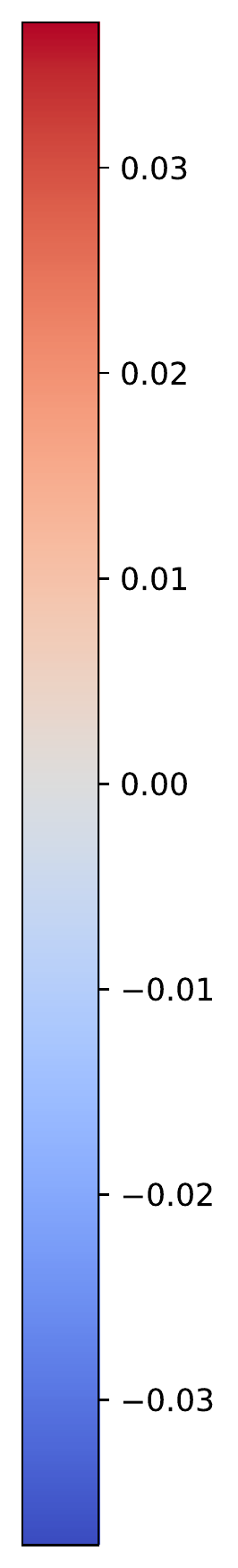}\\
    \end{tabular}
    \caption{\textbf{Denoising results for real data.} (a) An experimentally-acquired atomic-resolution transmission electron microscope image of a CeO2-supported Pt nanoparticle. A description of the experimental data acquisition is given in Section~\ref{sec:dataset}. The average image intensity is 0.45 electrons/pixel (i.e., a large fraction of pixels register zero electrons), which results in an extremely low signal-to-noise ratio. (b) Denoised image obtained via Fourier-based filtering by a domain expert. (c) Denoised image obtained via the wavelet-based PURE-LET method~\cite{luisier2010image}. (d) Denoised image obtained by the proposed simulation-based denoising (SBD) framework. (e) Likelihood map quantifying to what extent the atomic structure identified from the SBD denoised image is consistent with the data (see Section~\ref{sec:methodology}). Regions in red are more likely to correspond to atomic columns in the nanoparticle. Regions in blue are more likely to belong to the vacuum. }
    \label{fig:experiment-denoised}
\end{figure*}

Transmission electron microscopy (TEM) is a powerful and versatile characterization technique used to probe the atomic-level structure and composition of a wide range of materials, such as catalysts or semiconductors~\cite{Smith2015, Crozier-insitu2016}. The technique has had a huge impact in structural biology, as recognized with the award of the 2017 Nobel Prize in Chemistry~\cite{Cryo2016}. Recent advancements in direct electron detection systems enable experimentalists to 
image dynamic events at frame rates in the kilohertz range~\cite{FARUQI2018, ercius2020}. 
Imaging at these time scales is critical to advance our understanding of functional materials. In catalytic systems, for example, the chemical transformation process is accompanied by dynamic, atomic-level structural rearrangements which may occur over a time scale spanning tens of milliseconds~\cite{Sun2020, Guo2020, lawrence_levin_miller_crozier_2020, LEVIN2020, crozier2019dynamics}. 
Acquiring image series at such high temporal resolution necessarily produces datasets that 
are severely degraded by shot noise, rendering traditional imaging processing approaches ineffective. 
It is typically not feasible to reduce the noise content by increasing the intensity of the incident electron beam, since the high-energy beam can also damage the material when exposed to high doses. 
Consequently, there is an acute need for novel denoising technology in this domain.

In order to apply the proposed SBD framework to TEM data, 
we generate a large simulated dataset of TEM images, containing 18,000 examples, and use it to train CNNs for noise removal. This approach outperforms existing techniques by a wide margin on held-out simulated data, as well as on real TEM measurements (see Figures~\ref{fig:experiment-denoised},~\ref{fig:simulation-denoised} and \ref{fig:real-denoised-appendix1} and Sections~\ref{sec:sbd_comparison} and \ref{sec:exp_real_data}). We perform a thorough analysis of the generalization capability of our models, demonstrating that the CNNs are robust to variations of imaging parameters and of underlying signal structure. Our results indicate that architectures optimized for natural photographic images may have fundamental shortcomings when applied to domain-specific data. For instance, we show that substantially increasing the field-of-view of denoising CNNs has almost no effect on photographs, but produces a significant boost in performance for TEM images. We also demonstrate that standard performance metrics for photographs, such as peak signal-to-noise ratio (PSNR) and structural similarity index (SSIM)~\cite{wang2004image}, often fail to produce a scientifically-meaningful evaluation of the denoising results. For example, the presence or absence of a single atomic column often results in a negligible change in these metrics. This is highly problematic, because detecting these columns is one of the main motivations for our case study.
To remedy this issue, we propose several scientifically-motivated metrics to evaluate our results (see Section~\ref{sec:metrics}). 
In addition, we propose a likelihood-based visualization of the agreement between the observed measurements and structures of interest (such as atomic columns) in the denoised image. This visualization can be used to flag denoising artefacts, which may be mistaken for scientifically-relevant structure (see Figure~\ref{fig:likelihood_sim}). 
Finally, to encourage further development of deep-learning methodologies for scientific imaging, we have made our benchmark dataset of TEM images publicly available online\footnote{\href{https://sreyas-mohan.github.io/electron-microscopy-denoising/}{https://sreyas-mohan.github.io/electron-microscopy-denoising/}}. More details on applying the proposed methodology to TEM data, and domain-specific insights derived from the denoised images are described in our companion paper~\cite{vincent2021developing}.

\section{Related work}
\label{sec:related_work}

\subsubsection*{Denoising in scientific imaging}

A wide variety of denoising methods have been applied across different scientific imaging modalities, including traditional linear filters~\cite{nellist1998accurate}, nonlinear filters~\cite{tomasi1998bilateral,milanfar2012tour,jiang2003applications}, wavelet-based methods~\cite{chang2000adaptive,portilla2003image,zhu2015survey,meiniel2018denoising}, and sparsity-based approaches~\cite{meiniel2018denoising,beckouche2013astronomical}. Deep convolutional networks have been shown to outperform all of these approaches in photographic images~\cite{zhang2017beyond,chen2016trainable}. The rapidly growing literature on this methodology focuses almost exclusively on photographic images. We are aware of only a few very recent exceptions. In the medical domain, CNN-based denoising has been applied to low-dose computer tomography~\cite{kim2019performance}, positron-emission tomography~\cite{gong2018pet} and scintillation-camera data~\cite{minarik2020denoising}. Refs. \cite{giannatou2019deep,vasudevan2019deep,ede2019improving} apply CNNs to denoise simulated electron microscopy data, without validating on real data. Ref. \cite{manifold2019denoising} trains CNNs to denoise Raman scattering microscopy data, using measurements gathered at a higher signal-to-noise ratio (SNR) as ground-truth images. These results showcase the potential of deep denoising for scientific imaging, but also the challenge of gathering adequate datasets to train the deep networks. In this work, we propose to address this challenge by training denoising CNNs on carefully-designed simulated datasets, and validate our approach on experimental measurements.

\subsubsection*{Unsupervised denoising}

\noindent Unsupervised denoising is a promising approach for applications where ground-truth images are not available. Unsupervised methods based on wavelets have achieved performance comparable to their supervised counterparts on photographic images~\cite{lusier2007SURE,helor2008dscrim,raphan2008optimal}.\\ Noise2Noise~\cite{lehtinen2018noise2noise}, a deep-learning approach that requires access to pairs of noisy images corresponding to the same underlying signal, has been applied to cryo-electron microscopy  \cite{buchholz2019cryo}. More recent methods can be trained directly on noisy images~\cite{krull2019noise2void,batson2019noise2self,laine2019high}. Several recent works apply this approach to fluorescence microscopy data~\cite{krull2019probabilistic,khademi2020self,prakash2020fully,zhang2019poisson}. 
In the case of our TEM data, standard unsupervised methods do not perform as well as the proposed supervised approach (see Section~\ref{sec:unsupervised}). This is possibly due to the limited number of training data (see Figure~\ref{fig:unsup_data}) and to the low input SNR. The SNR of our TEM data (around 3 dB) is orders of magnitude lower than that reported in typical unsupervised denoising works (e.g. around 27 dB for \cite{zhang2019poisson}). 

\subsubsection*{Deep Learning for TEM}

Deep CNNs have been applied to other image-processing tasks in TEM beyond denoising, see Ref.~\cite{ede2020deep} for a comprehensive review. Ref.~\cite{suveer2019super} proposes a CNN-based method for TEM image super-resolution, wherein CNNs are trained on pairs of low-resolution and high-resolution images acquired experimentally.
Ref.~\cite{horwath2020understanding} applies CNNs to perform segmentation and systematically studies the influence of the design of the training dataset and network architecture on the generalization capabilities of these models. In this work, we provide a similar analysis for denoising. Refs. \cite{madsen2018deep} and \cite{ragone2020atomic} propose a CNN-based method to identify  structures of interest in TEM images. They train on carefully designed simulated data and show that the model generalizes to real data. Our work provides further evidence that CNNs trained on simulated data can generalize effectively to real measurements. 


\begin{figure}
    \centering
    \includegraphics[width=\linewidth]{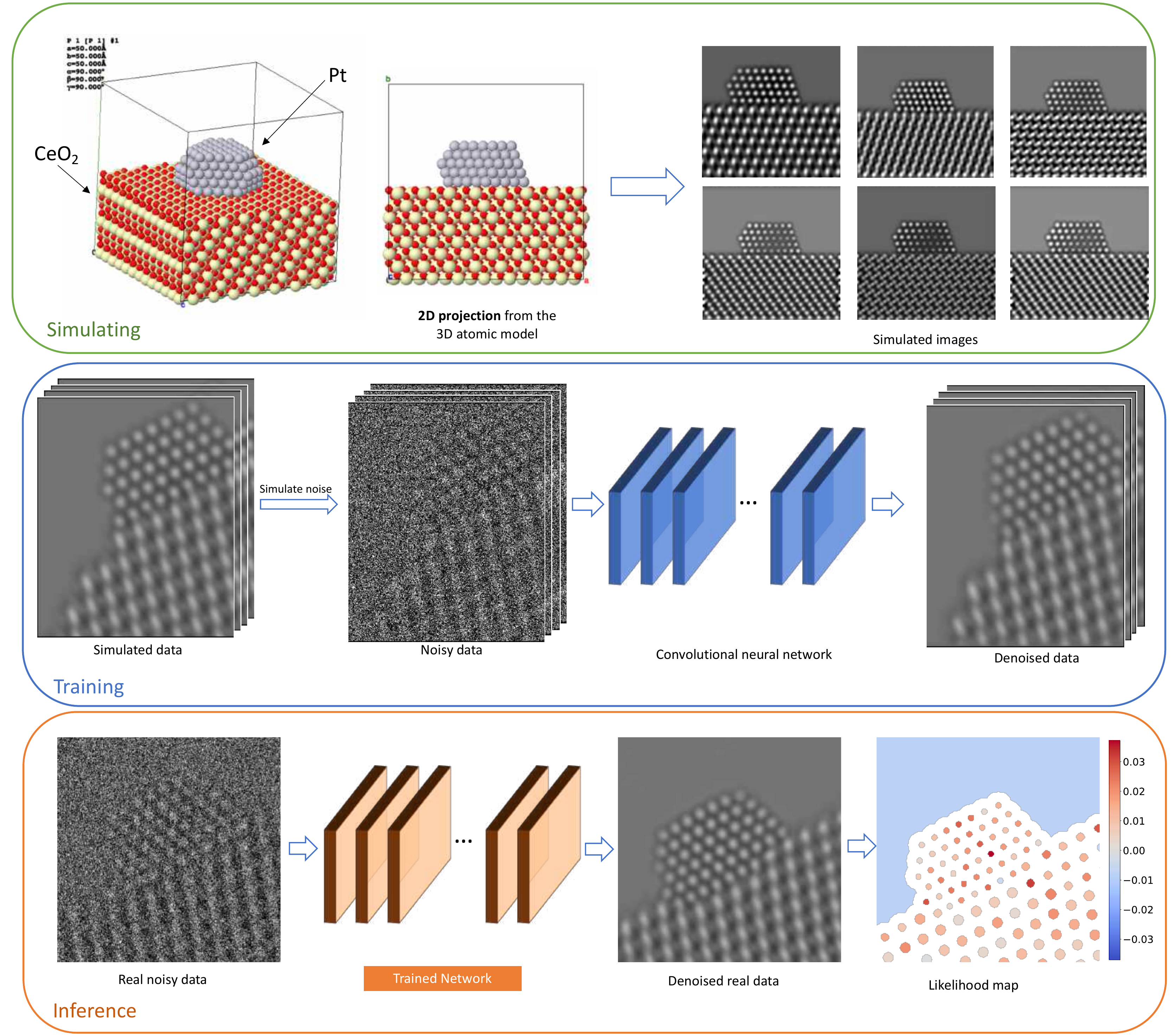}
    \caption{\textbf{Simulation-based denoising framework}. (Top) A training dataset is generated by simulating TEM images of different structures at varying imaging conditions. Here we focus on structures of Pt nanoparticles supported on CeO$_2$. (Middle) A CNN is trained using the simulated images, paired with noisy counterparts obtained by simulating the relevant noise process. (Bottom) The trained CNN is applied to real data to yield a denoised image. After analyzing the image to extract structures of interest, a likelihood map is generated to quantify the agreement between this structure and the noisy data. }
    \label{fig:methodology}
\end{figure}

\section{Methodology}
\label{sec:methodology}

\subsection{Simulation-based denoising}

Current state-of-the-art deep-learning techniques for denoising photographic images require a training set of ground-truth images~\cite{zhang2017beyond}. Typically these clean images are corrupted with additive Gaussian noise, and the CNNs are trained to minimize the mean squared error between the network output and the original images. The main obstacle to leveraging this approach in scientific imaging is the lack of ground-truth data; in many applications there is no such thing as a \emph{clean image}. We address this by using a dataset of simulated images to train the CNNs. We call this framework simulation-based denoising (SBD). 

Simulation-based denoising (SBD) consists of three stages: simulation of the training set, training of the CNNs using the simulated data, and inference on the real data (see Figure~\ref{fig:methodology} for an overview of the methodology). In order to generate the training set, we simulate clean images $x_1$, \ldots, $x_{N} \in \mathbb{R}^{M}$ (where $M$ is the number of pixels) according to a predefined physical model. These clean images are then corrupted using a noise model, which can follow a predefined model or be learned from the data, to generate the simulated noisy data. We provide a detailed account of how we generate the simulated dataset for our case study in Sections~\ref{sec:dataset} and~\ref{sec:data_simulation}, and of the noise model in Section~\ref{sec:noise_model}. Let $Y(x_{i})$ denote the random vector representing the noisy image corresponding to the clean simulated image $x_{i}$ and let $y(x_i)$ represent a realization of $Y(x_{i})$. We parameterize the denoising function as a CNN $f_\theta:\mathbb{R}^{M} \rightarrow \mathbb{R}^M$ where the parameters $\theta$ are the weights of the network. 
To find a good denoising function $f_\theta$, we minimize a loss function  $\mathcal{L}:\mathbb{R}^{M} \times \mathbb{R}^{M} \rightarrow \mathbb{R}$ which quantifies how close the estimate from the CNN $f_\theta(y(x_i))$ is to the clean image $x_i$. In our case study, we use mean squared error, which is a standard choice in CNN-based denoising~\cite{zhang2017beyond}. More concretely, during the training stage, we compute the parameters by solving
\begin{equation}
\label{eq:mse}
\hat{\theta} = \arg\min_\theta \mathbb{E} \left[\sum_{i=1}^{N} \mathcal{L}( f_\theta(Y(x_i)), x_i)  \right] = \arg\min_\theta \mathbb{E} \left[\sum_{i=1}^{N} \| f_\theta(Y(x_i)) - x_i\|_2^2  \right]
\end{equation}
\noindent
We perform minimization iteratively using a variant of stochastic gradient descent. 
We approximate the expectation in Equation~\ref{eq:mse} by sampling new realizations of the noisy image $Y(x_i)$ every time we compute the gradient. 
Once the network 
is trained, it can be directly applied to new noisy images to perform denoising.

\begin{figure}
\centering
\begin{subfigure}{.5\textwidth}
  \centering
  \includegraphics[width=1\linewidth]{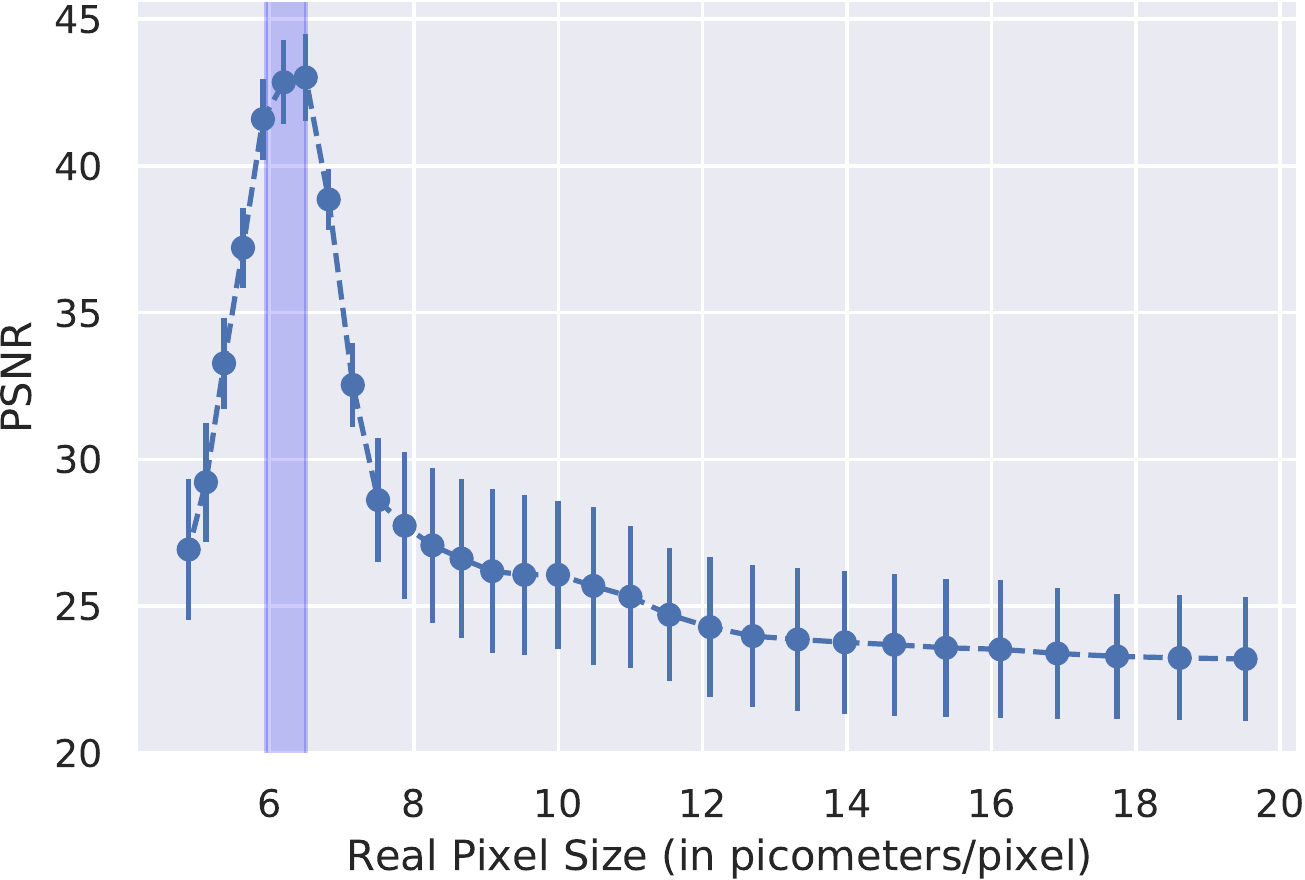}
\end{subfigure}%
\begin{subfigure}{.5\textwidth}
  \centering
  \includegraphics[width=1\linewidth]{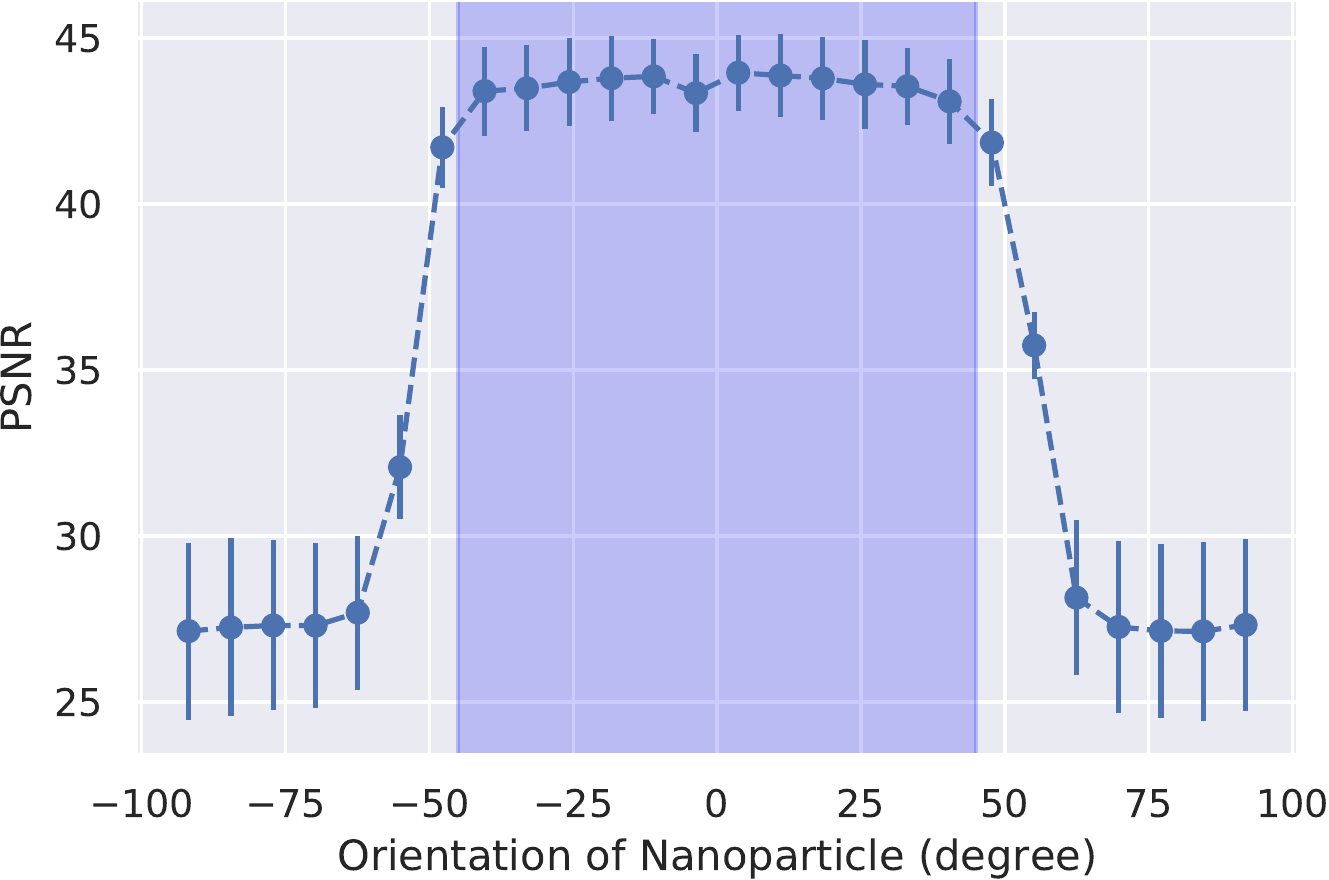}
\end{subfigure}
\caption{\textbf{Overfitting scaling and orientation}. The plots show the performance of the proposed network in PSNR for simulated images that have different scalings (on the left, measured in terms of the corresponding pixel size in picometers) and orientations (on the right). The CNN was trained on data augmented with rescaled and rotated images corresponding to the regions shaded in purple. When tested out of those regions, the denoising performance deteriorates significantly. This shows that careful data augmentation is required to ensure invariance to scaling and orientation.
}
\label{fig:pitfall}
\end{figure}

A crucial difference between SBD and previous methodology for deep denoising is that the training set needs to be explicitly \emph{designed}. In order to ensure effective generalization to real data, we must include sufficient variation of imaging parameters and image structure in the training dataset. In addition, particular care is needed to enforce  invariance to small changes in the geometry of the image. Figure~\ref{fig:pitfall} shows that a denoising CNN can easily overfit the specific alignment and scale of the training data. This issue can be addressed by augmenting the training set with rotated and scaled versions of the simulated images. 
Determining how to optimally sample the space of possible simulation parameters when generating data to train CNNs for denoising is an important methodological question for future research.

\subsection{Exploiting non-local signal structure}
\label{sec:non_local}

Images in scientific applications often have pixel-intensity distributions that differ significantly from those of natural images. Our case study shows that it is crucial to take this into account in order to achieve successful denoising. 
Current state-of-the-art networks for denoising photographic images have very small fields of view. For example, the field of view of DnCNN~\cite{zhang2017beyond} and DURR~\cite{zhang2018dynamically} are $41 \times 41$ pixels, and $45 \times 45$ pixels respectively. Unlike photographic images, the TEM images in our case study exhibit very prominent global regularities, due to periodicity in the atomic structure of the imaged materials. 
In addition, electron-microscopy images are often measured at very low SNRs (in our case, the SNR for the real TEM data is about 3 dB, but most works for photographic images focus on an SNR above 22 dB, see e.g.~\cite{zhang2017beyond}). As the SNR decreases, denoising CNNs tend to average over larger regions of the surrounding pixels, as demonstrated in Ref.~\cite{Mohan2020Robust} (qualitatively, this is the same behavior observed in a classical linear Wiener filter~\cite{wiener1950extrapolation}). These considerations motivate using networks with large field of view to denoise TEM data. 

\begin{table}[]
    \centering
    \textbf{(a) TEM Images}
    \vspace{2mm}
    \begin{tabular}{lcccc}
        \toprule
        \textsc{Model} & Parameters & FoV & PSNR & SSIM  \\
        \midrule
        SBD + DnCNN \cite{zhang2017beyond} & 668K & 41 $\times$ 41 & 30.47 $\pm$ 0.64 & 0.93 $\pm$ 0.01 \\
        SBD + Small UNet \cite{zhang2018dynamically} & 233K & 45 $\times$ 45 & 30.87 $\pm$ 0.56 & 0.93 $\pm$ 0.01 \\
        SBD + UNet (32 base channels) & 352K & 221 $\times$ 221 & 36.39 $\pm$ 0.77 & 0.98 $\pm$ 0.01 \\
        SBD + UNet (64 base channels) & 1.41M & 221 $\times$ 221 & 37.24 $\pm$ 0.76 & 0.99 $\pm$ 0.01 \\
        SBD + UNet (128 base channels) & 5.61M & 221 $\times$ 221 & 38.05 $\pm$ 0.81 & 0.99 $\pm$ 0.01 \\
        SBD + UNet (128 base channels) & 70.15M & 893 $\times$ 893 & 42.87 $\pm$ 1.45 & 0.99 $\pm$ 0.01 \\
        \bottomrule
    \end{tabular}\\
    \vspace{2mm}
    \textbf{(b) Photographic Images}
    
     \begin{tabular}{ccccccc}
        \toprule
        \textsc{Model} & Parameters & FoV & \multicolumn{2}{c}{PSNR} & \multicolumn{2}{c}{SSIM}  \\
        
        \cmidrule(lr){4-5}
        \cmidrule(lr){6-7} 
        
        & & & $\sigma=30$ & $\sigma=70$ & $\sigma=30$ & $\sigma=70$ \\
        
        \midrule
        UNet & 102K & 49 $\times$ 49 & 29.67 $\pm$ 2.84 & 26.16 $\pm$ 2.79 & 0.83 $\pm$ 0.06 & 0.70 $\pm$ 0.09 \\
        UNet & 352K & 221 $\times$ 221 & 29.65 $\pm$ 2.76 & 26.08 $\pm$ 2.68 & 0.83 $\pm$ 0.05 & 0.70 $\pm$ 0.08 \\
        UNet & 4.4M & 893 $\times$ 893 & 29.54 $\pm$ 2.82 & 26.07 $\pm$ 2.80 & 0.83 $\pm$ 0.06 & 0.70 $\pm$ 0.09 \\
        \bottomrule
    \end{tabular}
    \caption{\textbf{Field of view of CNN architectures and performance.} Mean PSNR and SSIM ($\pm$ standard deviation) of different CNN architectures on the (a) held-out simulated test set of TEM data described in Section~\ref{sec:sbd_comparison} and (b) validation set of the DIV2K photographic image dataset~\cite{Agustsson_2017_CVPR_Workshops}. For TEM images,  increasing the field of view~(FoV) of the UNet from $45\times 45$ pixels to $221\times 221$ produces a dramatic increase of around 6 dB in PSNR, even if the number of parameters remains similar. Increasing the number of parameters while keeping the field of view constant produces a modest gain in performance. In contrast, changing the field of view has almost no effect on denoising photographic images. The networks used for photographic images were trained on $512 \times 512$ patches extracted from training images of DIV2K~\cite{Agustsson_2017_CVPR_Workshops} corrupted with additive Gaussian noise with standard deviation $\sigma \in [0, 100]$.}
    \label{tab:network_comparison}
\end{table}

Here we propose to denoise TEM data using deep networks with very large fields of view: $221\times 221$ pixels and $893 \times 893$ pixels, a 25-fold and 400-fold increase with respect to generic denoising architectures respectively. In order to obtain a large field of view efficiently (i.e. without dramatically increasing the number of parameters in the network), we propose using a UNet network architecture~\cite{ronneberger2015u}. We use $4$ downsampling operations to achieve the $221 \times 221$ field of view and $6$ downsampling operations to achieve the $893 \times 893$ field of view (see Section~\ref{sec:architectures} for a detailed description of the architecture). Table~\ref{tab:network_comparison} compares the influence of the field of view in denoising photographic and TEM images. For photographic images the performance of the network remains almost constant as we increase the field of view. In contrast, for TEM images increasing the field of view produces a dramatic improvement in performance (6 dB and 10 dB, when the field of view is $221 \times 221$ and $893 \times 893$ respectively). Increasing the number of parameters, while keeping the field of view constant, has a very modest effect, which suggests that the increase in field of view is the primary reason for the improvement. 

\newcolumntype{C}[1]{>{\centering\arraybackslash}m{#1}}
\begin{figure*}
    \centering
    \begin{tabular}{C{0.21\linewidth}@{\hskip 0.07in}C{0.21\linewidth}@{\hskip 0.07in}C{0.21\linewidth}@{\hskip 0.07in}C{0.21\linewidth}@{\hskip 0.07in}C{0.049\linewidth}} \\
        \footnotesize{(a) Noisy} & \footnotesize{(b) Denoised} & \footnotesize{(c) Denoised (zoomed)} & \footnotesize{(d) Gradient} \\
        \includegraphics[width=0.21\textwidth]{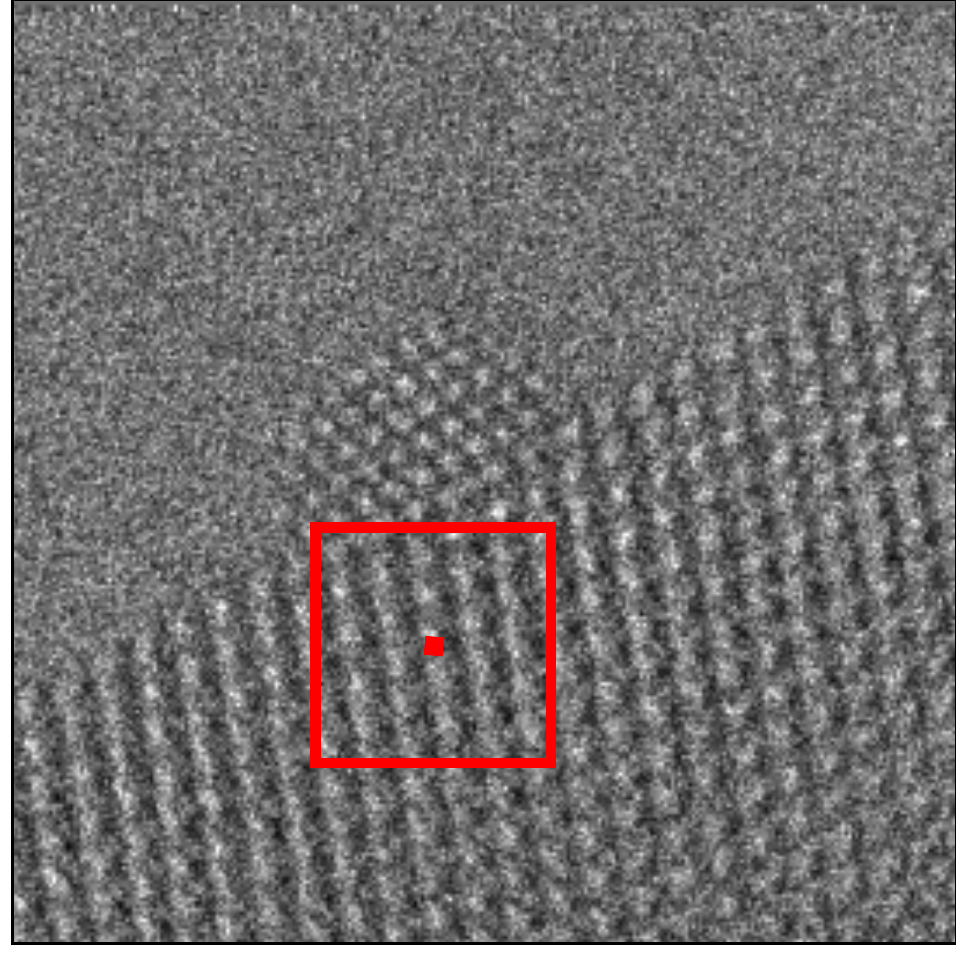} &
        \includegraphics[width=0.21\textwidth]{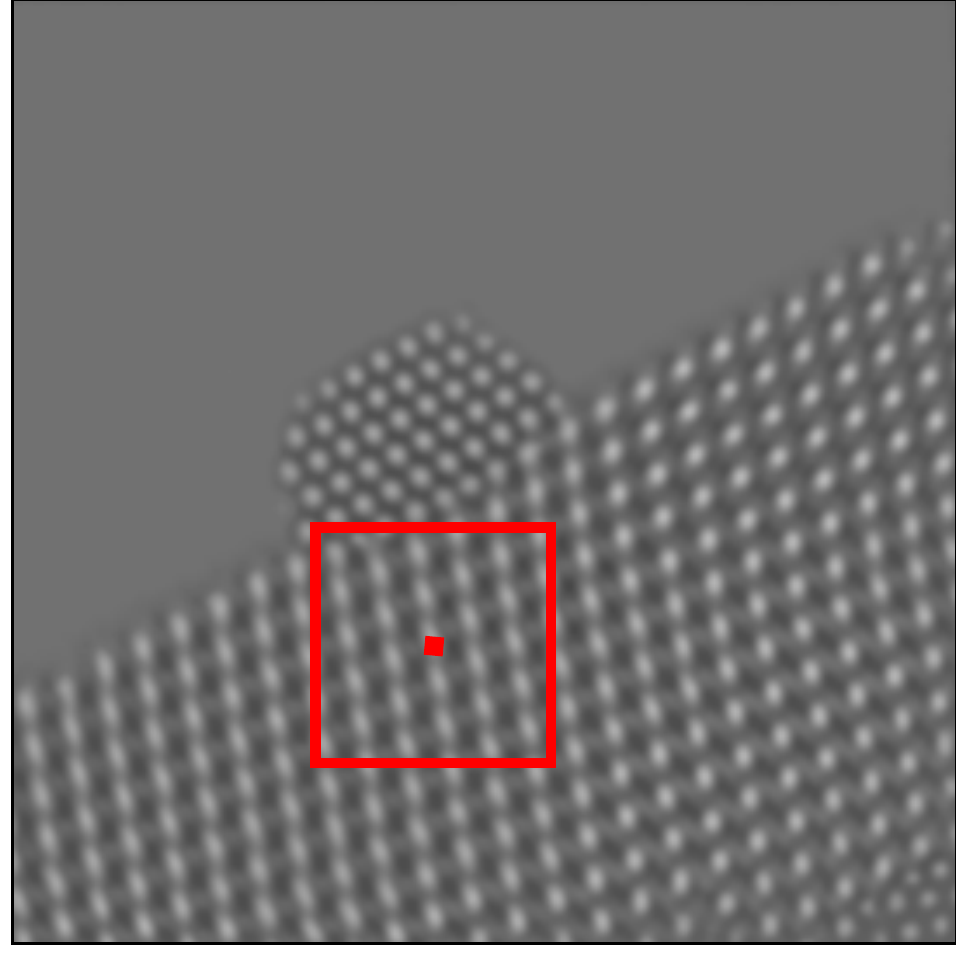} &
        \includegraphics[width=0.21\textwidth]{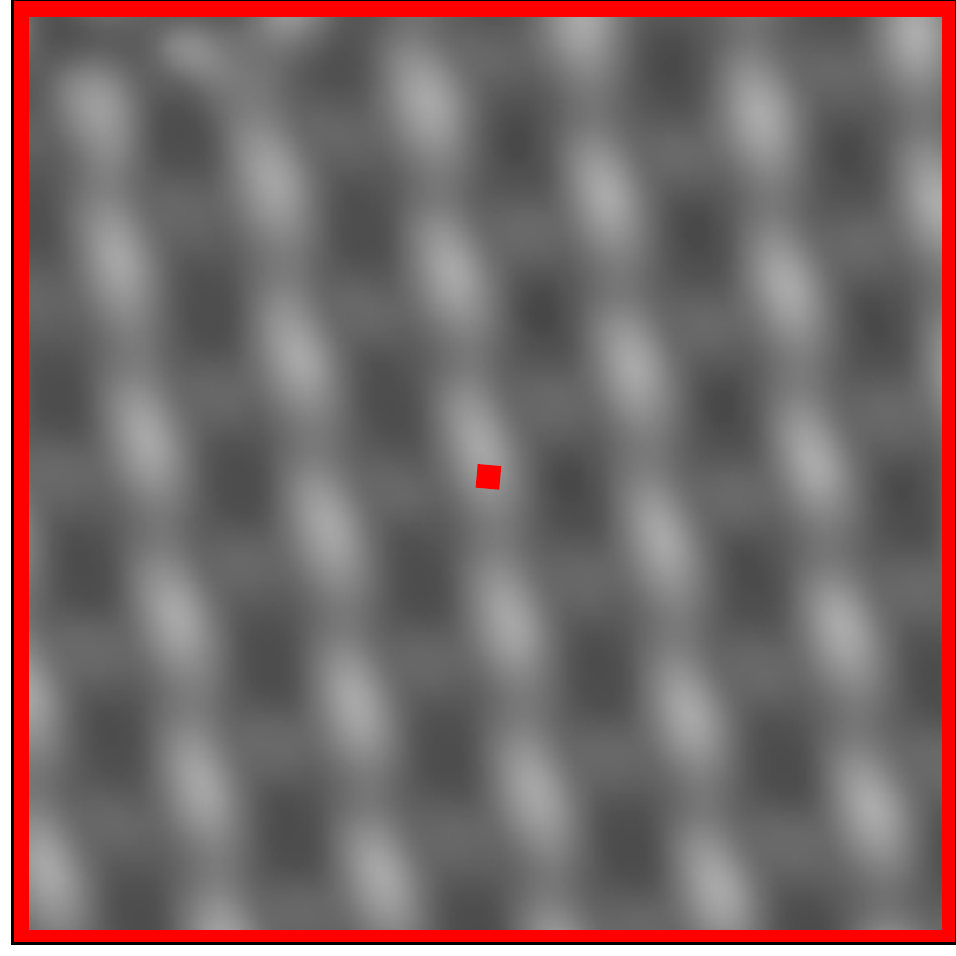} &
        \includegraphics[width=0.21\textwidth]{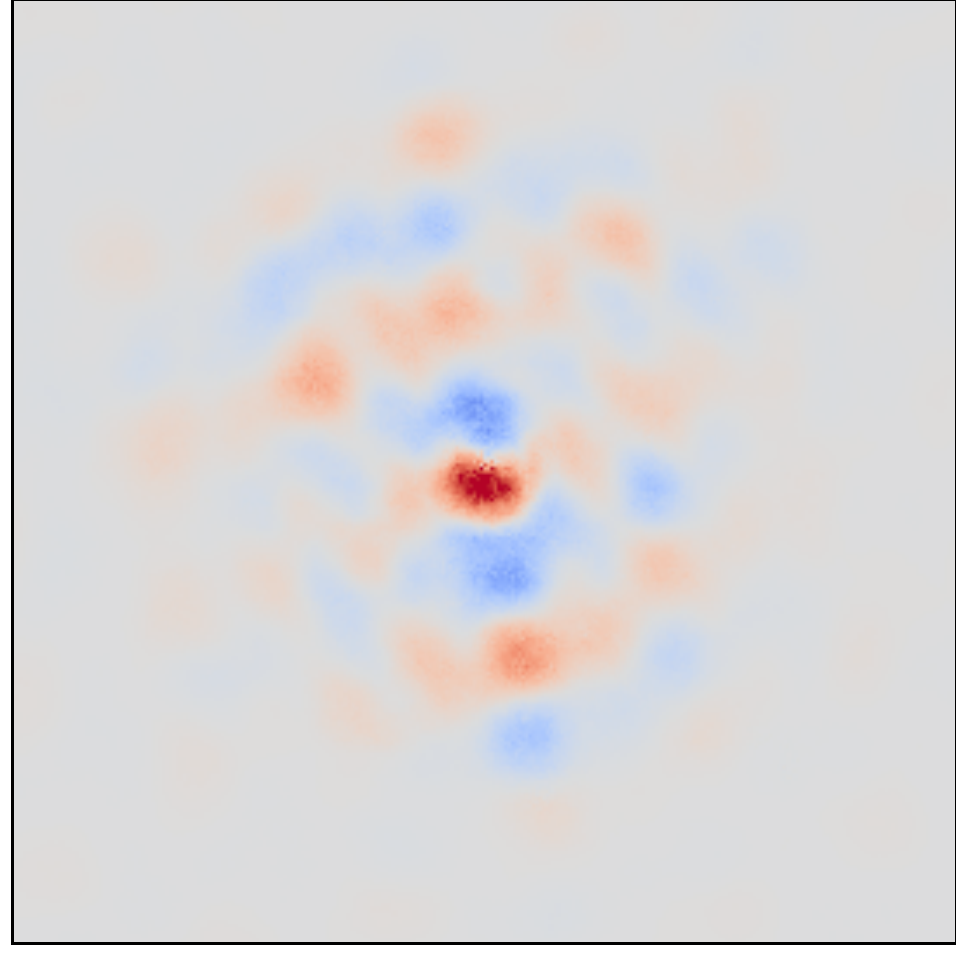} &
        \includegraphics[width=0.049\textwidth]{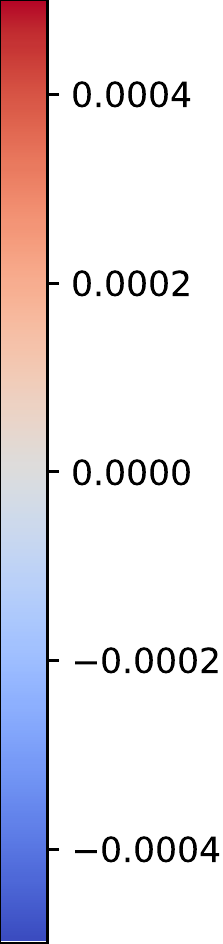} \\
        \includegraphics[width=0.21\textwidth]{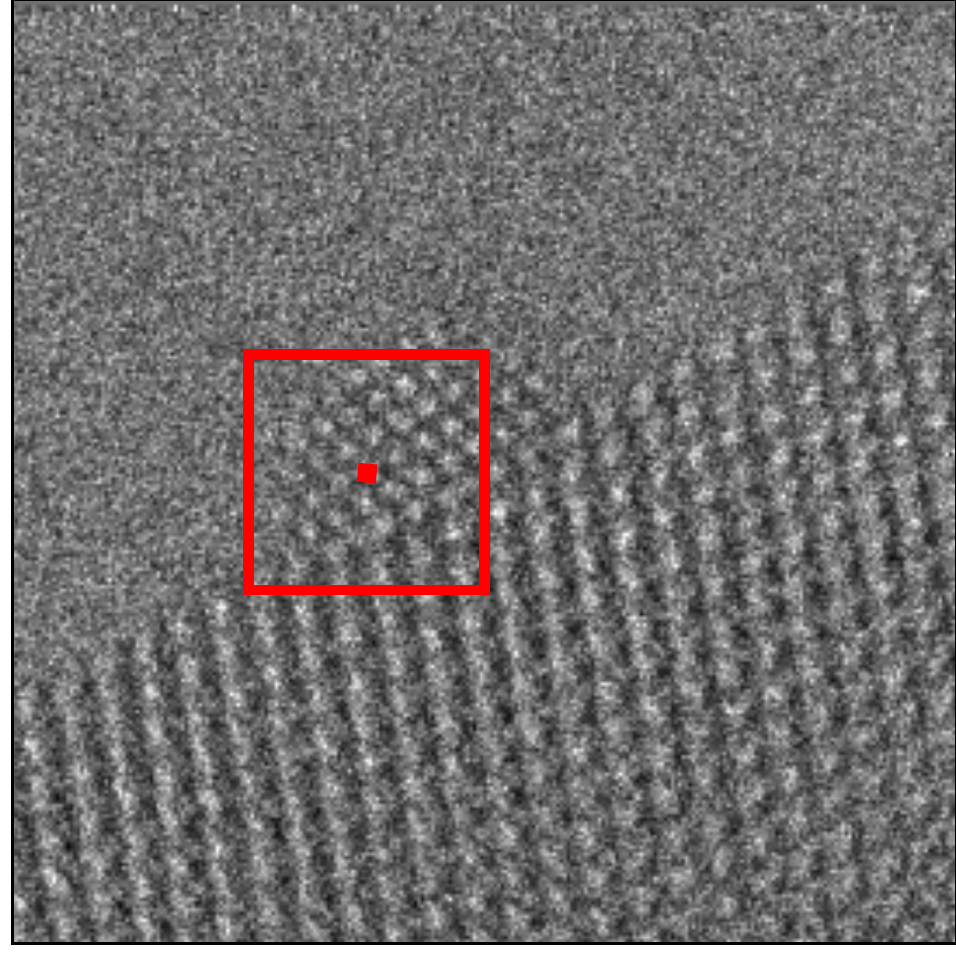} &
        \includegraphics[width=0.21\textwidth]{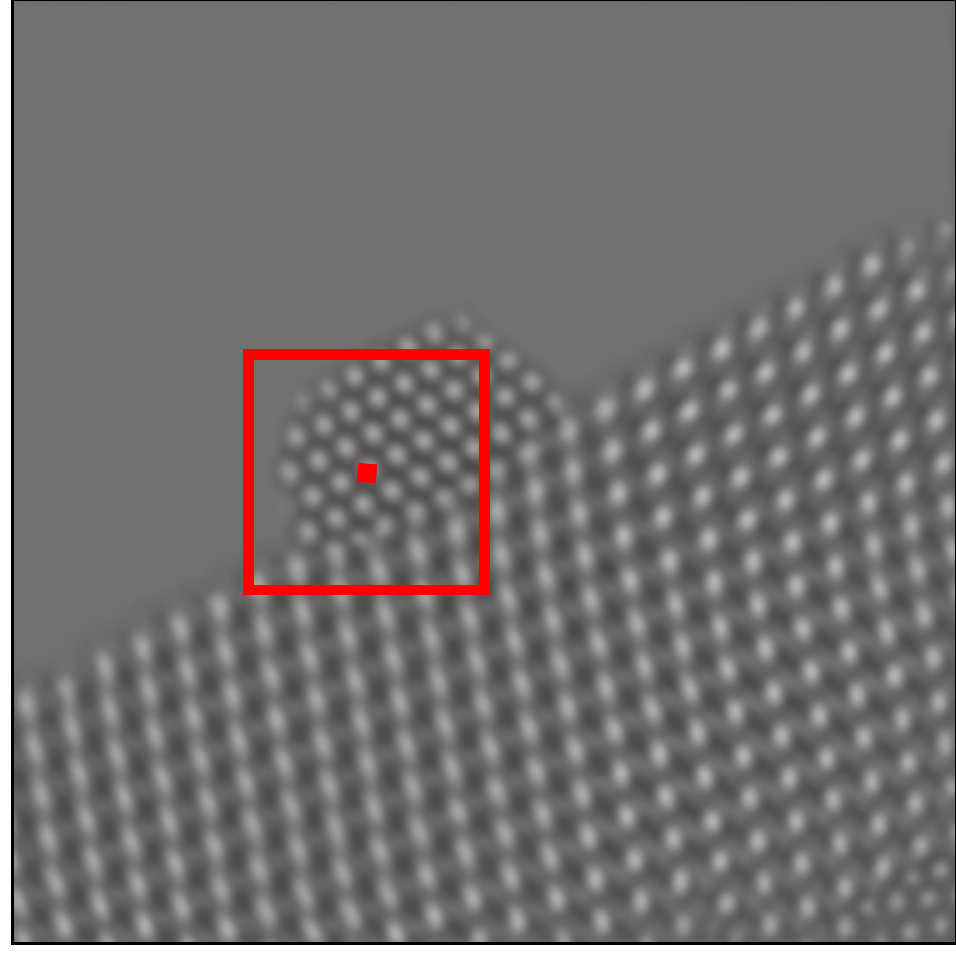} &
        \includegraphics[width=0.21\textwidth]{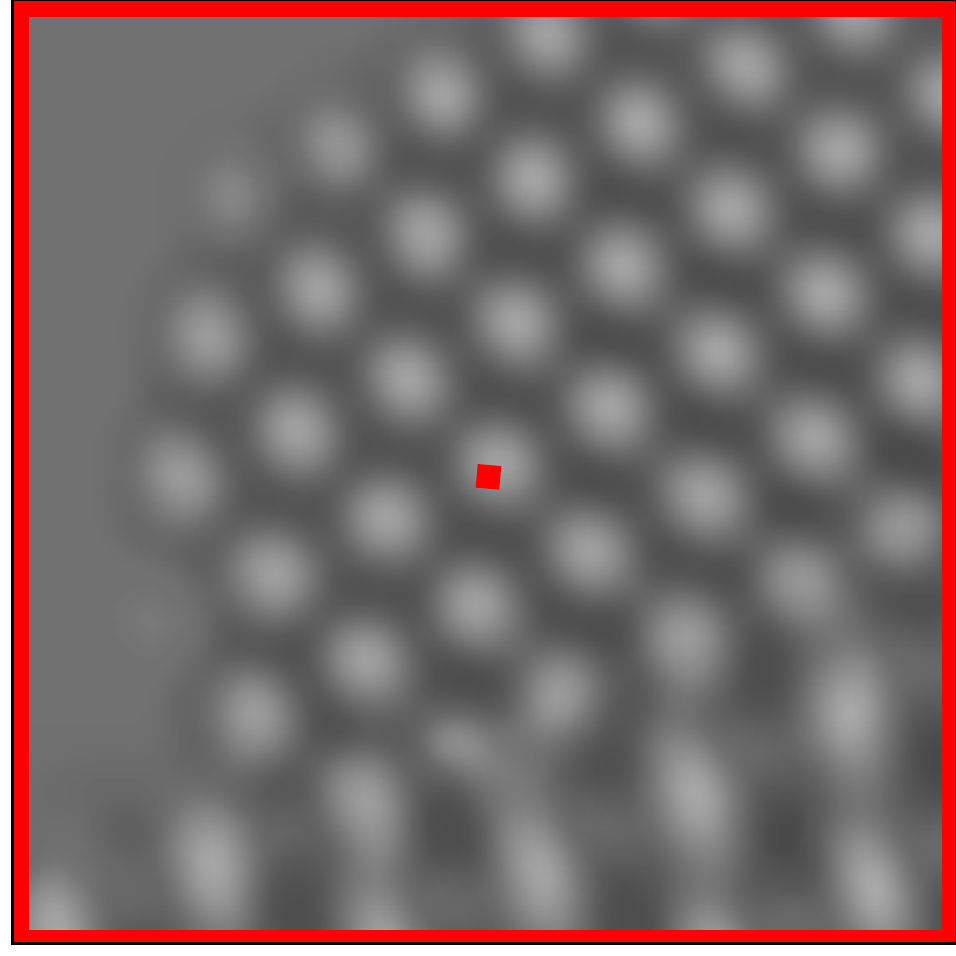} &
        \includegraphics[width=0.21\textwidth]{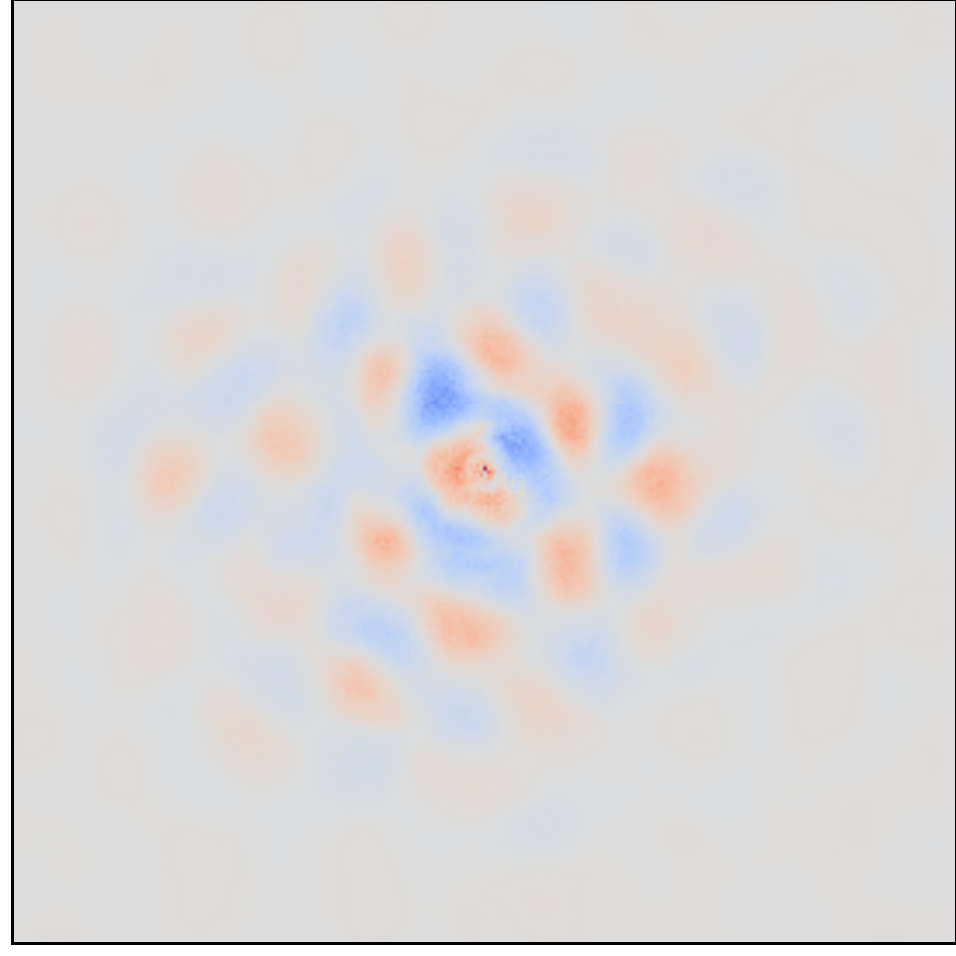} &
        \includegraphics[width=0.049\textwidth]{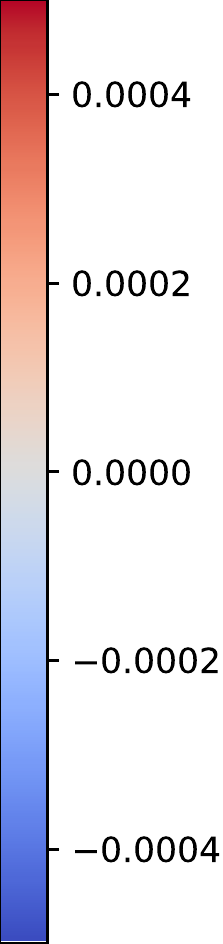} \\
        \includegraphics[width=0.21\textwidth]{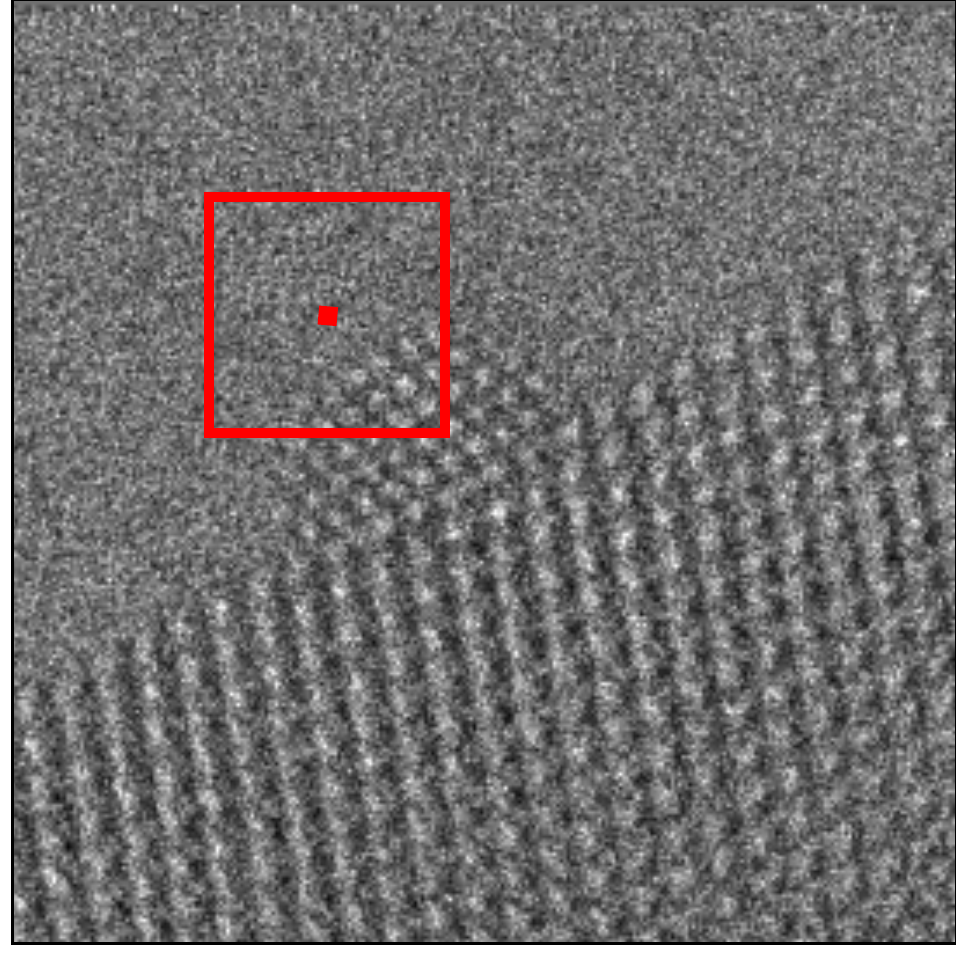} &
        \includegraphics[width=0.21\textwidth]{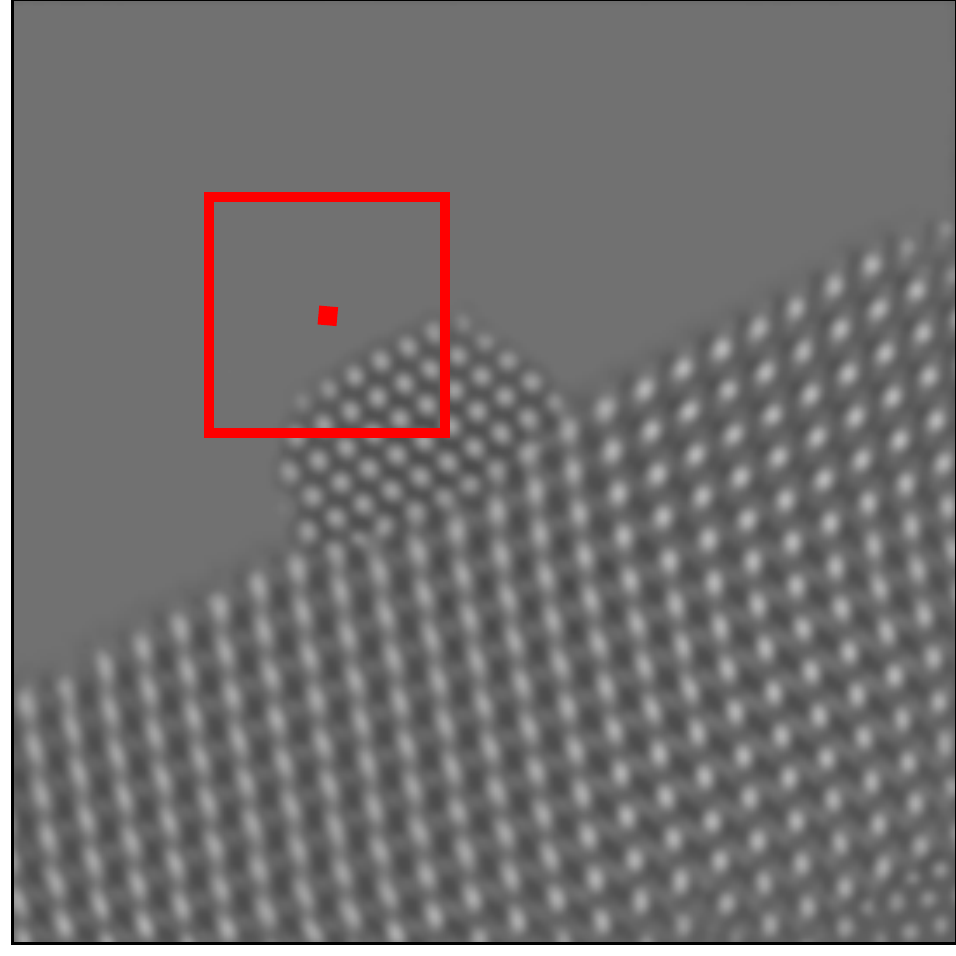} &
        \includegraphics[width=0.21\textwidth]{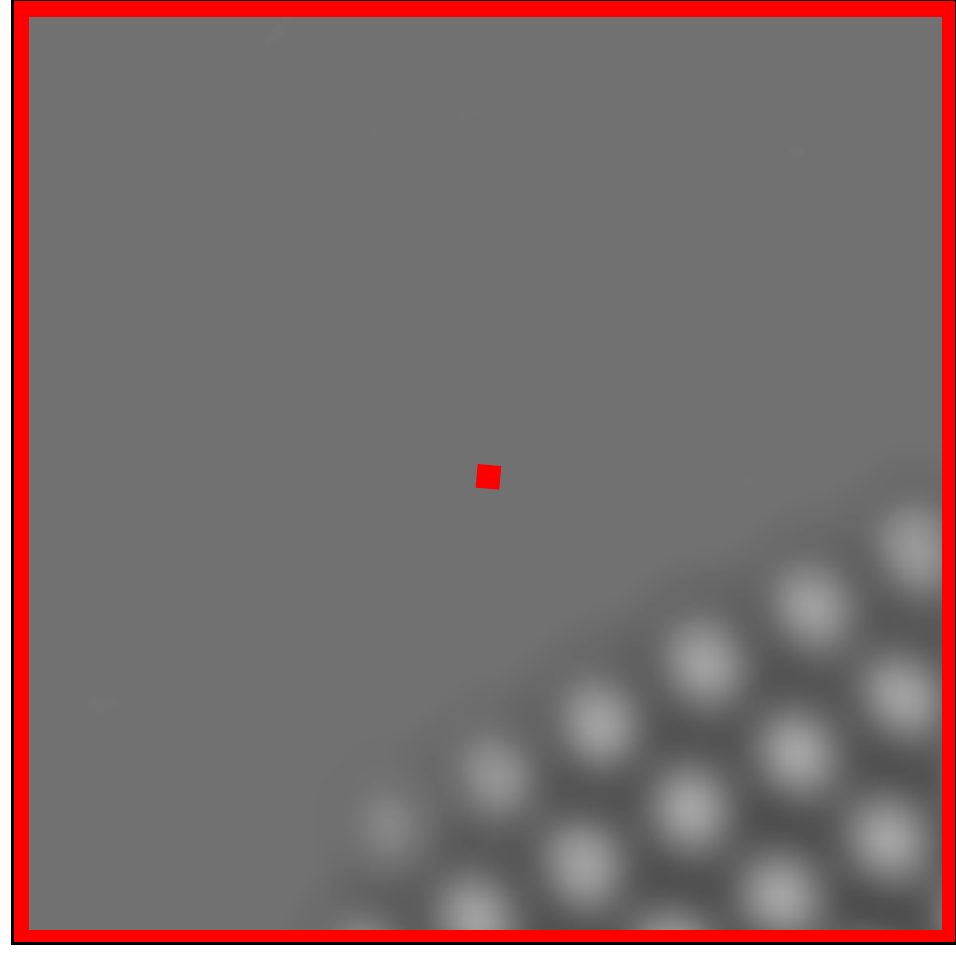} &
        \includegraphics[width=0.21\textwidth]{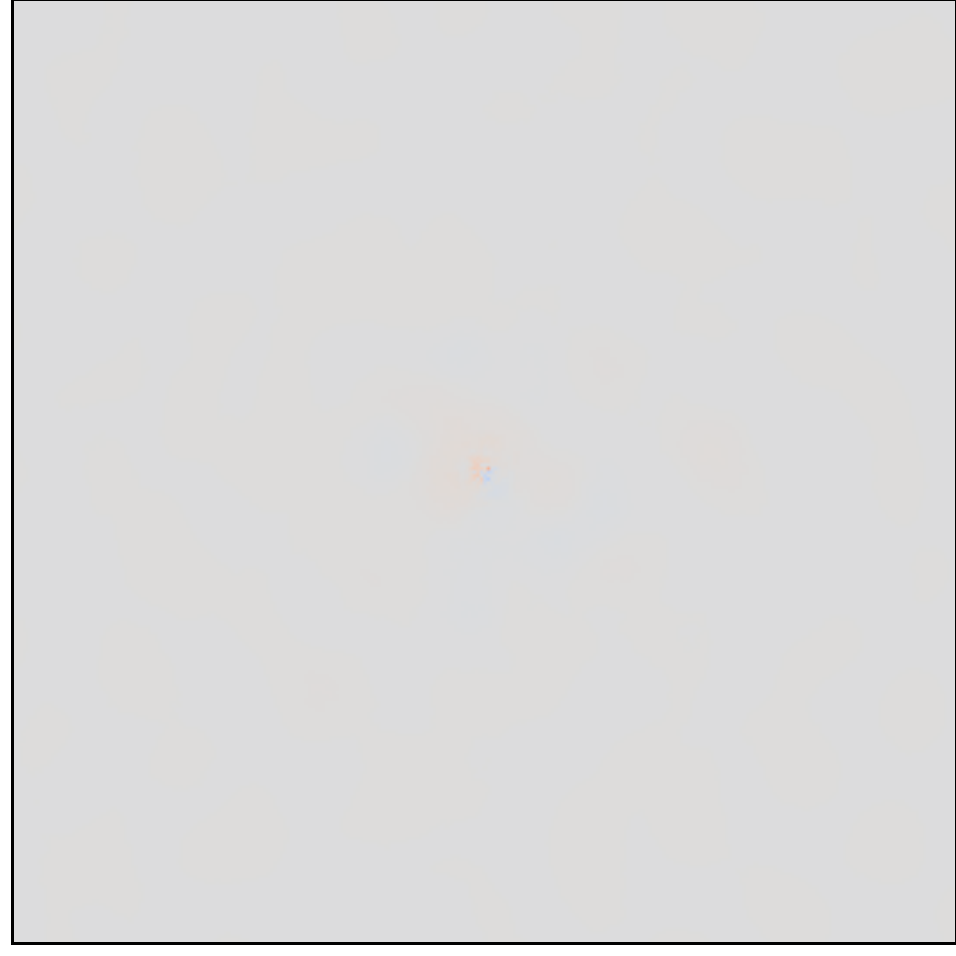} &
        \includegraphics[width=0.049\textwidth]{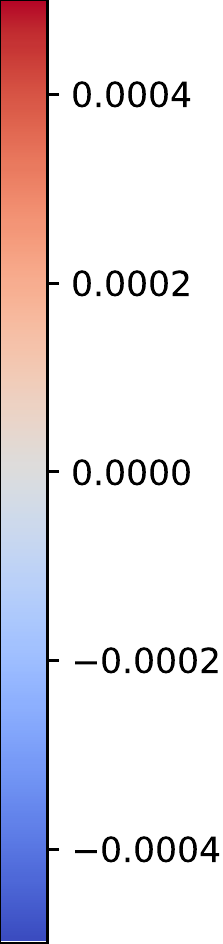} \\
        \includegraphics[width=0.21\textwidth]{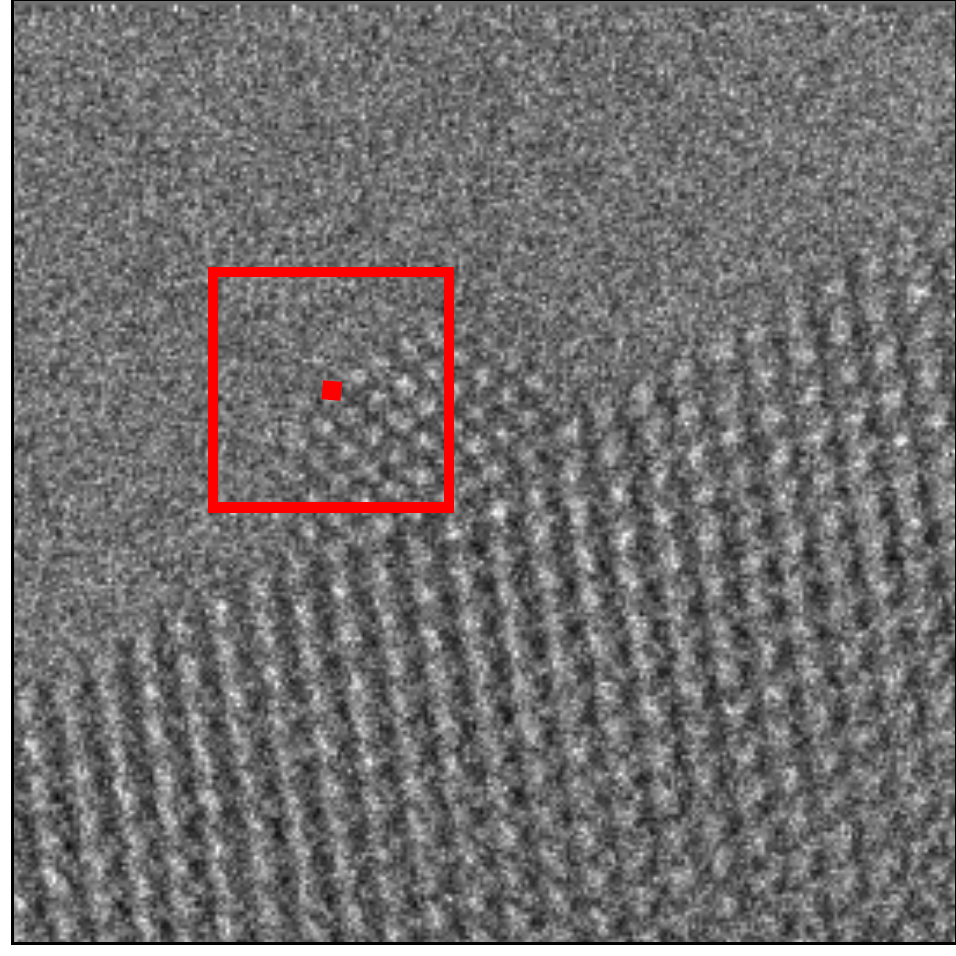} &
        \includegraphics[width=0.21\textwidth]{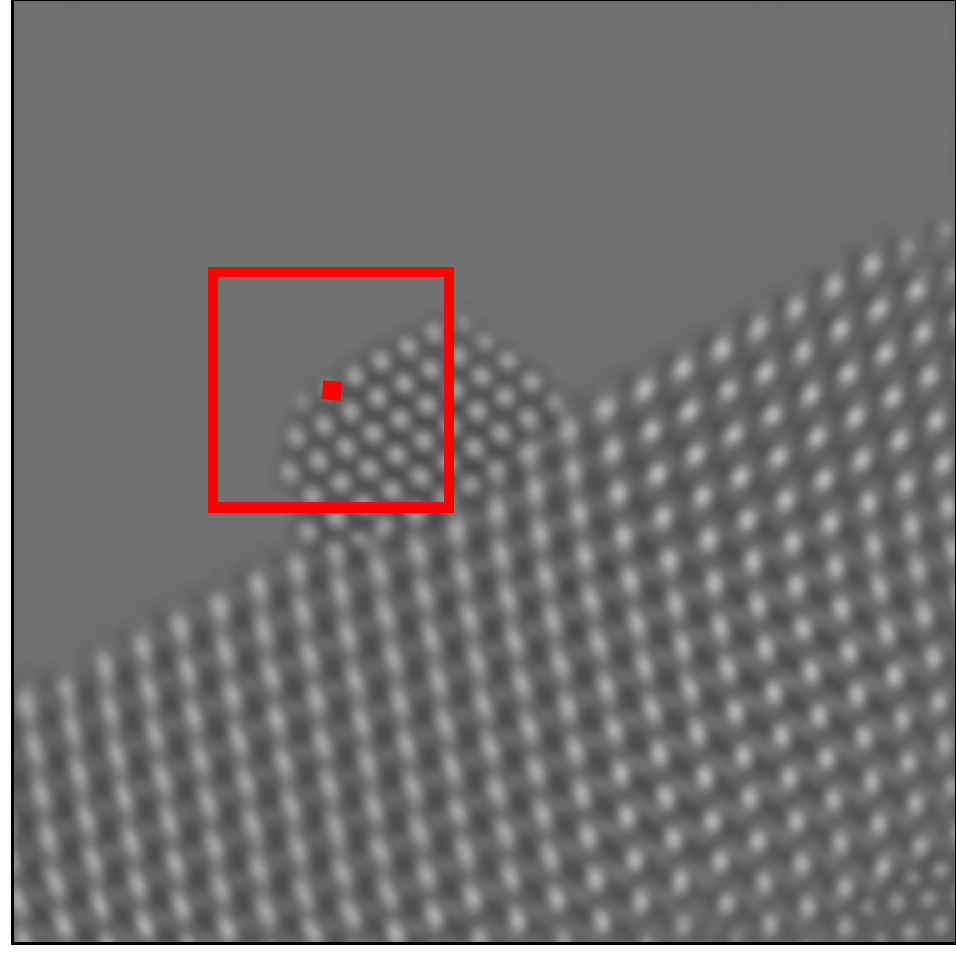} &
        \includegraphics[width=0.21\textwidth]{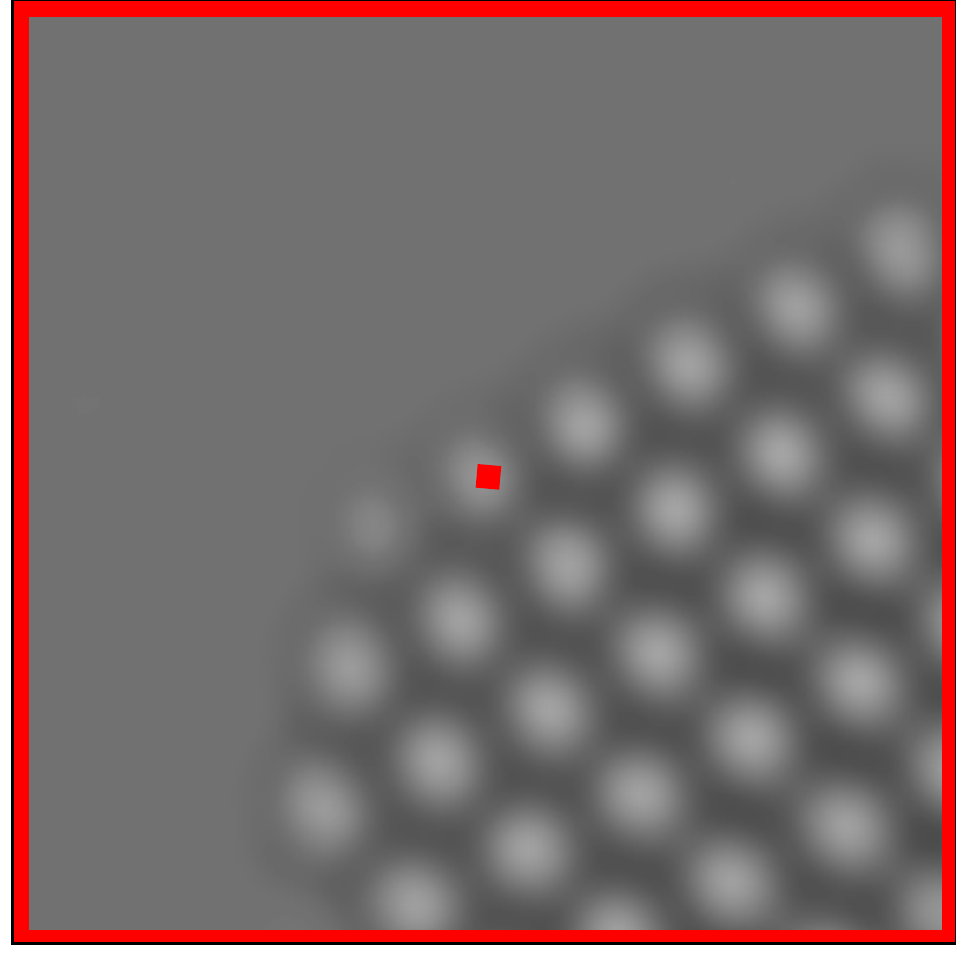} &
        \includegraphics[width=0.21\textwidth]{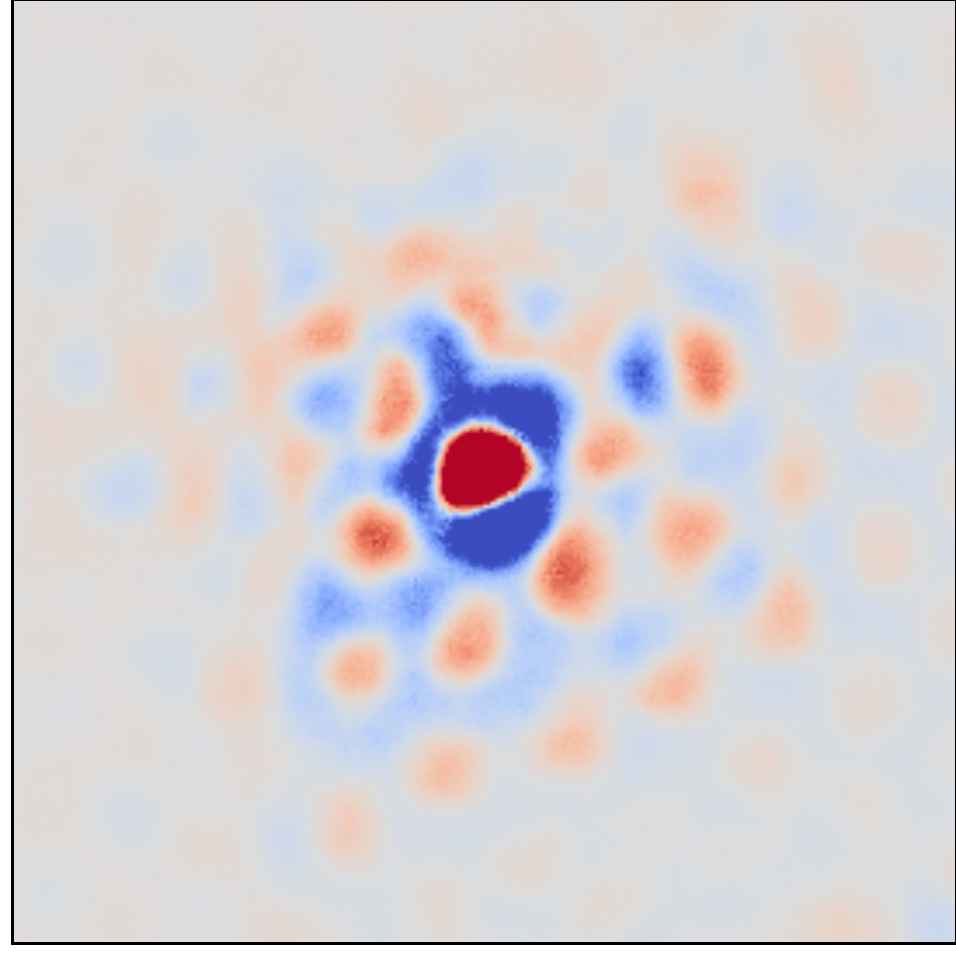} &
        \includegraphics[width=0.049\textwidth]{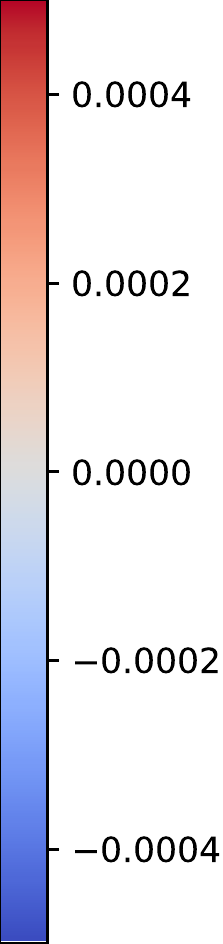}
    \end{tabular}
    \caption{\textbf{Gradient analysis of the learned denoising function on real data.} (a) To compute the red pixel in the denoised image (b), the proposed CNN can use noisy pixels in a neighbourhood of up-to $893 \times 893$, of which a $300 \times 300$ area (red box) (c) is highlighted. The gradient of the denoised pixel with respect to its input indicates what regions in the noisy image have a greater influence on the estimate (according to a first-order Taylor approximation to the denoising map). The gradient (d) weights nearby pixels more heavily, but also has significant magnitude at pixels located on different atoms. This suggests that the CNN combines local and non-local information to estimate the pixel. The colorbar is shared across all the images.}
    \label{fig:jacobian}
\end{figure*}

\def\nsp{\hspace*{-.07in}}

In order to gain some insight into the denoising mechanisms learned by our models, we apply the gradient-based analysis proposed by Ref.~\cite{Mohan2020Robust}. We visualize the linear term in the first-order Taylor decomposition of the denoising map with respect to its input for specific pixels. In more detail, we compute the gradient of a pixel in the denoised image $f(y)_i$ with respect to the input noisy image $y$. This vector (or image) $\nabla_y(f(y)_i)$, makes it possible to visualize the influence of different regions of the noisy image on the denoised pixel $f(y)_i$. This approach is similar to visualization methods proposed in the context of image classification  (e.g.~\cite{simonyan2013deep,montavon2017explaining}). Our analysis reveals that the network learns to  simultaneously exploit local structure as well as non-local periodicities in the data (see Figure~\ref{fig:jacobian}). This demonstrates the remarkable flexibility of data-driven denoising based on deep learning.

\subsection{Likelihood maps}
\label{sec:likelihood}

\begin{figure}[t]
    \centering
    \begin{tabular}{C{0.21\linewidth}@{\hskip 0.08in}C{0.21\linewidth}@{\hskip 0.08in}C{0.21\linewidth}@{\hskip 0.0in}C{0.051\linewidth}@{\hskip 0.08in}C{0.25\linewidth}}
        \footnotesize{(a) Data} & \footnotesize{(b) Denoised image} & \footnotesize{(c) Likelihood map} && \footnotesize{(d) Zoomed} \\
        \includegraphics[width=0.21\textwidth]{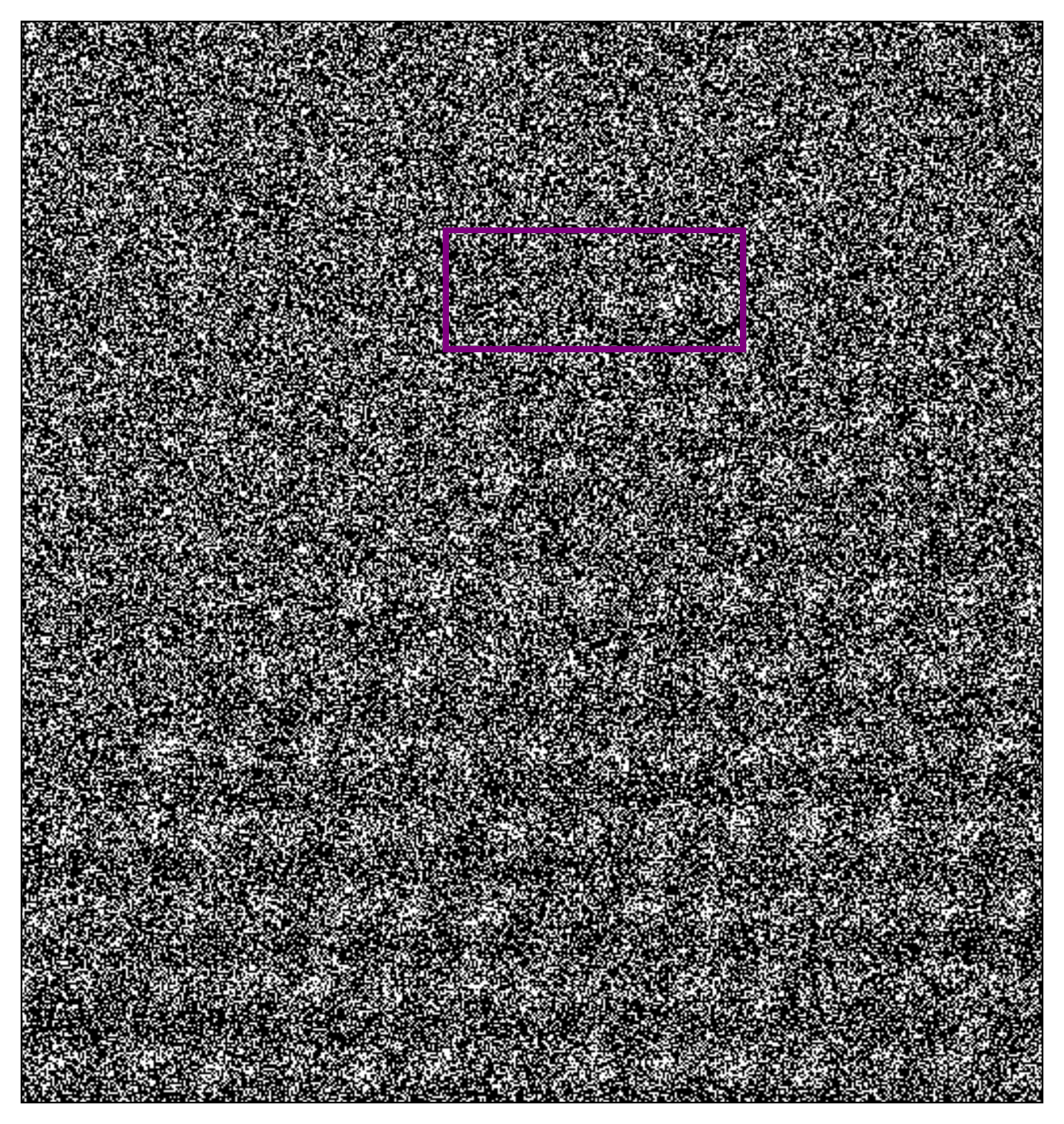} &
        \includegraphics[width=0.21\textwidth]{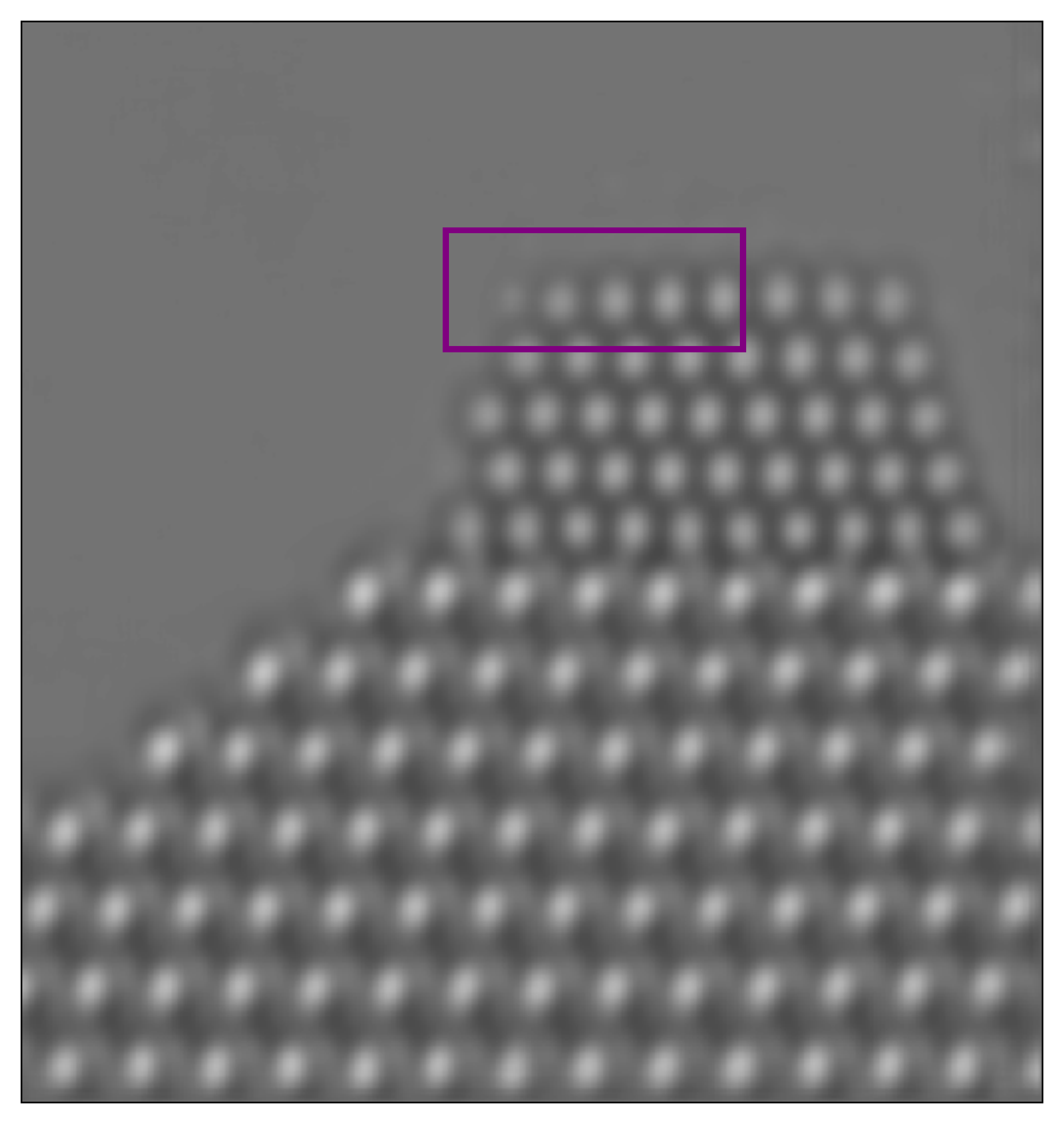} &
        \includegraphics[width=0.21\textwidth]{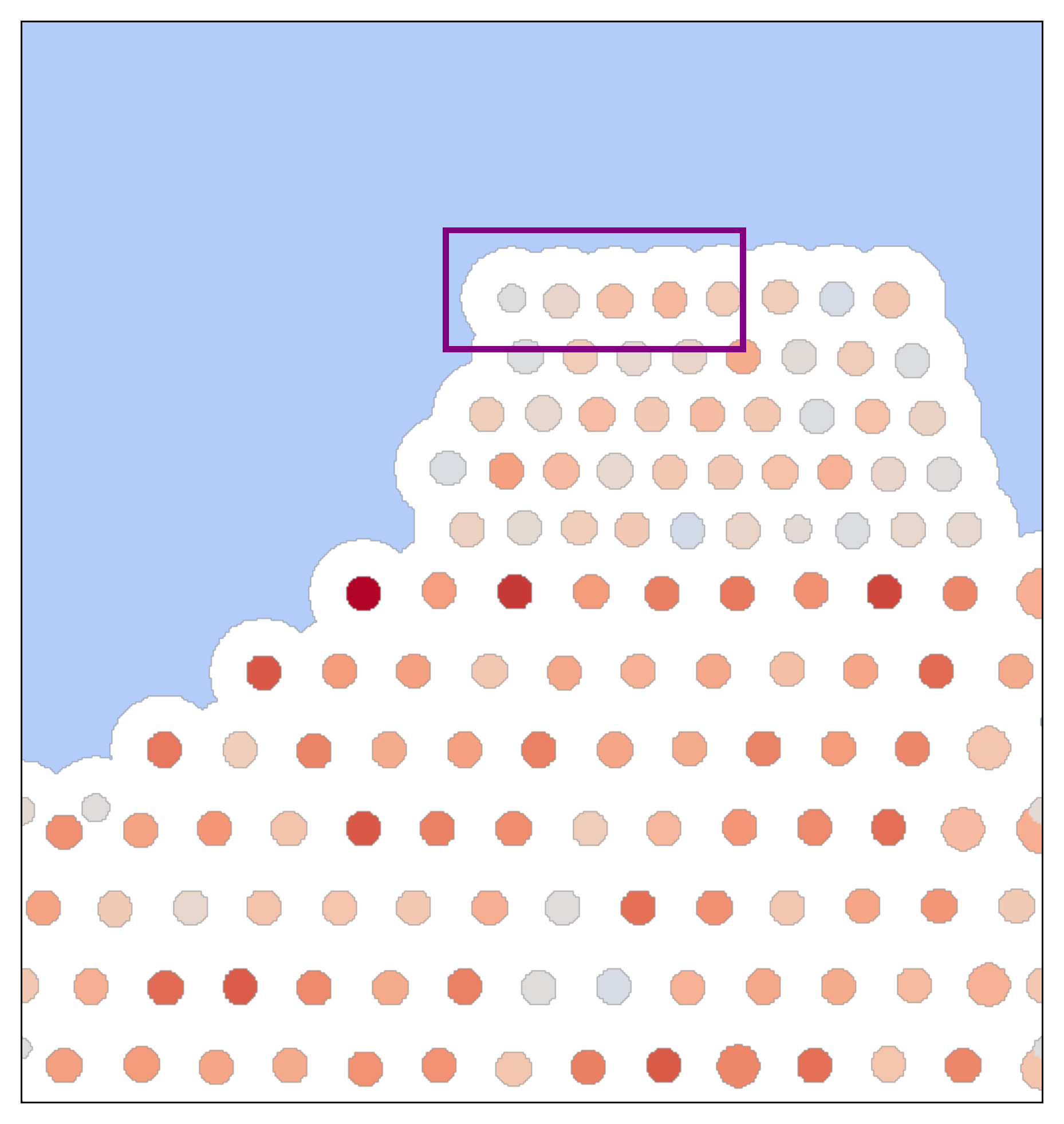} &
        \includegraphics[width=0.051\textwidth]{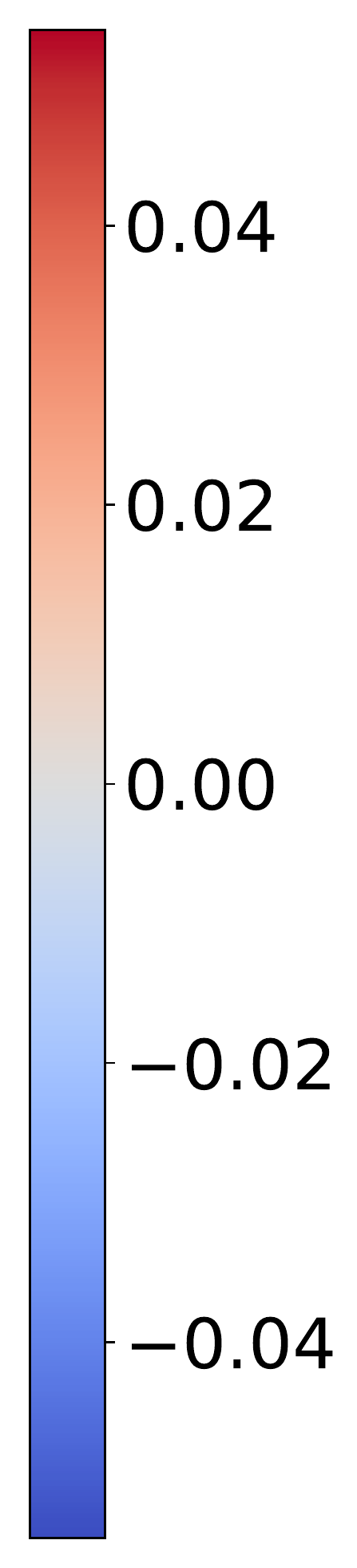} &
        \includegraphics[width=0.25\textwidth]{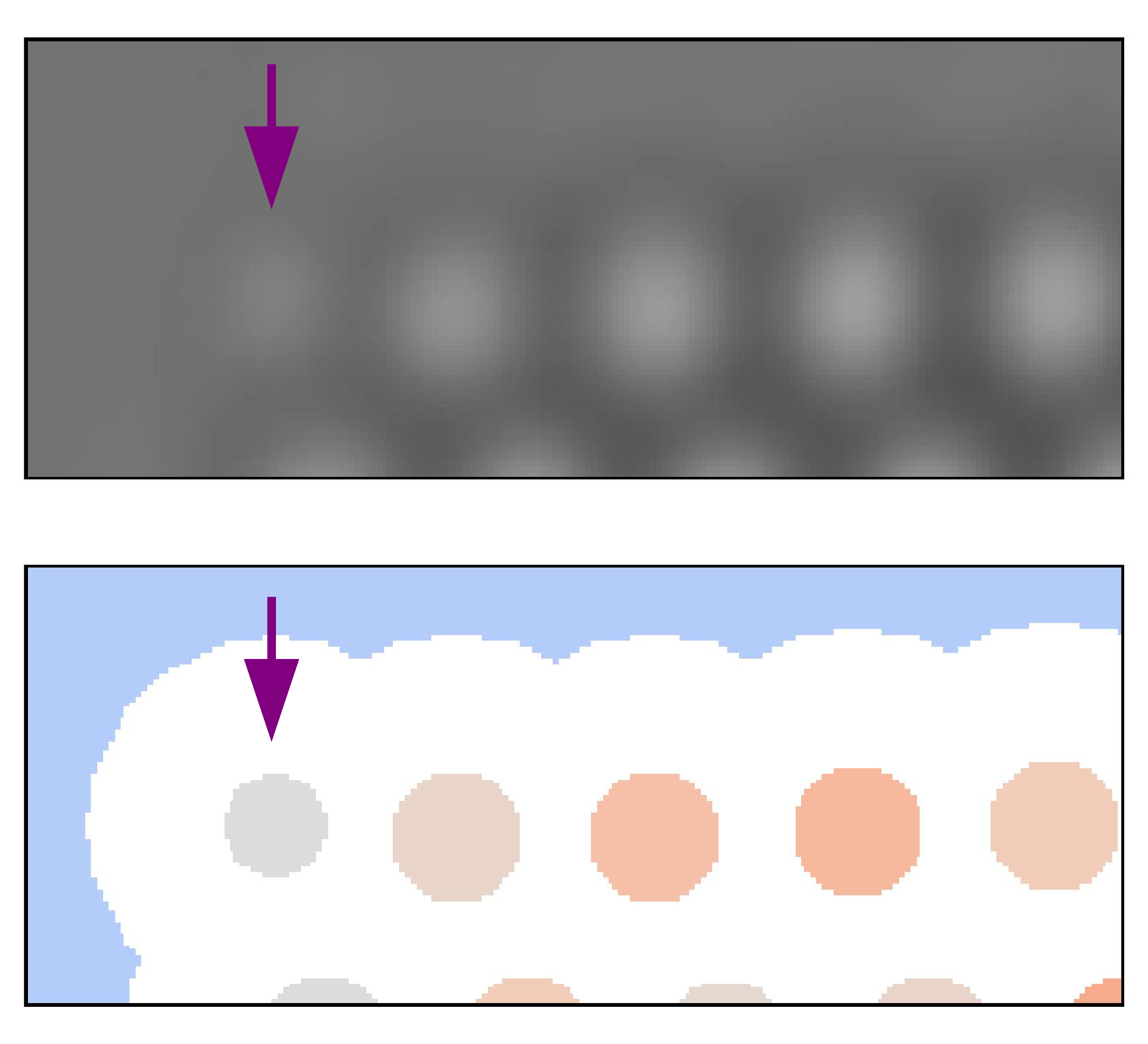}
    \end{tabular}
    \caption{\textbf{Likelihood map.} When the simulated noisy image in (a) is denoised using the proposed framework (b), a spurious atom appears at the left edge of the nanoparticle (see zoomed image (d)). The value of the likelihood map (c) at that location is very low, indicating that the presence of an atom is less consistent with the observed data than its absence. }
    \label{fig:likelihood_sim}
\end{figure}


In most applied domains, the goal of denoising is to uncover image structure of scientific interest. In our case study, this corresponds to the location and intensity of projected columns of atoms in a catalytic nanoparticle that is surrounded by a vacuum. Quantifying to what extent such structure is consistent with the observed measurements is therefore of great interest. We propose to achieve this by computing the likelihood of the data \emph{with respect to meaningful features identified in the denoised image}. The general procedure, and its implementation in the case of our case study, are as follows: 

\begin{enumerate}[leftmargin=*]
    \item Identify a region of interest $\mathcal{R}$. In our case study, this would correspond to an atomic column, located for example via blob detection~\cite{lindeberg2013scale}, or to the vacuum.
    \item 
    Fit a low-dimensional model to the denoised image within the region of interest. The low-dimensional model provides an estimate of the image value $x_i$ at each pixel location $i \in \mathcal{R}$. In our case, we assume that the intensity of each atomic column and the vacuum are constant, so the estimate is obtained by averaging over all denoised pixels in $\mathcal{R}$.
    \item Compute the likelihood of the noisy data in $\mathcal{R}$ with respect to the estimated pixel values. In our case, the noise is approximately independent and individually distributed (iid) Poisson (see Section~\ref{sec:noise_model}), so the likelihood is given by
    \begin{align}
    \mathcal{L}(\mathcal{R}) & := \prod_{i \in \mathcal{R}} p_{x_i}(y_i),   
    \end{align}
    where $y_i$ denotes the noisy value in the $i$th pixel, and $p_{x_i}$ is a Poisson probability mass function (pmf) with rate parameter $x_i$. Note that in the low-dimensional model, which assumes constant intensities, $x_i$ is constant for all pixels in $\mathcal{R}$.
\end{enumerate}

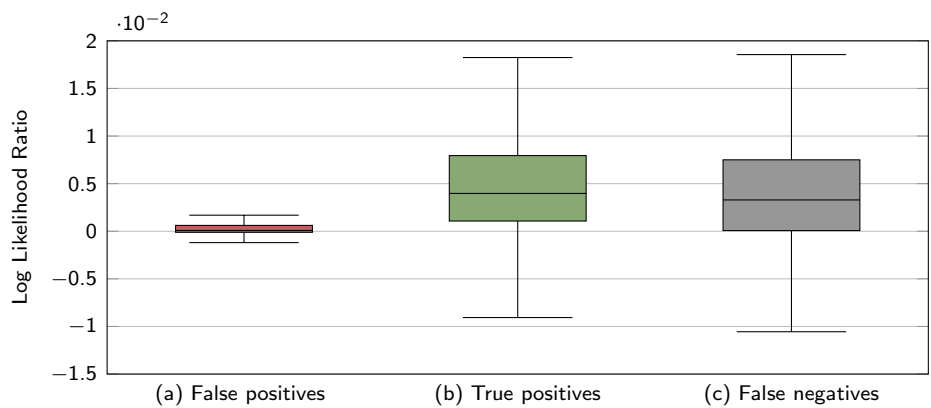
\begin{figure}[t]
    \centering
    \begin{tikzpicture}
        \pgfplotsset{
            ticklabel style={font=\scriptsize\sffamily\sansmath},
            every axis label/.append style={font=\sffamily\sansmath\scriptsize},
            title style={font=\footnotesize\sansmath}
        }
        \begin{axis}[
            height=6cm, width=0.8\linewidth,
            boxplot/draw direction = y,
            ymin=-0.015,ymax=0.02, ymajorgrids=true, ytick distance=0.005,
            boxplot={box extend=0.5},
            xtick={1, 2, 3}, xtick style={draw=none},
            xticklabels={ (a)~False positives  , (b)~True positives, (c)~False negatives}, xticklabel style={align=center},
            ylabel=Log Likelihood Ratio,
        ]
            \addplot+ [draw=black, fill=knred!80, boxplot prepared={lower whisker=-0.00119385, lower quartile=-0.00011067, median=8.39671761e-05, upper quartile=0.00061419, upper whisker=0.00168654}] coordinates {};
            \addplot+ [draw=black, fill=kngreen!80, boxplot prepared={lower whisker=-0.00906631, lower quartile=0.00107292, median=0.00397347, upper quartile=0.00794723, upper whisker=0.01824576}] coordinates {};
            \addplot+ [draw=black, fill=kngrey!80, boxplot prepared={lower whisker=-1.05549512e-02, lower quartile=7.14804899e-05, median=0.00328948, upper quartile=7.50264777e-03, upper whisker=0.0185543}] coordinates {};
        \end{axis}
    \end{tikzpicture}
    \caption{\textbf{Distribution of likelihood ratio.} The figure shows the distribution of log-likelihood ratio of over $25,000$ regions of interest computed from the surface of $1550$ denoised images using the dataset described in Section~\ref{sec:metrics_dataset_eval}. The empirical distribution is visualized as a box plot indicating the median, 25$^{th}$ quartile, 75$^{th}$ quartile, minimum and maximum value of the distribution. The regions containing  spurious atoms (false positives, (a)) have a much lower log-likelihood ratio than the regions containing accurately recovered atoms (true positives, (b)). Regions where existing atoms were not detected (false negatives, (c)) have a higher log-likelihood ratio, comparable to that of the regions with accurately recovered atoms. The occurrence of missing and spurious atoms in denoised images is quite rare: out of the $25,732$ regions of interest, only $2,457$ and $2,368$ were false positives and false negatives respectively.}
     \label{fig:likelihood_box}
\end{figure}

This technique makes it possible to consider different hypotheses about the underlying image structure and compare their agreement with the observed data. In our case study, we evaluate the hypotheses that a detected atomic column is (1) truly there, or (2) an artefact introduced by the denoising procedure. The likelihood under hypothesis (1) is computed as above. The likelihood under hypothesis (2) is computed by setting the estimate $x_i$ equal to the average intensity of the noisy pixels identified as belonging to the vacuum region. To visualize the consistency of the two hypotheses with the measured data, we plot the difference in their log likelihood for each region of interest. This is equivalent to performing a log-likelihood ratio test. We call this visualization a \emph{likelihood map}.

Figures~\ref{fig:experiment-denoised} and~\ref{fig:likelihood_sim} show likelihood maps for the real data and for a simulated example. In the simulated example (Figure~\ref{fig:likelihood_sim}), a spurious atom is detected at the left end of the zoomed region. However, the likelihood map at that location is very low, which indicates that the presence of an atom is not consistent with the observed data at that location. 
Figure~\ref{fig:likelihood_box} shows the distribution of log-likelihood ratio of over $25,000$ regions of interest extracted from the surface of over $1,550$ denoised images obtained from the dataset described in Section~\ref{sec:metrics_dataset_eval}.  
As shown in Figure~\ref{fig:likelihood_box}, the log-likelihood ratio values of spurious atoms (false postives, (a)) are much lower than those of correctly-identified atoms (true positives, (c)). When the network fails to detects atoms (false negatives, (b)), we observe that the log-likelihood ratio in such regions tends to be high. It is worth noting that the occurrence of spurious and missing atoms in the denoised images is quite rare: out of the $25,732$ regions identified, only $2,457$ and $2,368$ regions correspond to spurious and missing atoms respectively.

Visualizing the likelihood is useful to quantify the agreement between the output of deep-learning models and the observed data, but it is important to note that the approach suffers from sampling bias. We focus on regions of the input that have been selected because they resemble structures of scientific interest. The data in those regions are therefore more likely to be in agreement with the presence of such structures, just by the sheer fact that they have been selected. This is a manifestation of the notorious multiple-comparisons problem~\cite{benjamini1995controlling,benjamini2010simultaneous}. Overcoming this issue is an important challenge for future research.



\section{Dataset}
\label{sec:dataset}

The TEM image data used in this work correspond to images from a widely utilized catalytic system, which consist of platinum (Pt) nanoparticles supported on a larger cerium (IV) oxide (CeO$_2$) nanoparticle. This bi-functional catalytic system is ubiquitously used in clean energy conversion and environmental remediation applications, in addition to a broad range of other chemical reactions \cite{montini2016fundamentals, yu2012review, nie2015recent}. From a general point of view, this system can be considered as a model for supported nanoparticle catalysts, since a large number of heterogeneous catalysts are based on  metallic nanoparticles supported over different oxides. Thus, results and conclusions extracted from the current work are relevant to a great number of similar samples in the field of catalysis (e.g., oxide crystals supporting metal nanoparticles).

\subsection{Real Data}
\label{sec:real_dataset}
The real data used to test the proposed SBD framework consist of a series of images of the Pt/CeO$_2$ catalyst. The images were acquired in a N$_2$ gas atmosphere using an aberration-corrected FEI Titan transmission electron microscope (TEM), operated at 300 kV and coupled with a Gatan K2 IS direct electron detector. The detector was operated in electron counting mode with a time resolution of 0.025 sec/frame and an incident electron dose rate of 5,000 e$^-$/{\AA}$^2$/s. The electromagnetic lens system of the microscope was tuned to achieve a highly coherent parallel beam configuration with minimal low-order aberrations (e.g., astigmatism, coma), and a third-order spherical aberration coefficient of approximately -13 $\mu$m. 

\subsection{Simulation Dataset}
The simulated TEM image dataset was generated using the multi-slice TEM image simulation method, as implemented in the Dr. Probe software package~\cite{barthel2018dr} (see Section~\ref{sec:data_simulation} for more details on the simulation process). Images were simulated with 1024 x 1024 pixels and then binned to match the approximate pixel size of the experimentally acquired image series. To equate the intensity range of the simulated images with those acquired experimentally, the intensities of the simulated images were scaled by a factor which equalized the vacuum intensity in a single simulation to the average intensity measured over a large area of the vacuum in a single 0.025 second experimental frame (i.e., 0.45 counts per pixel in the vacuum region).

In the type of phase-contrast TEM imaging performed in this work, multiple electron-optical and specimen parameters can give rise to complex, non-linear modulations of the image contrast. These parameters include the objective lens defocus, the specimen thickness, the orientation of the specimen, and its crystallographic shape/structure. Various combinations of these parameters may cause the contrast of atomic columns in the image to appear as black, white, or an intermediate mixture of the two (see, e.g., Figure \ref{fig:schematic_contrast}). When designing the simulated dataset for the SBD framework, it is necessary to include images simulated under widely varied conditions, in order to cover the breadth of possibilities which may arise during a typical experiment. A skilled microscopist attempts to acquire images under conditions in which the image contrast can be interpreted, which limits the overall size of the parameter space under consideration. However, various instances of defocus, tilt, thickness, and shape/structure inevitably arise. To generate our dataset we systematically varied these parameters to produce a large number of potential combinations (approximately 18,000), as described in Sections~\ref{sec:experimental_parameters} and~\ref{sec:nanoparticle_structures}.

\subsection{Noise model}
\label{sec:noise_model}
\begin{figure*}
    \centering
    \begin{tabular}{c@{\hskip 0.1in}c@{\hskip 0.1in}c@{\hskip 0.1in}c}
    \footnotesize{(a) Mean Image} & \footnotesize{\hspace{0.6cm} (b) Histogram of Intensities} & \footnotesize{(c) Variance \& Mean} \\
    \includegraphics[trim={0 0 18.5cm 0},clip,height=0.28\linewidth]{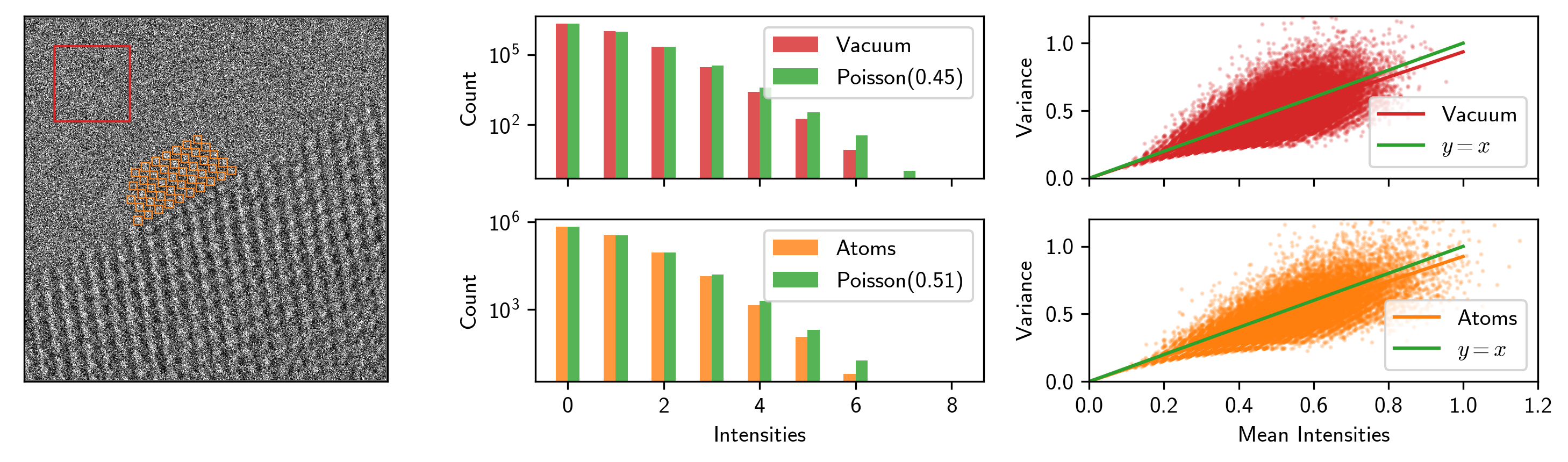} &
    \includegraphics[trim={7cm 0 9cm 0},clip,height=0.28\linewidth]{images/mean_variance/mean_var.png} &
    \includegraphics[trim={16cm 0 0 0},clip,height=0.28\linewidth]{images/mean_variance/mean_var.png}\\
    \end{tabular}
    \caption{\textbf{Analysis of the noise in the real data}. The analysis shows that the noise is approximately iid Poisson. (a) Pixel-wise mean over $40$ frames of the real data described in Section~\ref{sec:dataset}. (b) Histogram of noisy pixel intensities from highlighted regions in the image compared to a simulated Poisson distribution. (c) The plot of empirical mean and standard deviation of pixels approximately follows a line with unit slope, as expected from iid Poisson samples (the spread is due to averaging over only 40 frames).}
    \label{fig:noise-statistics}
\end{figure*}


The TEM images were acquired on a direct electron detector operating in electron counting mode. In such conditions, the electron dose rate per pixel is sufficiently low enough that individual electron arrivals can be detected and registered. It is well known that the statistical fluctuations of such counting processes for discrete events are governed by shot noise, which can be modeled with a Poisson distribution~\cite{barford1967experimental}. We expect that other sources of noise, including fixed pattern noise, dark noise, and thermal noise are minimal after applying a gain correction and a dark reference to the raw image, and by cooling the detector to -20 $^{\circ}$C, respectively.  

We empirically verified that the noise follows Poisson statistics through the analysis shown in Figure~\ref{fig:noise-statistics}. Our real dataset, described in Section~\ref{sec:real_dataset} consists of $40$ noisy frames acquired sequentially across time in 0.025 sec intervals. Figure \ref{fig:noise-statistics}(a) shows the mean image over all $40$ frames. The region of the image containing no material (red box) corresponds to the vacuum, where the electron beam intensity is uniformly illuminating the detector. Fluctuations of the intensity in this region therefore purely arise from the noise. We validate that the histogram of these pixels aggregated over the identified spatial region closely follows a Poisson distribution (Figure \ref{fig:noise-statistics}(b)). Pixels aggregated over spatial domains corresponding to the Pt atomic columns (orange boxes) show similar behavior. Further, if the distribution is indeed Poisson, the mean and variance of the noisy pixels should be approximately the same. The empirical mean and variance, computed by averaging over the 40 time frames at every spatial location, follows a linear trend, thus further confirming that the noise distribution has Poisson properties (Figure \ref{fig:noise-statistics}(c)). The spread in the scatter plot is due to the limited number of time frames over which we average.

\section{Experiments and Results}
\label{sec:experiments}

In this section, we evaluate the performance of our proposed methodology and show that we outperform other methods by a large margin (more than $12$ dB in PSNR on held-out simulated data). We also perform a thorough analysis of the generalization capability of our models, demonstrating that the CNNs are robust to variations in imaging parameters and in underlying signal structure. Furthermore, we demonstrate that standard performance metrics for photographs, such as peak signal-to-noise ratio (PSNR) and SSIM~\cite{wang2004image}, may fail to produce a scientifically-meaningful evaluation of the denoising results, and we propose a few alternative metrics to remedy this. Finally, we show that our approach achieves effective denoising of real experimental data. 

We use CNNs with the proposed UNet architecture with $128$ base channels and $6$ scales in all of our experiments (see Sections~\ref{sec:non_local} and \ref{sec:architectures} for more details). The networks were trained on $400 \times 400$ patches extracted from the training images and augmented with horizontal flipping, vertical flipping, random rotations between $-45^{\circ}$ and $+45^{\circ}$, and random resizing by a factor of $0.75$-$0.82$. The models were trained using the Adam optimizer~\cite{kingma2014adam}, with a default starting learning rate of $10^{-3}$, which was reduced by a factor of $2$ every time the validation PSNR plateaued. Training was terminated via early stopping based on validation PSNR. The details of training, validation and test data for each experiment are provided in the corresponding section. Since the models are trained on $400 \times 400$ patches, when applying them to larger images we divide the images into overlapping $400 \times 400$ patches, denoise them, and then combine them via averaging.

\subsection{Generalization to unseen structures and acquisition conditions}
\label{sec:generalization}

\begin{figure*}
    \centering
    \begin{tabular}{c@{\hskip 0.15in}c@{\hskip 0.15in}c@{\hskip 0.15in}c}
    \includegraphics[height=0.27\linewidth]{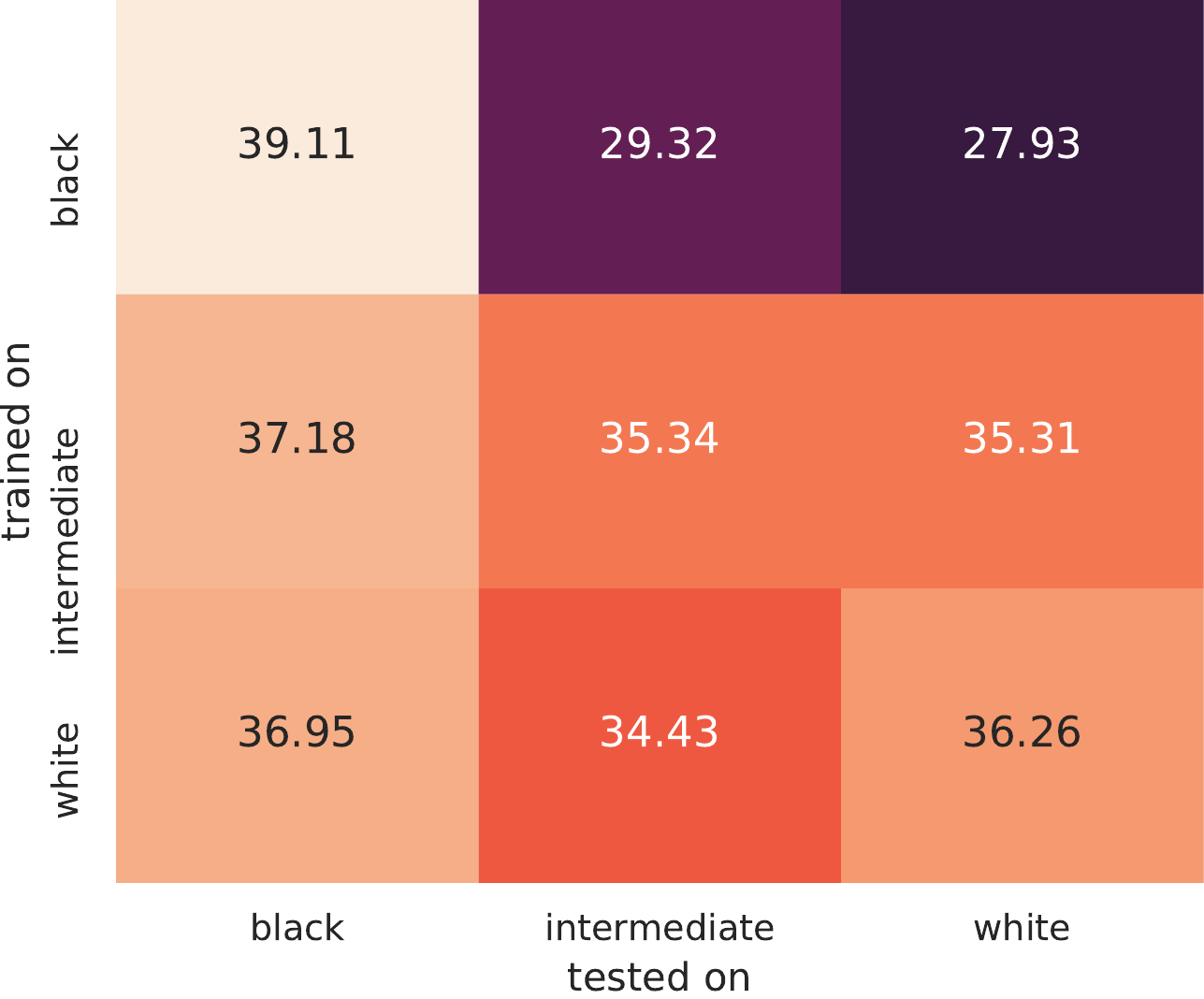} &
    \includegraphics[height=0.27\linewidth]{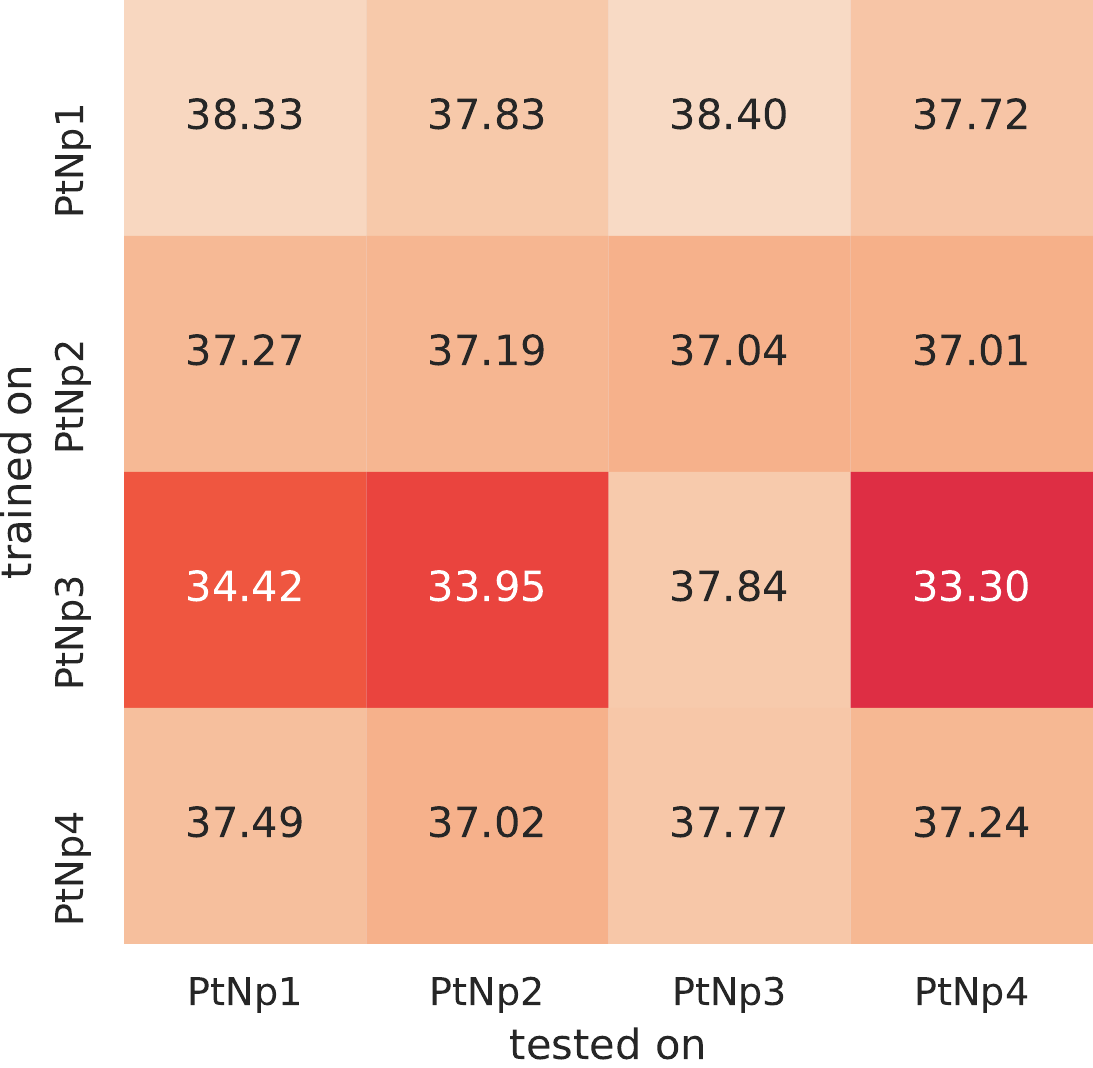} &
    \includegraphics[height=0.27\linewidth]{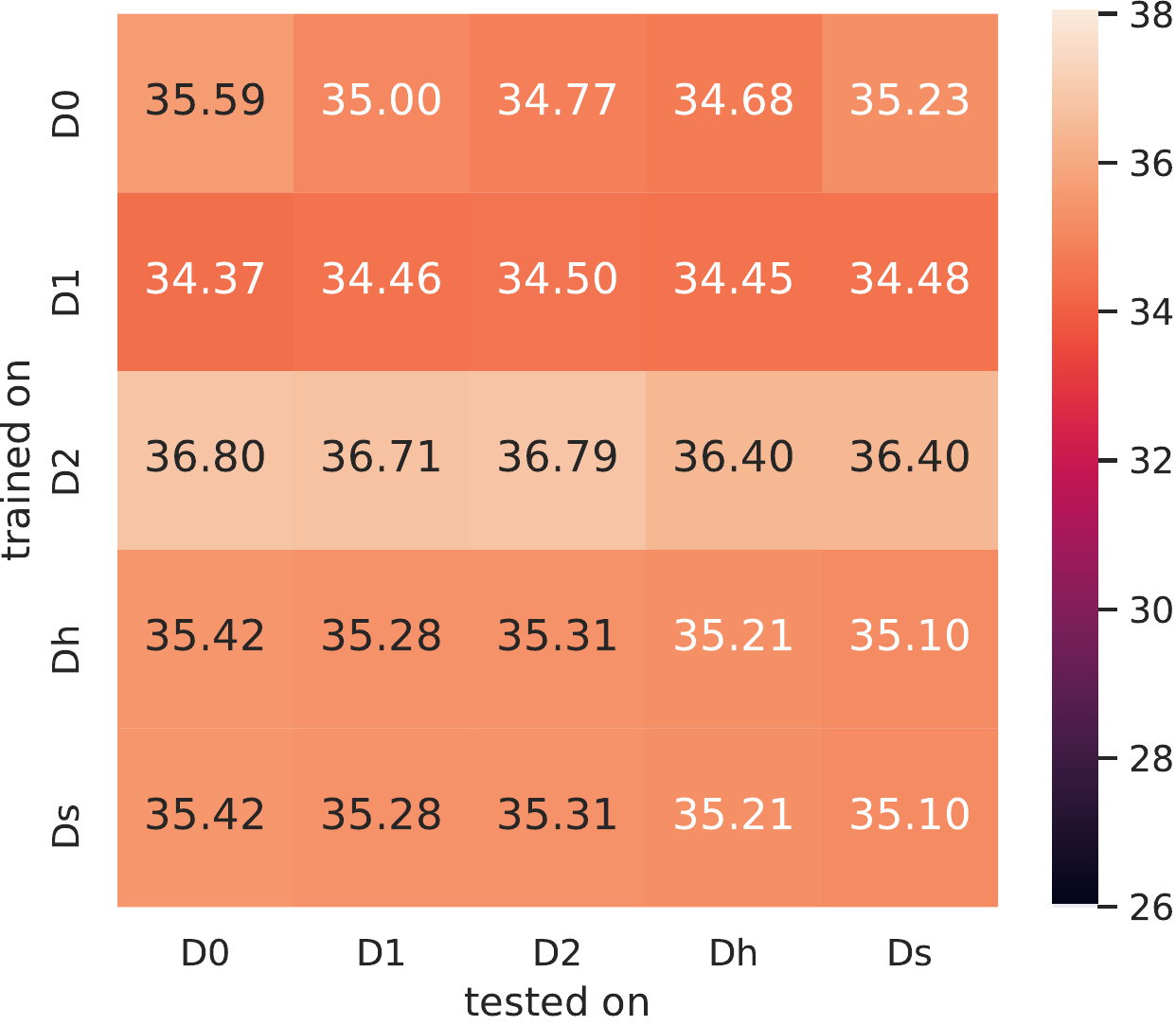} \\
    \footnotesize{(a) Contrasts} & \footnotesize{(b) Structures} & \footnotesize{(c) Defects} \\
    \end{tabular}
    \caption{\textbf{Generalization across different imaging parameters and signal structures}. In order to study the generalization ability of the proposed method we divided the simulated dataset described in Section~\ref{sec:dataset} into subsets, based on the atomic column contrast, the structure/size of the supported Pt nanoparticle, and the defects of the Pt surface structure. The tables show the test PSNR for networks trained and tested on different combinations of the subsets.
    (a) Networks trained on white and intermediate contrast generalize well to all other contrasts. The network trained on black contrast does not generalize as well. 
    (b) Networks trained on a type of nanoparticle structure generalize well to all other types. (c) Networks trained on one type of defect or no defects generalize well to different types.  Figures~\ref{fig:schematic_contrast}, \ref{fig:schematic_defect}, \ref{fig:schematic_structure} and \ref{fig:schematic_thickness_defocus} show the effect of these parameters on the images.}
    \label{fig:generalization}
\end{figure*}

 In order to study the generalization ability of the proposed approach across different imaging parameters and signal structures we divided the simulated dataset described in Section~\ref{sec:dataset} into different subsets. These subsets were classified based on (1) the character of the atomic column contrast, (2) the structure/size of the supported Pt nanoparticle, and (3) the defects of the Pt surface structure. 
The contrast was classified into three divisions,
black, intermediate, or white contrast, by a domain expert (see Figure \ref{fig:schematic_contrast} in the supplementary material). 
The nanoparticle structure was classified into four categories, “PtNp1” through “PtNp4”. PtNp1 and PtNp2 correspond to supported Pt nanoparticles of size 2 nm, which differ in the presence or absence of an atomic column located at the interface between the Pt and the CeO$_2$ support. PtNp3 corresponds to a Pt nanoparticle ~1 nm in size. PtNp4 corresponds to a Pt nanoparticle ~3 nm in size. Finally, the defects were divided into five categories: 
“D0”, “D1”, “D2”, “Dh”, and “Ds” in accordance with the atomic-scale structural models presented in \ref{sec:data_simulation} and in particular in Figure \ref{fig:schematic_defect}. D0 is the initial structure, D1/D2 a structure in which 1/2 atomic columns have been removed respectively, Dh a structure in which a column has been reduced to half its original occupancy, and Ds a structure in which a column has been reduced to a single atom. The generalization ability of the proposed CNN was evaluated by systematically training on each of the subsets and testing on the rest. The number of images in each subset was fixed to be equal in order to ensure a fair comparison. 

The performance of SBD is robust to variations in imaging parameters and in the underlying signal structure, as shown in Figure~\ref{fig:generalization}. We only observe a significant decrease in performance when the network is trained on black-contrast images and tested on other contrasts (interestingly the network generalizes well from white and intermediate contrasts to black contrasts).  

\begin{table}[]
\centering

    \begin{tabular}{lrr}
        \toprule
        \textsc{Methods} & PSNR & SSIM \\
        \midrule
        Raw & 3.56 $\pm$ 0.03 & 0.00 $\pm$ 0.00 \\
        Low Pass Filter~\cite{nellist1998accurate} & 21.59 $\pm$ 0.07 & 0.44 $\pm$ 0.03 \\
        Adaptive Wiener Filter \cite{lim1990two} & 22.42 $\pm$ 1.08 & 0.63 $\pm$ 0.02 \\
        VST + NLM \cite{buades2005non} & 26.55 $\pm$ 0.16 & 0.73 $\pm$ 0.01 \\
        VST + BM3D \cite{makitalo2012optimal} & 25.27 $\pm$ 0.15 & 0.80 $\pm$ 0.01 \\
        PURE-LET \cite{luisier2010image} & 28.36 $\pm$ 0.88 & 0.93 $\pm$ 0.01 \\
        SBD + DnCNN \cite{zhang2017beyond} & 30.47 $\pm$ 0.64 & 0.93 $\pm$ 0.01 \\
        SBD + Small UNet \cite{zhang2018dynamically} & 30.87 $\pm$ 0.56 & 0.93 $\pm$ 0.01 \\
        \textbf{SBD + Proposed Architecture} & \textbf{42.87} $\pm$ 1.45 & \textbf{0.99} $\pm$ 0.01 \\
        \bottomrule
    \end{tabular}

    \caption{\textbf{Results on simulated test data.} Mean PSNR and SSIM ($\pm$ standard deviation) of different denoising methods on the held-out simulated test set described in Section~\ref{sec:sbd_comparison}. SBD approaches achieve the best results. SBD combined with the proposed architecture outperforms all other techniques by about 12 dB. The performance of SBD applied to additional architectures is reported in Table~\ref{tab:network_comparison}. 
    }
    \label{tab:psnr_table}
\end{table}

\begin{figure*}[!ht]
    \centering
    \begin{tabular}{c@{\hskip 0.01in}c@{\hskip 0.01in}c@{\hskip 0.01in}c@{\hskip 0.01in}c}
    \footnotesize{Noisy} & \footnotesize{WF} & \footnotesize{LPF} & \footnotesize{VST+NLM} & \footnotesize{VST+BM3D} \\
    \includegraphics[width=1.05in]{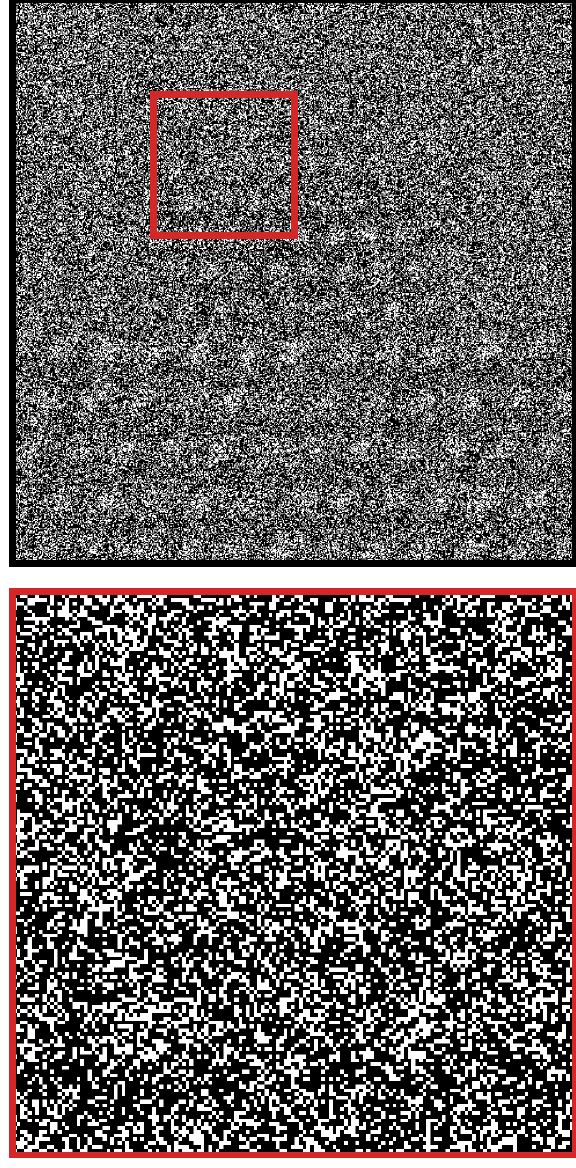}&
    \includegraphics[width=1.05in]{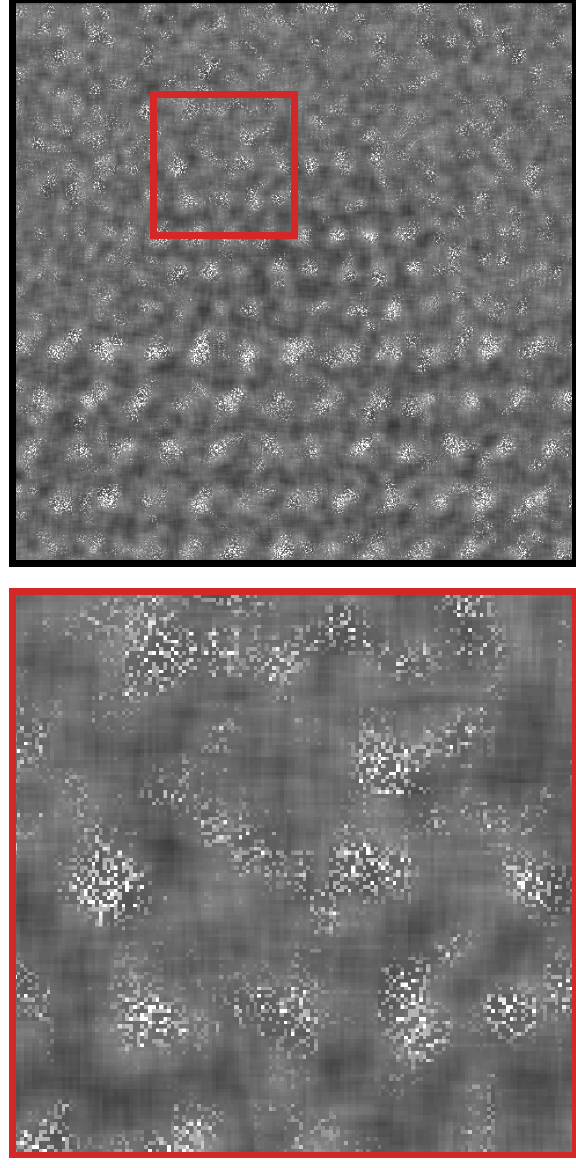}&
    \includegraphics[width=1.05in]{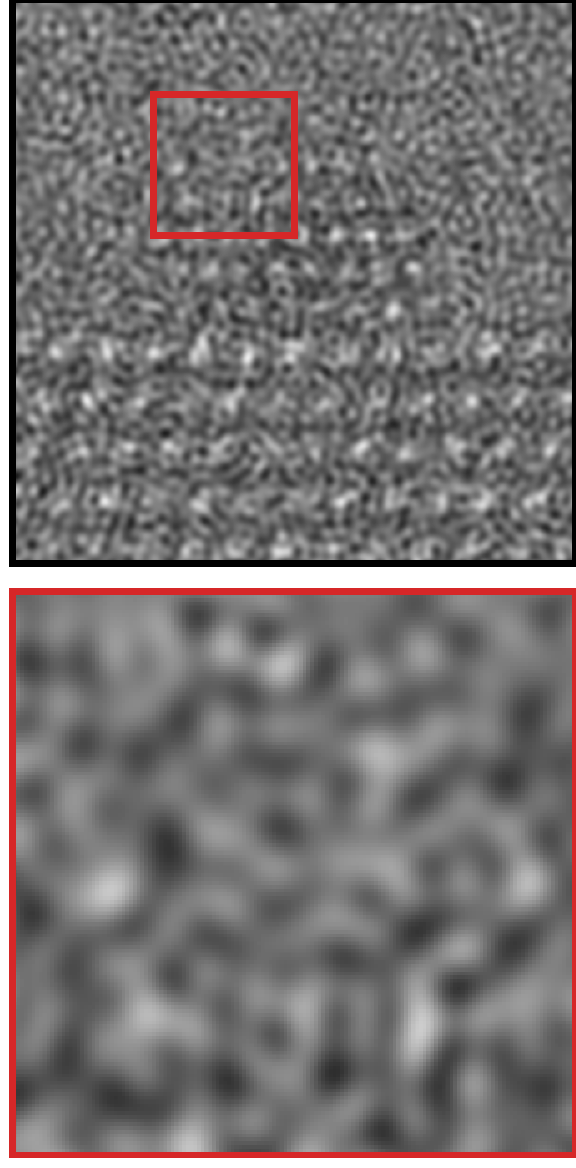}&
    \includegraphics[width=1.05in]{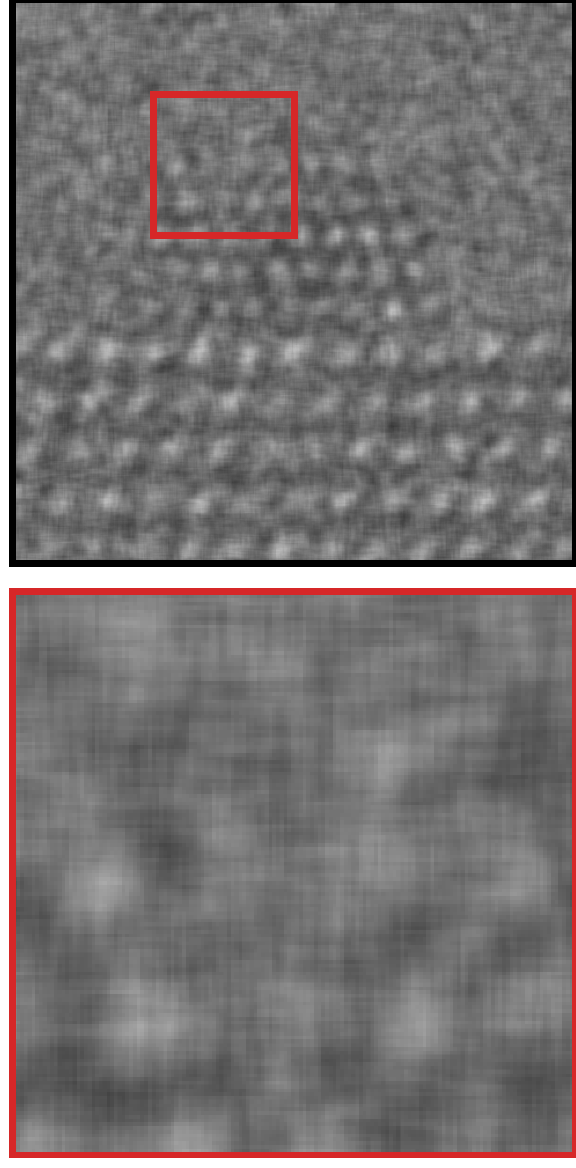}&
    \includegraphics[width=1.05in]{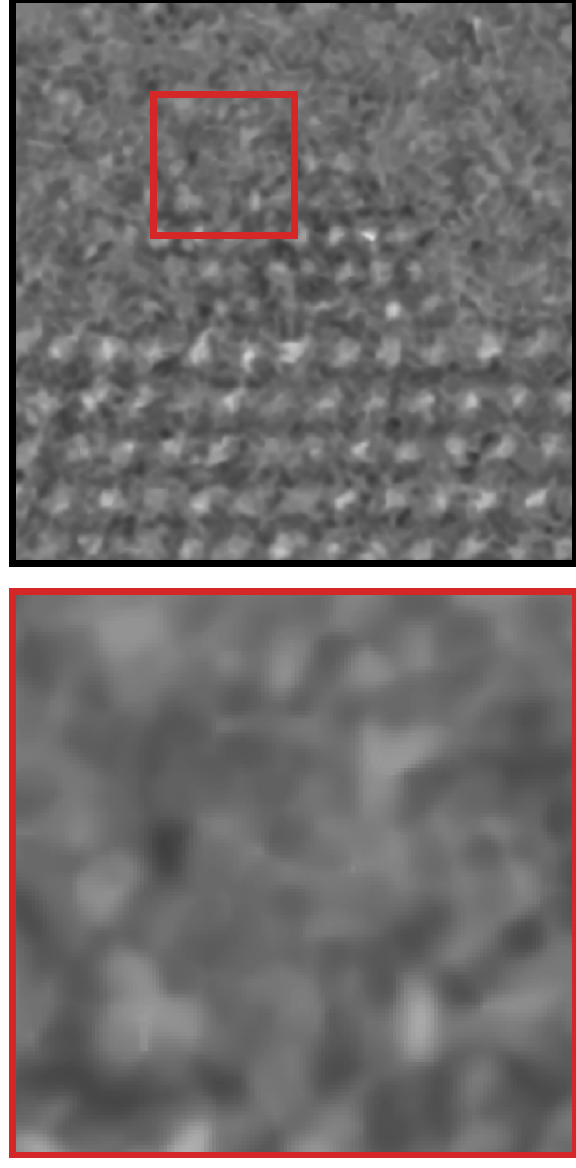}\\

    \footnotesize{PURE-LET} & \footnotesize{SBD+DnCNN} & \footnotesize{SBD+Small UNet} & \footnotesize{Ours} & \footnotesize{Ground Truth} \\
    \includegraphics[width=1.05in]{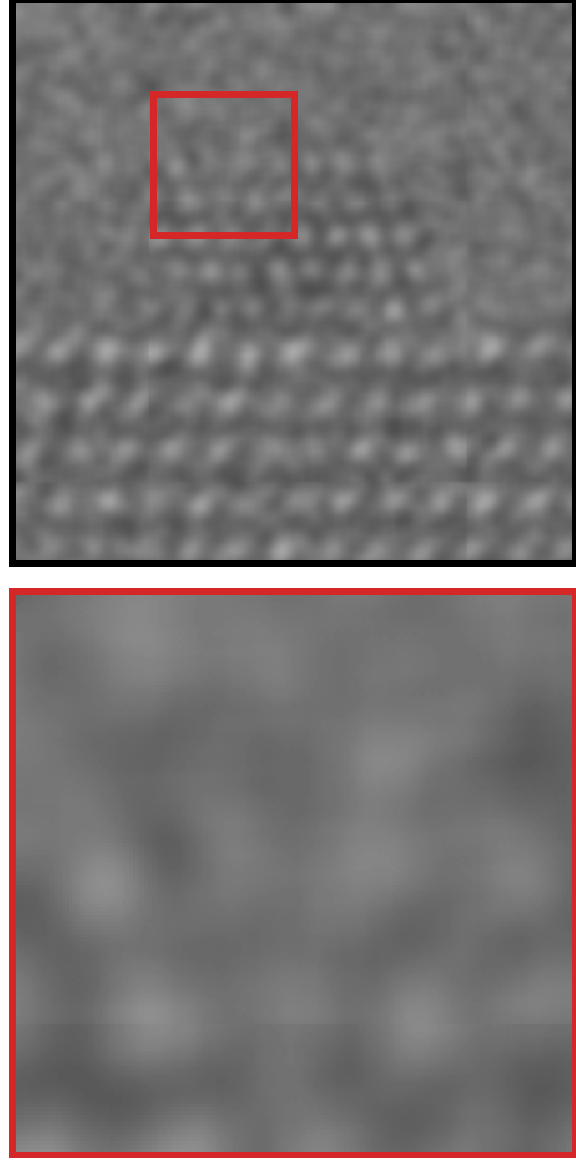}& \includegraphics[width=1.05in]{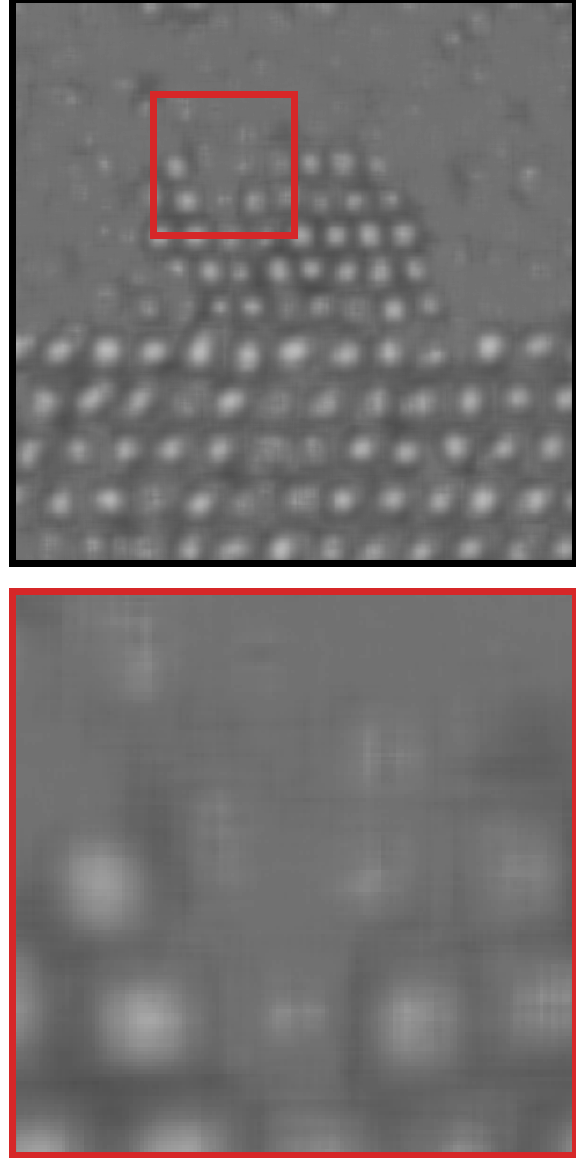}&
    \includegraphics[width=1.05in]{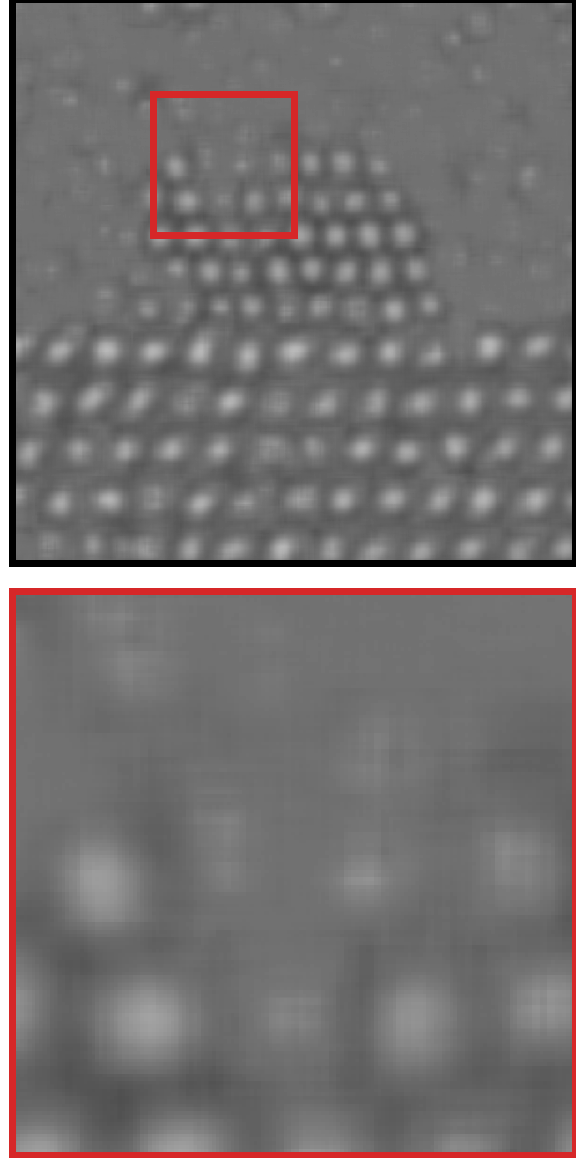}&
    \includegraphics[width=1.05in]{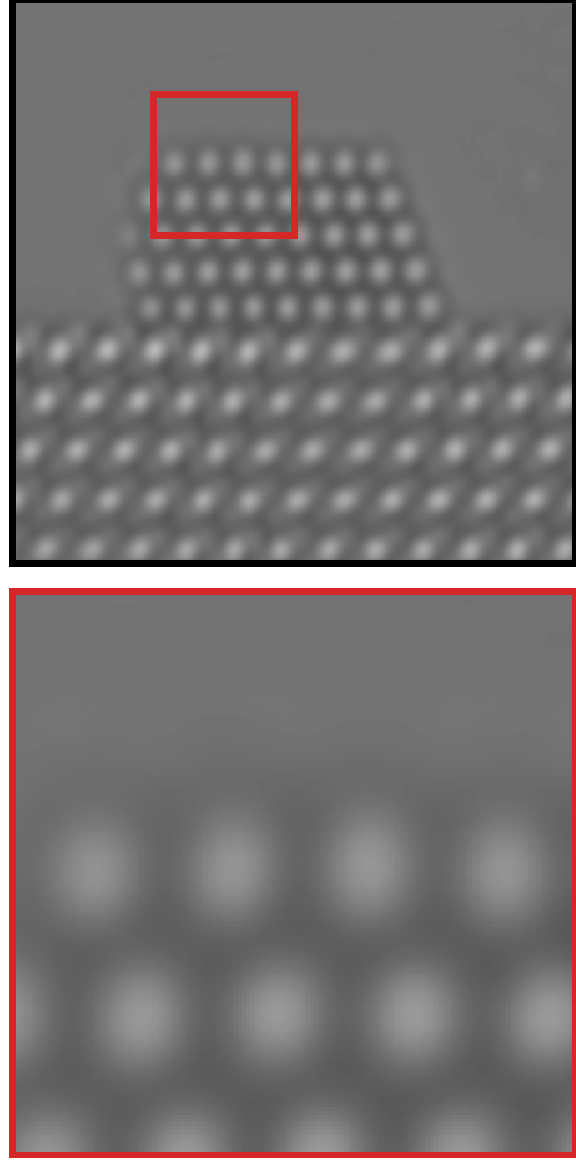}&
    \includegraphics[width=1.05in]{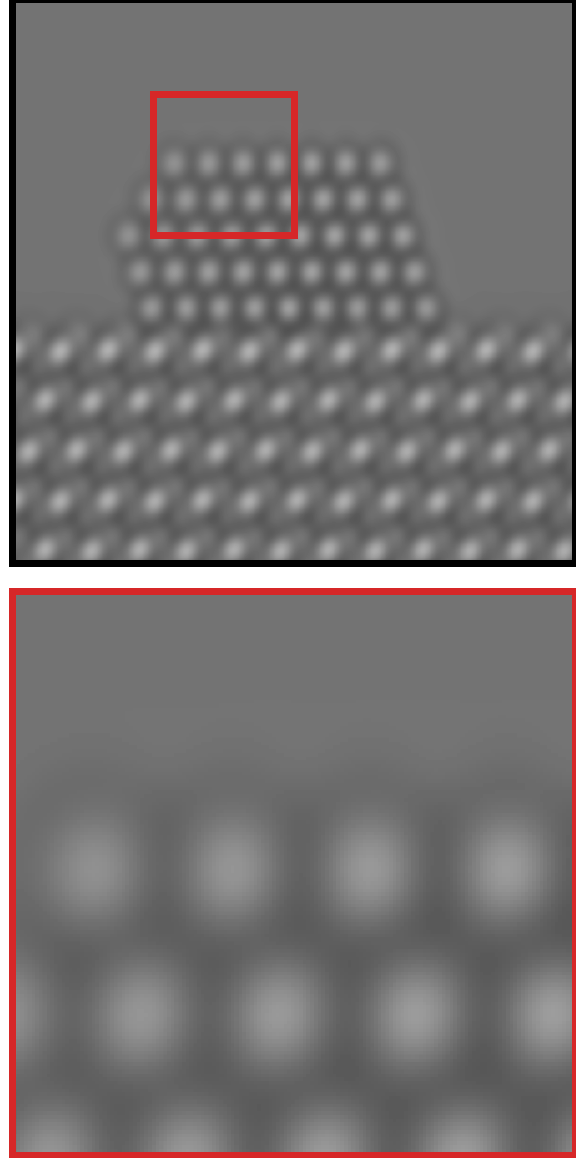}\\
    \end{tabular}
    \caption{\textbf{Denoising results for simulated data}. Comparison of SBD and the baseline methods described in Sections~\ref{sec:sbd_comparison} and~\ref{sec:architectures}. The second row zooms in on the region in red box. Our proposed approach produces images of much higher quality than the other approaches, and is able to accurately recover the atomic structure of the nanoparticle. For example, the vacuum region in images denoised by several of the baselines contain visible artefacts, including missing atoms. See Figure~\ref{fig:simulation-denoised-2} in the supplementary material for an additional example. 
    }
    \label{fig:simulation-denoised}
\end{figure*}

\subsection{Comparison of SBD with other methods}
\label{sec:sbd_comparison}

The imaging parameters of the real data, described in Section~\ref{sec:dataset}, are well described by the white contrast category defined in Section~\ref{sec:experimental_parameters}. We therefore used the subset of simulated dataset corresponding to this contrast ($5583$ images) to compare our proposed methodology to other models. $90\%$ of the data were used for training. The remaining $559$ images were evenly split into validation and test sets. We compare our proposed UNet architecture (see Section~\ref{sec:architectures}) with two state-of-the-art architectures for photographic-image denoising~\cite{zhang2017beyond,zhang2018dynamically} (see Sections~\ref{sec:dncnn} and~\ref{sec:Small UNet}), and with several classical denoising methods: 


\begin{itemize}
    
\item  \textbf{Low-pass Filter (LPF)}: We apply a standard low-pass filter~\cite{nellist1998accurate} designed by a domain expert. The cut-off frequency was determined based on inspection of the data in the Fourier domain. 

\item \textbf{Adaptive Wiener Filter}: We apply an adaptive low-pass Wiener filter \cite{lim1990two} to account for variations in local image statistics. The mean and variance around each pixel are estimated from a local neighborhood. 
We selected a neighborhood with radius $13$ pixels based on expert evaluation of the denoised images in the validation set. 

\item \textbf{VST + BM3D and VST + Non-Local Means (NLM)}: BM3D \cite{makitalo2012optimal} and NLM \cite{buades2005non} are popular denoising techniques for natural images with additive Gaussian noise, which can be adapted to Poisson noise by applying a nonlinear variance-stabilizing transformation (VST)~\cite{zhang2019poisson}. More specifically, we use the Anscombe transformation proposed in Ref.~\cite{makitalo2012optimal}.

\item \textbf{PURE-LET}: PURE-LET \cite{luisier2010image}
 is a transform-domain thresholding algorithm adapted to mixed Poisson–Gaussian noise. The method requires the input image to have dimensions of the form $(2^k, 2^k)$. To apply this method on our TEM images, we extracted $128 \times 128$ overlapping patches, denoised them and combined them via averaging.
 
\end{itemize}
 
For all methods, hyperparameters were chosen based on the validation data. Performance was measured in terms of SSIM~\cite{wang2004image} and peak signal-to-noise ratio (PSNR). 

The results demonstrate that SBD is an effective denoising methodology for TEM data. Our proposed CNN outperforms all other methods by a margin of 12 dB in PSNR on the simulated test data, as shown in Table~\ref{tab:psnr_table}, and Figures~\ref{fig:simulation-denoised} and~\ref{fig:simulation-denoised-2}. 
SBD recovers the overall shape of the nanoparticle, the interface between the nanoparticle and the support, and the different periodic patterns of the CeO$_2$ support and Pt nanoparticle. Contrast features, such as subtle patterns of bright, intermediate and dark features associated with the atomic structure of the CeO$_2$ crystal, are well reproduced in the images denoised via SBD, but are mostly absent from the results of the baseline approaches. 

\subsection{Beyond PSNR: Towards scientifically-meaningful evaluation metrics
}
\label{sec:metrics}

\def\nsp{\hspace*{0in}}
\begin{figure}[!t]
\def\f1ht{1.17in}
\centering 
\begin{tabular}{
>{\centering\arraybackslash}m{0.1\linewidth}>{\centering\arraybackslash}m{0.18\linewidth}>{\centering\arraybackslash}m{0.18\linewidth}>{\centering\arraybackslash}m{0.18\linewidth}>{\centering\arraybackslash}m{0.18\linewidth}}
&\footnotesize{(a) Reference} \vspace{0.1cm} & \footnotesize{(b) All atoms preserved} \vspace{0.1cm} & \footnotesize{(c) One extra atom} \vspace{0.1cm} & \footnotesize{(d) One missing atom} \vspace{0.1cm} \\
  \footnotesize{Images} & \nsp\includegraphics[height=\f1ht]{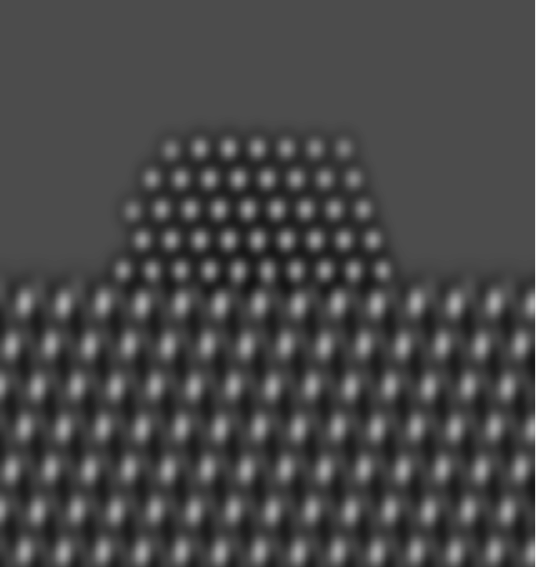}\nsp &
  \nsp\includegraphics[height=\f1ht]{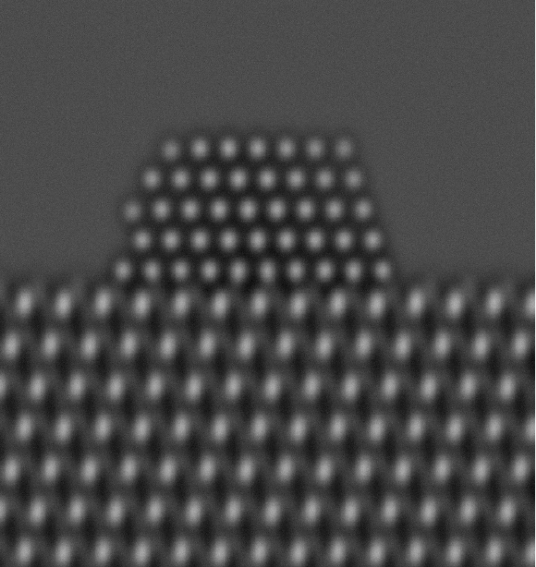}\nsp &
  \nsp\includegraphics[height=\f1ht]{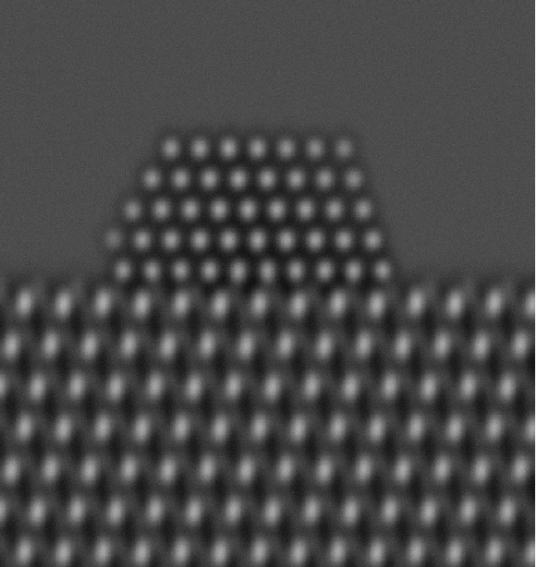} &
  \includegraphics[height=\f1ht]{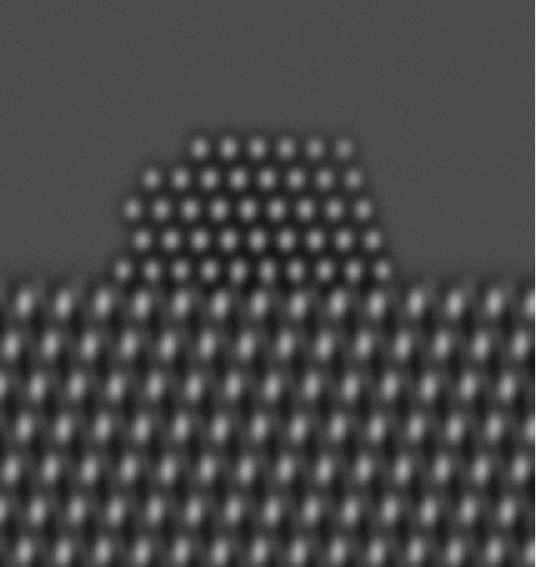}\nsp \\
  
  \footnotesize{Surface} & \nsp\includegraphics[height=\f1ht]{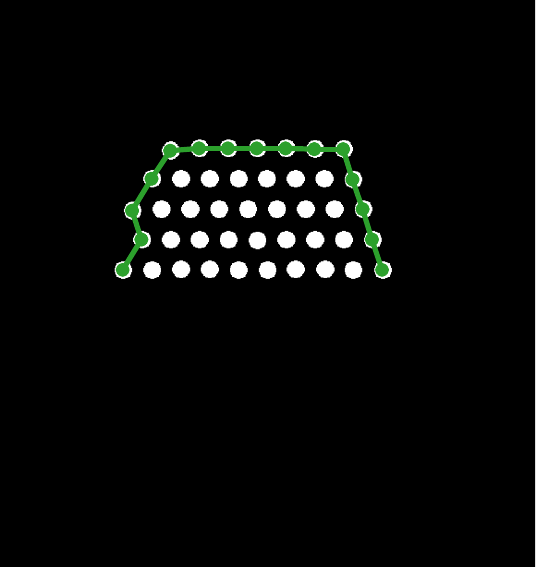}\nsp &
  \nsp\includegraphics[height=\f1ht]{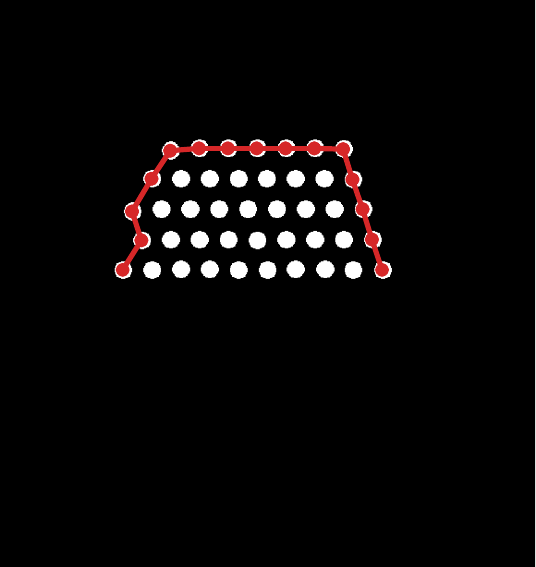}\nsp &
  \nsp\includegraphics[height=\f1ht]{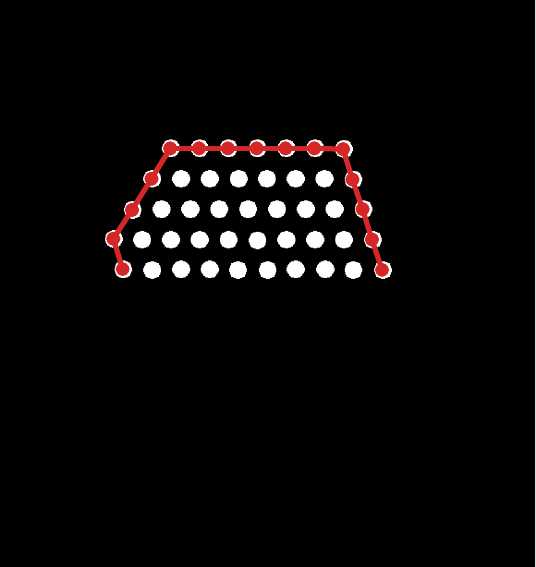} &
  \includegraphics[height=\f1ht]{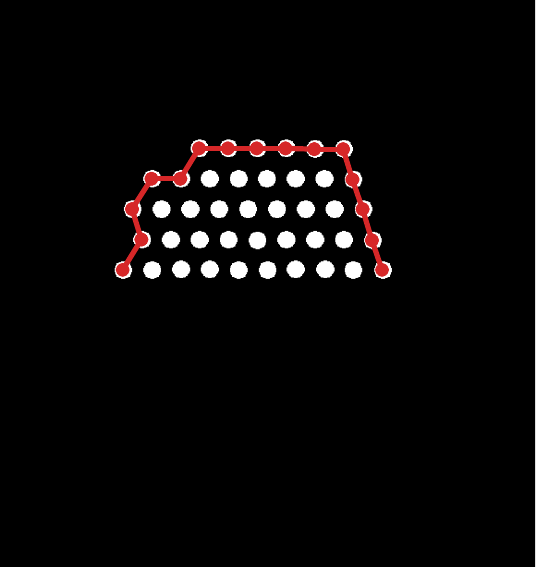}\nsp \\
  
  \vspace{0.1cm}
  
  \footnotesize{PSNR} &  - & \footnotesize{41.89} & \footnotesize{41.90} & \footnotesize{41.85} \\
  \footnotesize{SSIM} &  - & \footnotesize{0.946} & \footnotesize{0.948} & \footnotesize{0.949} \\
  \footnotesize{Precision} &  - & \footnotesize{1.00} & \footnotesize{0.933} & \footnotesize{0.933} \\
  \footnotesize{Recall} &  - & \footnotesize{1.00} & \footnotesize{0.933} & \footnotesize{0.933} \\
  \footnotesize{F1} &  - & \footnotesize{1.00} & \footnotesize{0.933} & \footnotesize{0.933} \\
  \footnotesize{Jaccard Index} &  - & \footnotesize{1.00} & \footnotesize{0.875} & \footnotesize{0.875}  \\
 
\end{tabular}

\caption{\textbf{Scientifically-meaningful metrics for atom detection}. In order to compare metrics in terms of their sensitivity to changes in atomic structure we perturb and compare the reference image in (a) to several modified images. 
In (b) the atomic structure is the same.
In (c) a spurious atom is added to the image in the top left. In (d) an atom is removed from the top left. The images in (b), (c), and (d) are further corrupted by adding iid Gaussian noise with a small deviation of around $2/255$. The PSNR and SSIM~\cite{wang2003new} of (b), (c) and (d) with respect to (a) are essentially constant, indicating that these metrics are not sensitive to changes in atomic configuration. In contrast, our proposed metrics reflect these changes more accurately: (b) is assigned a score of 1 in all metrics, whereas (c) and (d) are consistently assigned lower values. }
\label{fig:metrics_visual}
\end{figure}

Domain scientists denoise images in order to extract scientifically relevant information. In our case, the atoms on the surface of nanoparticles are of particular interest, because the atomic configuration at the surface regulates the nanoparticle's ability to catalyze chemical reactions. It is therefore of critical importance to understand how different denoising methods recover these atoms. We can verify visually that SBD achieves a largely successful recovery in held-out simulated data, whereas the baseline methods described in Section~\ref{sec:sbd_comparison} do not (see Figure~\ref{fig:simulation-denoised} for example). However, visual inspection is a limited and non-quantitative evaluation tool. Unfortunately, standard metrics like PSNR and SSIM 
are insensitive to changes in the atomic structure of the nanoparticle surface, because these changes have a small effect on the overall intensity of the images. We demonstrate the lack of sensitivity through a synthetic example in Figure~\ref{fig:metrics_visual}: when we add or remove an atom in the surface the PSNR and SSIM remain roughly constant. Motivated by the need for  scientifically-relevant performance evaluation, we propose several metrics explicitly designed to account for changes in surface atomic configuration in Section~\ref{sec:metrics_descr}. We report an evaluation of SBD using these metrics on a challenging test dataset in Section~\ref{sec:metrics_dataset_eval}.

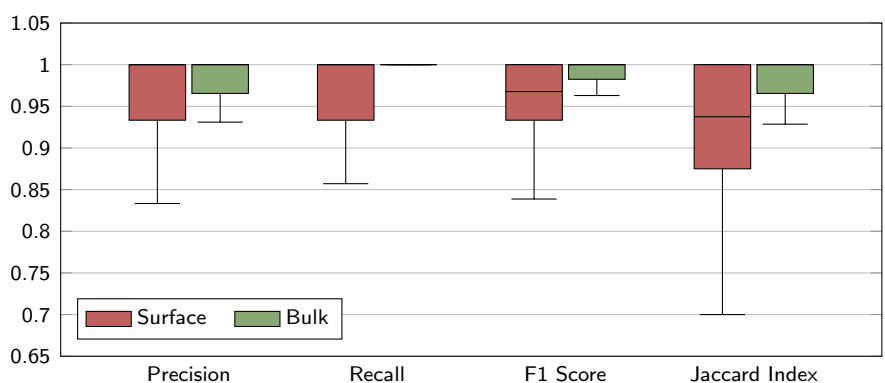
\begin{figure}[t]
    \centering
    \begin{tikzpicture}
        \pgfplotsset{
            ticklabel style={font=\scriptsize\sffamily\sansmath},
            every axis label/.append style={font=\sffamily\sansmath\scriptsize},
            title style={font=\footnotesize\sansmath}
        }
        \begin{axis}[
            height=6cm, width=0.8\linewidth,
            boxplot/draw direction=y,
            ymin=0.65,ymax=1.05, ymajorgrids=true, ytick distance=0.05,
            cycle list={knred!80, kngreen!80},
            boxplot={draw position={1/3 + floor(\plotnumofactualtype/2) + 1/3*mod(\plotnumofactualtype,2)}, box extend=0.3},
            xtick={0,1,2,...,10}, x tick label as interval, major x tick style=transparent,
            xticklabels={Precision, Recall, F1 Score, Jaccard Index}, x tick label style={text width=2.5cm, align=center},
            legend style={font=\scriptsize\sffamily\sansmath, legend columns=2, anchor=south west, at={(0.02, 0.05)}, /tikz/every even column/.append style={column sep=0.3cm}},
            legend image code/.code={\draw[#1, draw=black] (0cm,-0.1cm) rectangle (0.6cm,0.1cm);},
        ]
        \addplot+ [draw=black, fill, boxplot prepared={lower whisker=0.83333333, lower quartile=0.93333333,, median=1, upper quartile=1, upper whisker=1}] coordinates {};
        \addplot+ [draw=black, fill, boxplot prepared={lower whisker=0.93103448, lower quartile=0.96551724,, median=1, upper quartile=1, upper whisker=1}] coordinates {};

        \addplot+ [draw=black, fill, boxplot prepared={lower whisker=0.85714286, lower quartile=0.93333333,, median=1, upper quartile=1, upper whisker=1}] coordinates {};
        \addplot+ [draw=black, fill, boxplot prepared={lower whisker=1, lower quartile=1, median=1, upper quartile=1, upper whisker=1}] coordinates {};
        \addplot+ [draw=black, fill, boxplot prepared={lower whisker=0.83870968, lower quartile=0.93333333, median=0.96774194, upper quartile=1, upper whisker=1}] coordinates {};
        \addplot+ [draw=black, fill, boxplot prepared={lower whisker=0.96296296, lower quartile=0.98245614, median=1, upper quartile=1, upper whisker=1}] coordinates {};
        \addplot+ [draw=black, fill, boxplot prepared={lower whisker=0.7, lower quartile=0.875, median=0.9375, upper quartile=1, upper whisker=1}] coordinates {};
        \addplot+ [draw=black, fill, boxplot prepared={lower whisker=0.92857143, lower quartile=0.96551724, median=1, upper quartile=1, upper whisker=1}] coordinates {};
        \addlegendentry[]{Surface};
        \addlegendentry[]{Bulk};
        \end{axis}
    \end{tikzpicture}
    \caption{\textbf{Performance of SBD in terms of our proposed metrics}. We compute all our proposed metrics (see Section~\ref{sec:metrics_descr}) on over $7,000$ denoised images corresponding to $25$ unique noisy images sampled from the $308$ clean images described in Section~\ref{sec:metrics_dataset_eval}. The empirical distribution on the surface (red) and bulk (green) is visualized as box plots indicating the median, 25$^{th}$ quartile, 75$^{th}$ quartile, minimum and maximum value of the distribution. SBD has a near perfect performance in the bulk with all metric values hovering around $1$. On the surface, SBD achieves a median score of $1$ for precision and recall, and about $0.95$ for F1 score and Jaccard index.}
    \label{fig:metrics_box}
\end{figure}

\subsubsection{Evaluation metrics}
\label{sec:metrics_descr}


 
 To define metrics that evaluate detection of surface atoms, we assume that there is a predefined approach to perform detection based on the denoised images. In our case of interest, we apply a blob detection algorithm (Laplacian of Gaussian~\cite{lindeberg2013scale}) to locate the centers, and compute the $\alpha$-shape of all the atom centers using Delaunay triangulation~\cite{preparata1985convex}. Let $A$ and $B$ be the set of surface atoms of interest in the denoised image and the ground-truth clean image respectively. We propose the following four metrics to measure the fidelity of the recovered structure:
\begin{itemize}
    \item \textbf{Precision} is the fraction of atoms in the denoised image that are also present in the clean image.  
    \begin{equation}
        P(A, B) = \frac{|A \cap B|}{|B|}.
    \end{equation}
    \item \textbf{Recall} is the fraction of atoms in the clean image that are correctly recovered in the denoised image. 
    \begin{equation}
        R(A, B) = \frac{|A \cap B|}{|A|}.
    \end{equation}
    \item \textbf{F1 score} combines precision and recall by giving them equal importance. 
    \begin{equation}
        F(A, B) = 2 \frac{P(A, B)  R(A, B)}{P(A, B) + R(A, B)}.
    \end{equation}
    \item \textbf{Jaccard index} is an alternative measure consisting of the ratio between the size of the intersection between the recovered atoms and the ground truth divided by the size of their union. \begin{equation}
        J(A, B) = \frac{|A \cap B|}{|A \cup B|}.
    \end{equation}
\end{itemize}

When performing intersection and union operations, we consider two atoms to be the same if the distance between their centers is less than a threshold of $100$ pixels. All our metrics take values between $0$ and $1$ ($1$ is best). Figure~\ref{fig:metrics_visual} shows a synthetic example comparing three images with different atomic configurations: all the images have similar PSNR and SSIM values, but the precision, recall, F1 and Jaccard index show substantial differences.

\subsubsection{Evaluating atom detection accuracy}
\label{sec:metrics_dataset_eval}


To evaluate the performance of the proposed approach to recover atoms at the surface, we designed a new dataset with 308 images, where the imaging parameters are set based on the real dataset described in Section~\ref{sec:real_dataset}. This new dataset is similar to the one used for the generalization experiments in Section~\ref{sec:generalization}, but here we add more diverse surface defects. We created a series of 44 Pt/CeO$_2$ structural models with atomic-level surface defects such as the removal of an atom from a column, removal of two atoms, removal of all but one atom and addition of an atom at a new site (see Figure~\ref{fig:metrics_dataset_fig} for a visual overview). We hypothesize that these defects emulate dynamic atomic-level reconfigurations that could potentially be observed in real experiments. To match the image contrast of our real data, we simulated images under defocus values ranging from 6 nm to 10 nm, all with a tilt of 3$^{\circ}$ in x and -1$^{\circ}$ in y and a support thickness of 40 {\AA}. 
SBD recovers all the atoms in the bulk almost perfectly, as reflected in the different metrics. On the surface, SBD achieves a median score of $1$ for precision and recall, and more than $0.95$ on F1 and Jaccard index (see Figure~\ref{fig:metrics_box}).

\begin{figure}[t]
     \centering
     \includegraphics[width=\linewidth]{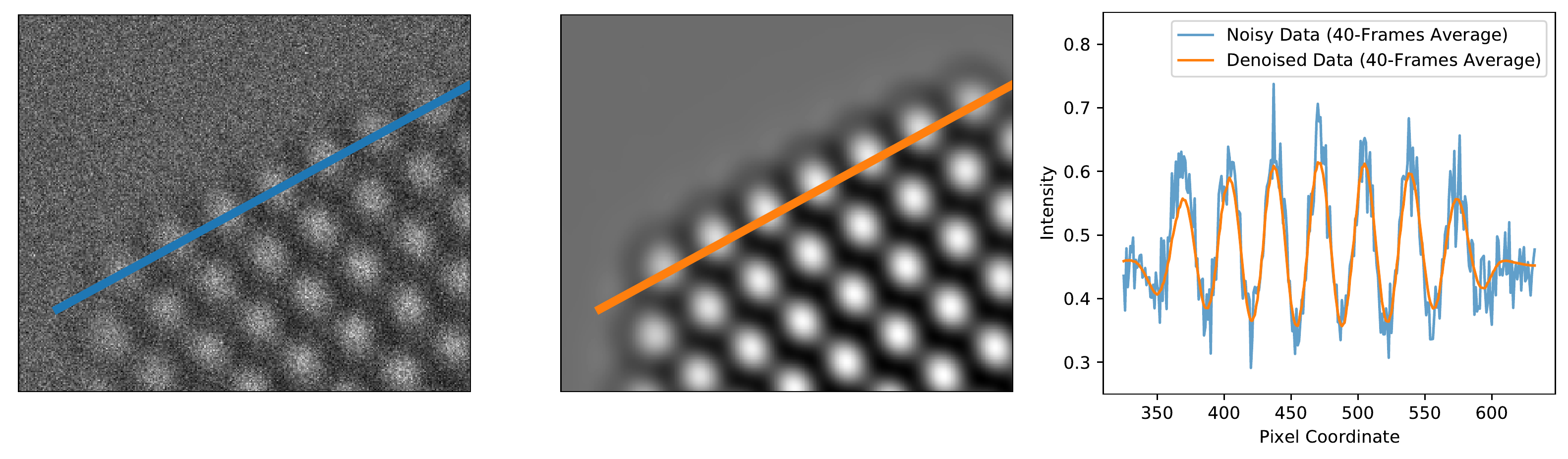}
     \caption{\textbf{Validation on real data}. The real data consist of $40$ frames which are approximately stationary and aligned. Their temporal average (left) therefore provides a reasonable estimate for the true intensity profile. In the image on the right, we compare the average intensity profile on the surface atomic columns of the platinum nanoparticle for the denoised data (middle) and the temporal average (left). The profiles are very similar (except for some spurious fluctuations in the temporal average), which suggests that the proposed approach achieves effective denoising on the real data.}
     \label{fig:linescan}
 \end{figure}

\begin{figure*}[!t]
    \centering
    \begin{tabular}{c@{\hskip 0.01in}c@{\hskip 0.01in}c@{\hskip 0.01in}c@{\hskip 0.01in}c@{\hskip 0.01in}c}
    \footnotesize{Noisy} & \footnotesize{WF} & \footnotesize{Spot Filter} & \footnotesize{VST+NLM} & \footnotesize{VST+BM3D}  &  \\
    \includegraphics[width=1.0in]{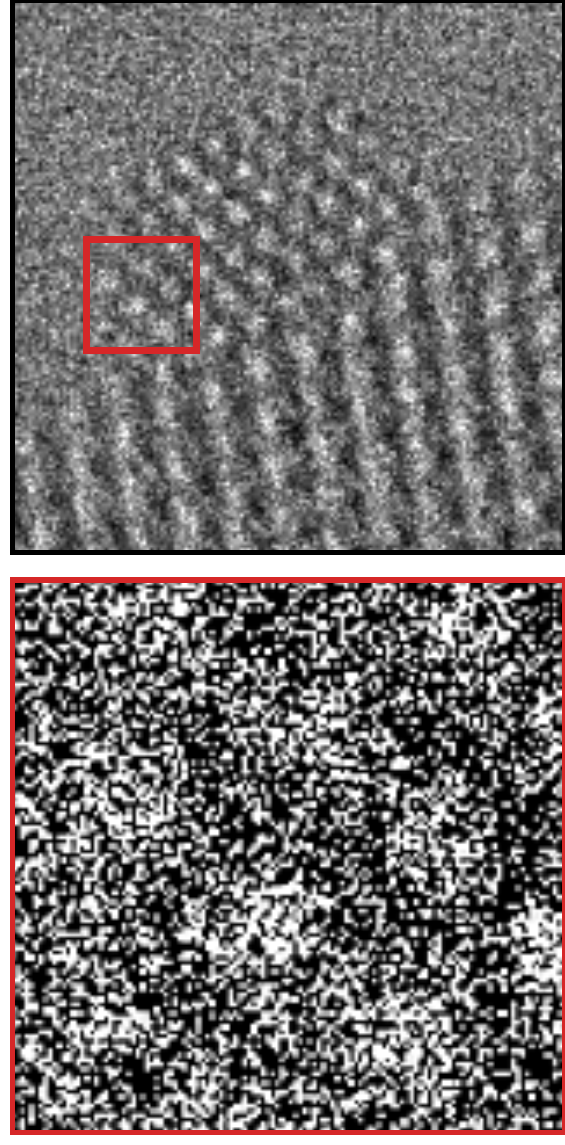}&
    \includegraphics[width=1.0in]{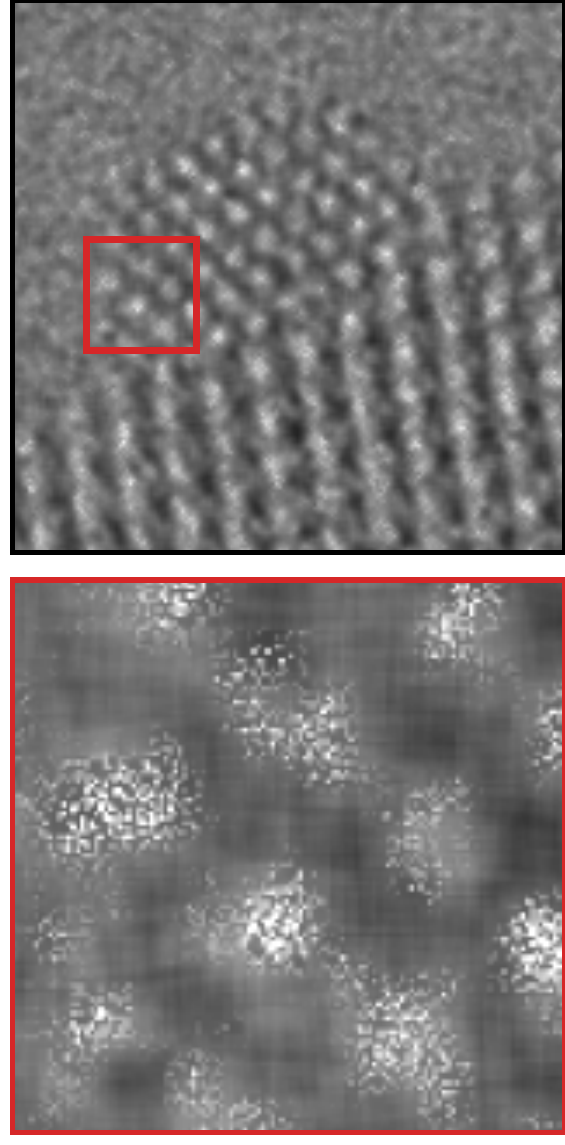}&
    \includegraphics[width=1.0in]{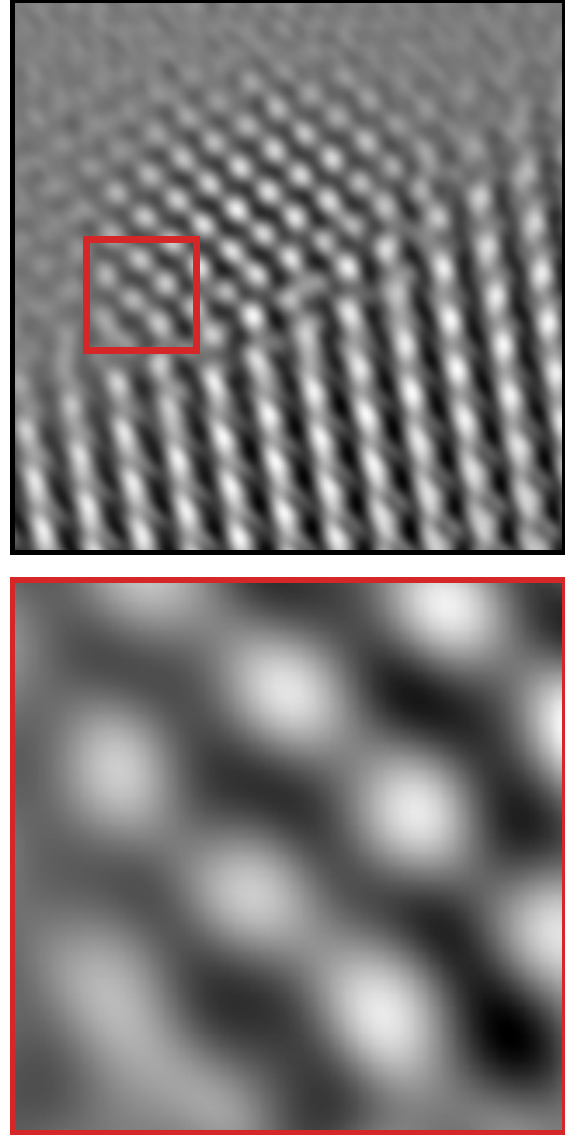}&
    \includegraphics[width=1.0in]{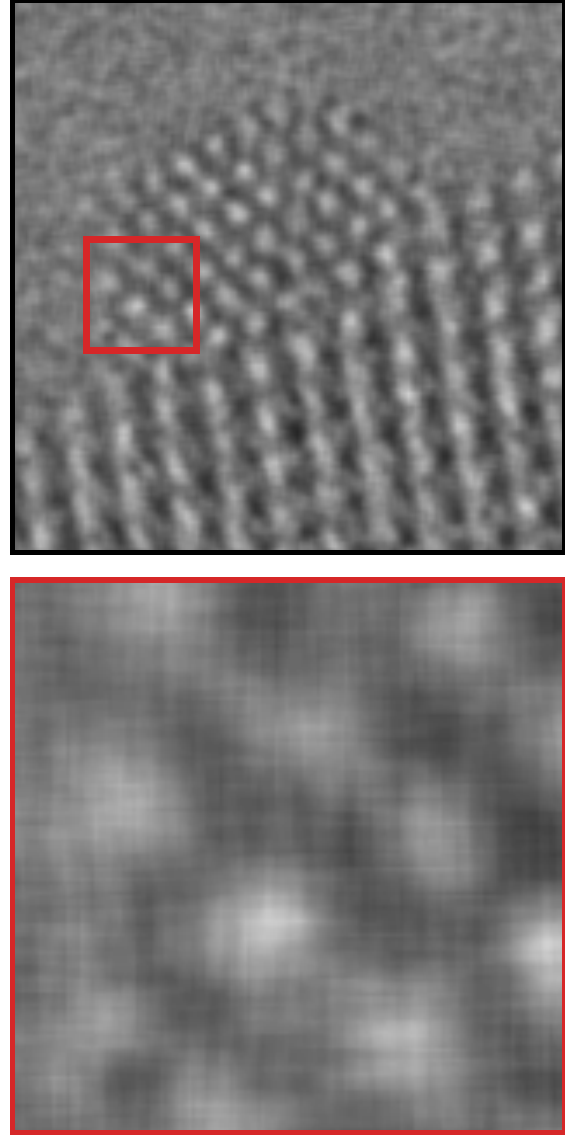}&
    \includegraphics[width=1.0in]{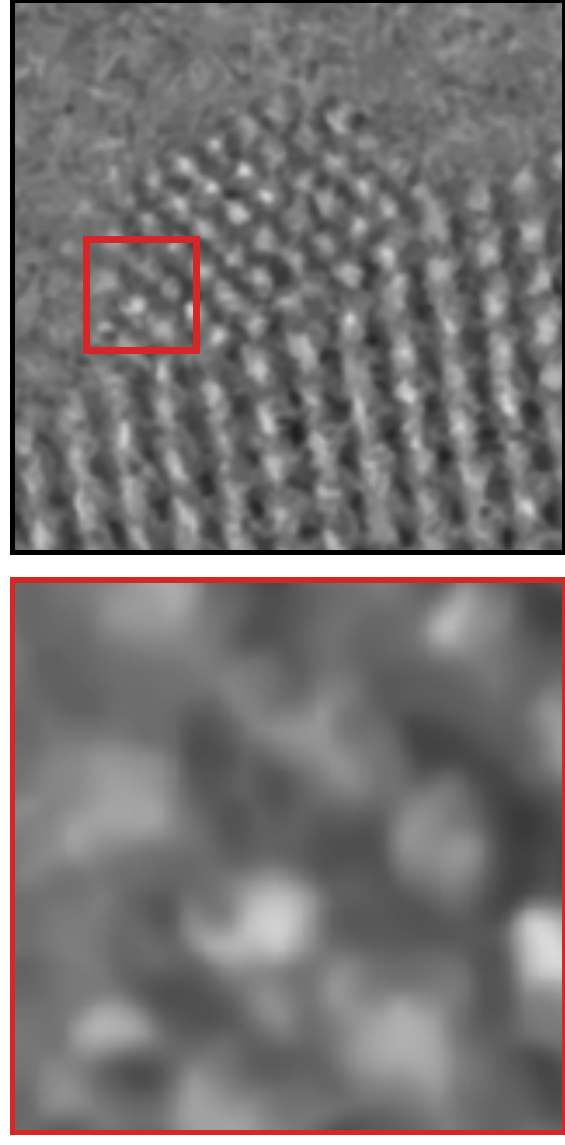}& \\

    \footnotesize{PURE-LET} & \footnotesize{SBD+DnCNN} & \footnotesize{SBD+Small UNet} & \footnotesize{Ours} & \footnotesize{Likelihood Map} &  \\
    \includegraphics[width=1.0in]{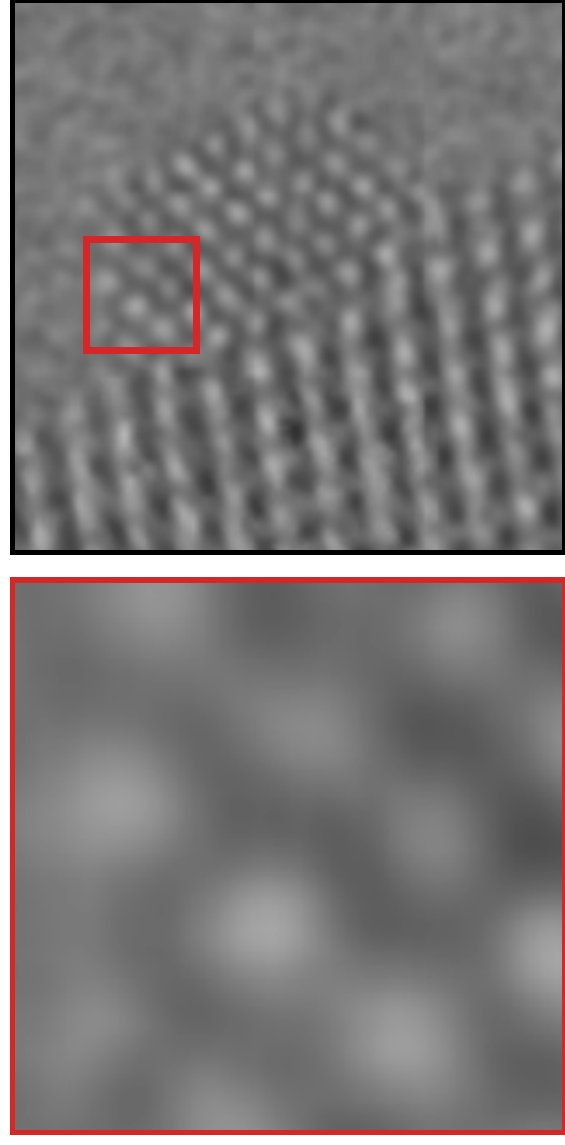}& \includegraphics[width=1.0in]{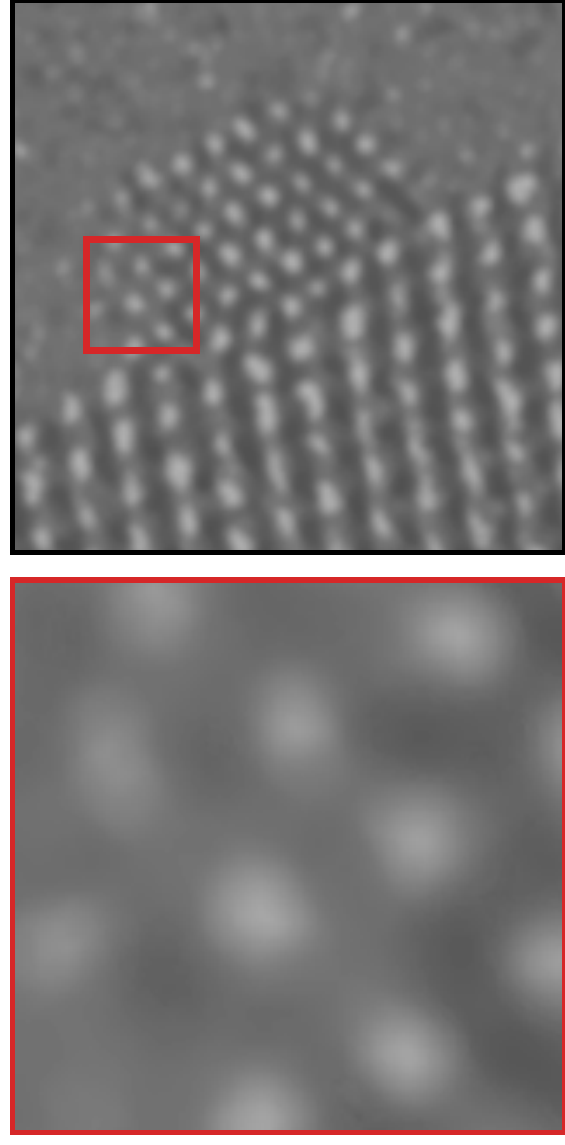}&
    \includegraphics[width=1.0in]{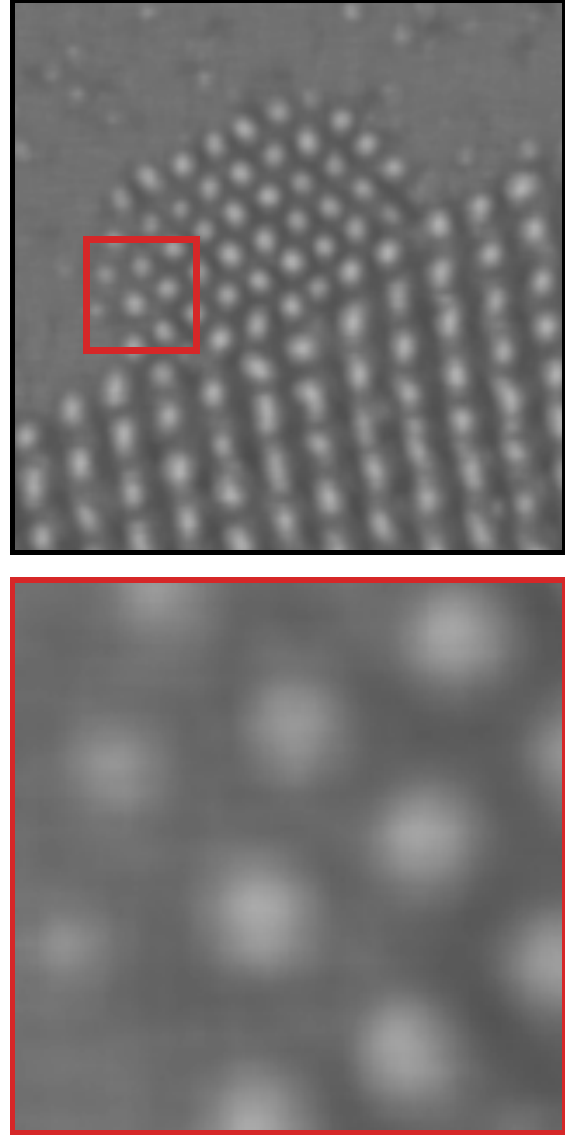}&
    \includegraphics[width=1.0in]{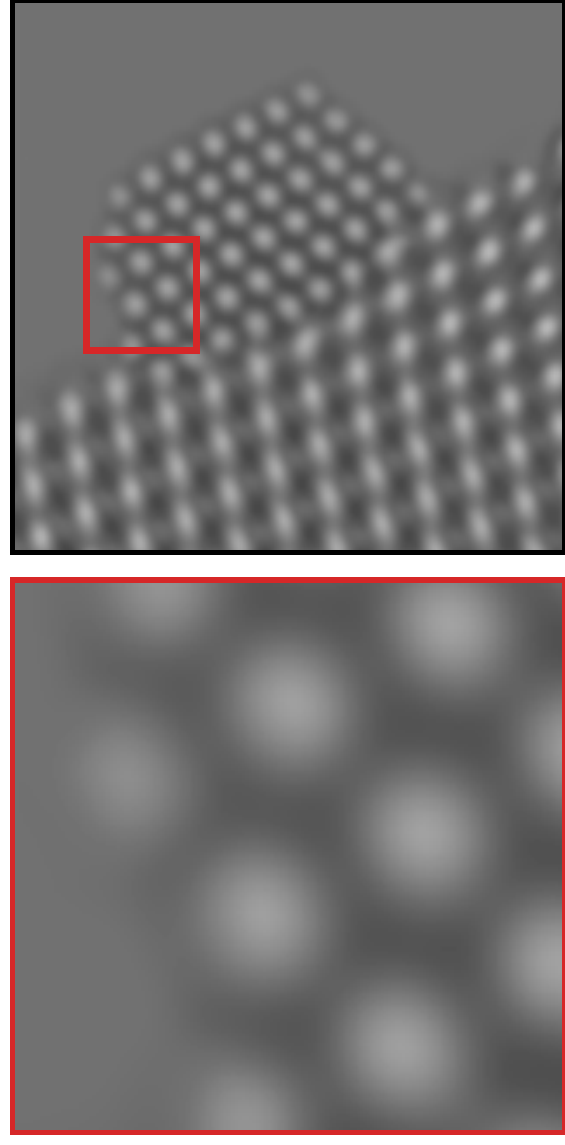}&
    \includegraphics[width=1.0in]{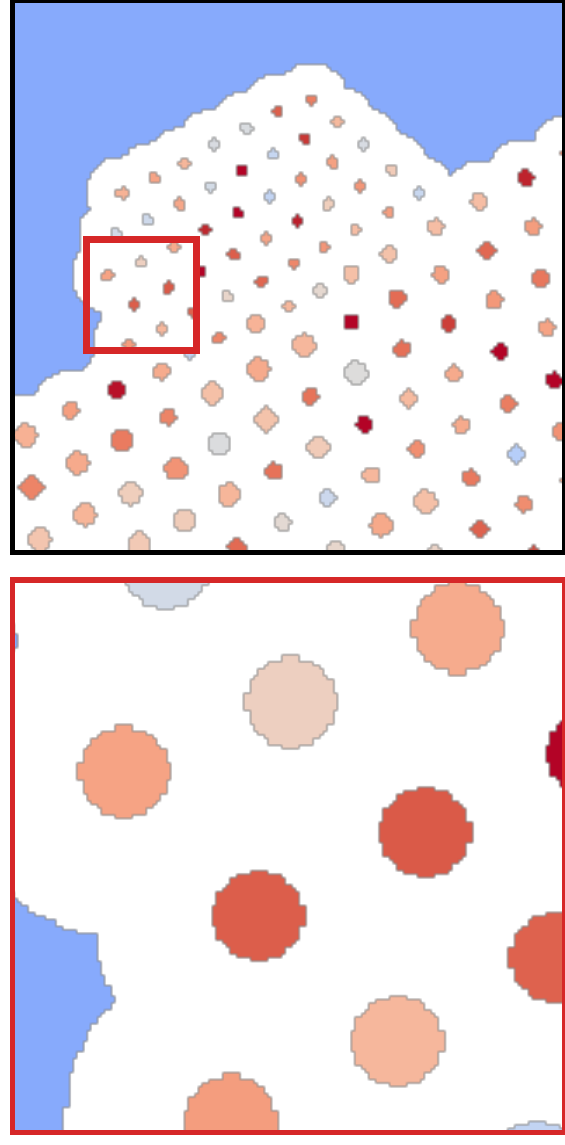}&
    \includegraphics[width=0.335in]{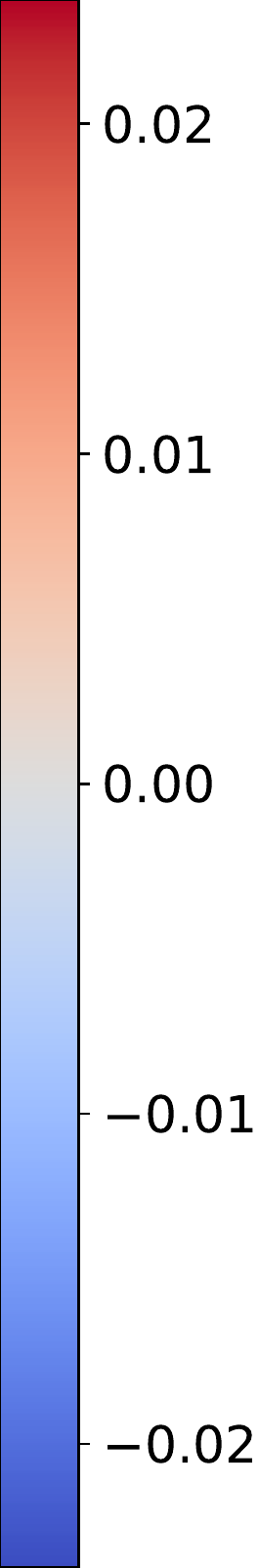}\\
    \end{tabular}
    \caption{\textbf{Denoising results for real data}. Comparison SBD and the baseline methods described in Sections~\ref{sec:sbd_comparison} and~\ref{sec:architectures} when applied on the real data described in Section~\ref{sec:real_dataset}. The second row zooms in on the region in red box. In contrast to the other methods, SBD combined with the proposed architecture is able to precisely recover the structure of the nanoparticle and has very few artefacts, particularly in the vacuum region. The likelihood map quantifies the agreement between recovered structures in the denoised images, such as atomic columns and the vacuum, and the observed data (see Section~\ref{sec:likelihood} for more details). See Figure~\ref{fig:real-denoised-appendix2} for an additional example}
    \label{fig:real-denoised-appendix1}
\end{figure*}

\subsection{Performance on real data}
\label{sec:exp_real_data}

In the experiments reported in Sections~\ref{sec:sbd_comparison} and \ref{sec:metrics} we used a network trained on all simulated images from the white contrast category defined in Section~\ref{sec:generalization}. However, the real data described in Section~\ref{sec:dataset} more closely corresponds to a subset of white contrast images satisfying the following conditions: structure limited to PtNP2, thickness between 40 {\AA} - 60 {\AA} and, defocus between 5 nm and 10 nm. We used 236 images from this subset for training, and another such 15 images for validation. We also trained two state-of-the-art architectures for photographic image denoising - DnCNN~\cite{zhang2017beyond} and DURR~\cite{zhang2018dynamically} on these data.

Results on real experimental data obtained using SBD trained on this relevant subset of white contrast are shown in Figures~\ref{fig:experiment-denoised}, \ref{fig:real-denoised-appendix1}, and \ref{fig:real-denoised-appendix2}. SBD produces denoised images that are of much higher quality than those of the baseline methods described in Section~\ref{sec:sbd_comparison}, which contain obvious artefacts. Further, we validate the denoising results of SBD by comparing to an estimated reference image obtained by temporal averaging. Our real dataset consists of $40$ frames that are approximately stationary and aligned. Therefore, their temporal average provides a good estimate for the ground-truth images. As shown in Figure~\ref{fig:linescan}, the denoised intensity values of the atomic column approximately match those of the estimated reference image. 

In the rest of this section, we compare the performance of SBD and unsupervised denoising techniques on the real experimental data, and analyze the effect of the design of the training dataset on the denoised output produced by SBD.

\begin{figure}[t]
\centering
        \begin{tabular}{c@{\hskip 0.005in}c@{\hskip 0.005in}c@{\hskip 0.005in}c@{\hskip 0.005in}c@{\hskip 0.005in}c@{\hskip 0.005in}c@{\hskip 0.005in}}
        \footnotesize{(a) Data} &  \footnotesize{(b) Blind-spot~\cite{laine2019high} } & \footnotesize{(c) Blind-spot$^*$} & \footnotesize{(d) UDVD$^{*\dagger}$~\cite{udvd} } & \footnotesize{(e) Self2Self~\cite{self2self}} & \footnotesize{(g) SBD}  & \footnotesize{(g) Estimated ref.} \\
        \includegraphics[width=0.13\linewidth]{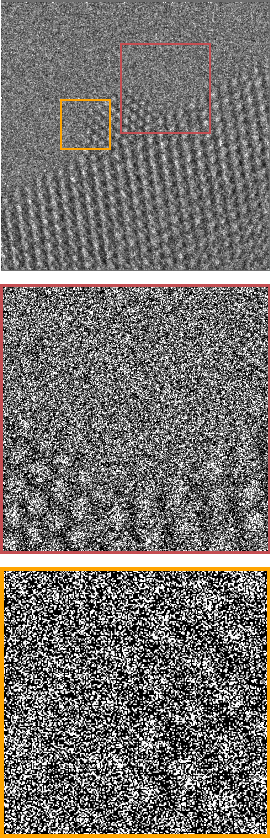} &
        \includegraphics[width=0.13\linewidth]{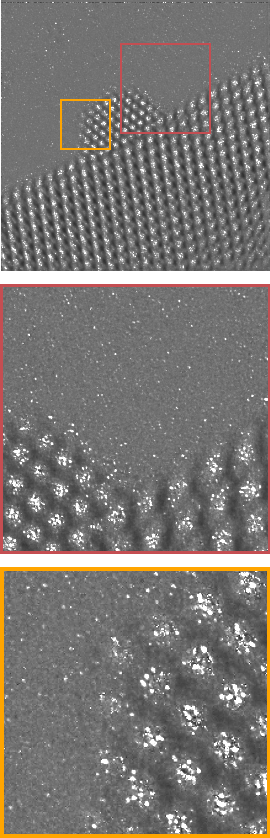} &
        \includegraphics[width=0.13\linewidth]{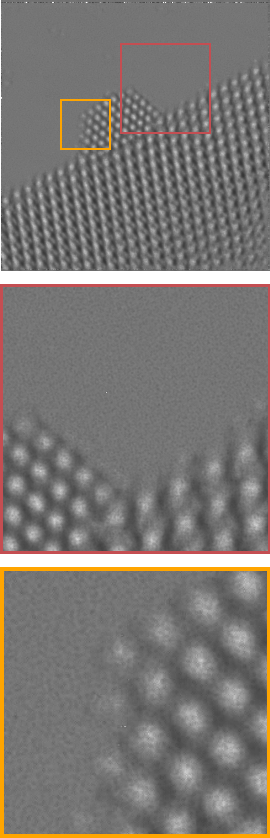} &
        \includegraphics[width=0.13\linewidth]{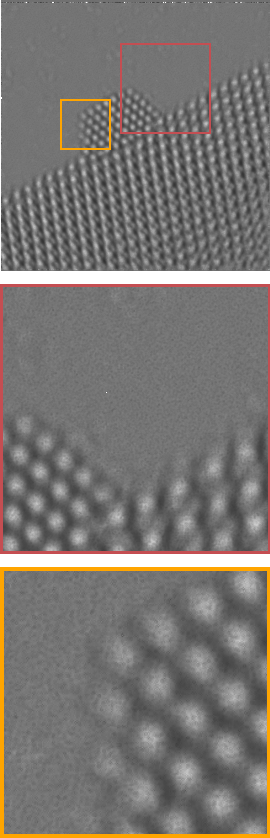} &
        \includegraphics[width=0.13\linewidth]{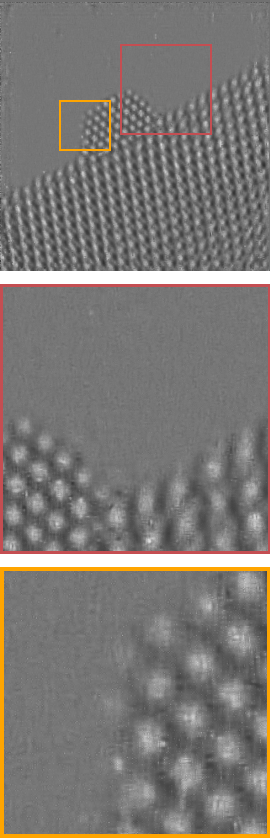} &
         \includegraphics[width=0.13\linewidth]{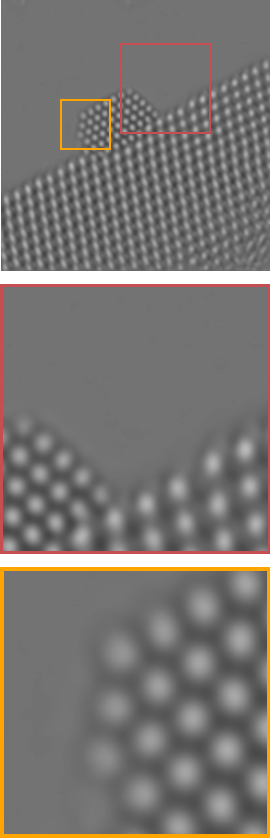}&
        \includegraphics[width=0.13\linewidth]{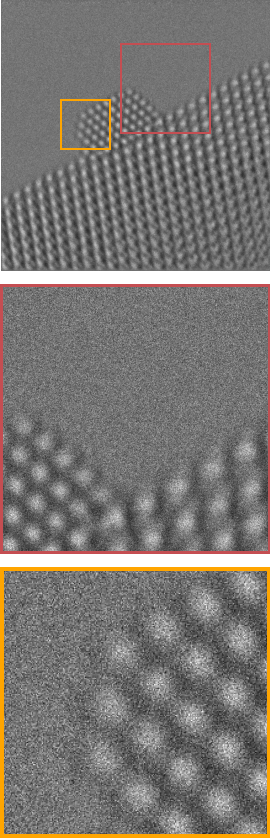} \\
        \end{tabular}
\caption{\textbf{Comparison of unsupervised denoising methods with SBD on real data}. The real data described in Section~\ref{sec:real_dataset} denoised using SBD and unsupervised methods described in Section~\ref{sec:unsupervised}. The second and third rows zoom in on the region in red and green boxes respectively. Our proposed method denoises the real data more effectively than the unsupervised approaches. SBD is able to precisely recover the structure of the nanoparticle and has very few artefacts (compare visually to the estimated reference image obtained via time averaging; there are three missing atoms for most unsupervised methods in the third row). A $*$ indicates that the method used early stopping and $\dagger$ indicates that the method uses $5$ noisy frames as input. }
\label{fig:unsup_comparison}
\end{figure}

\definecolor{knred}{RGB}{175,57,55}
\begin{figure}[t]
    \centering
    \begin{tikzpicture}[declare function={barWidth=10pt; barShift=barWidth/2;}]
        \pgfplotsset{
            xtick=data, major x tick style=transparent,
            ticklabel style={font=\scriptsize\sffamily\sansmath, yshift=3pt},
            every axis label/.append style={font=\sffamily\sansmath\scriptsize},
            title style={font=\footnotesize\sansmath, yshift=-3pt},
        }
        \begin{axis}[
                height=5cm, width=0.8\linewidth,
                ybar=0pt, bar width=barWidth,
                xmin=40, xmax=5024, symbolic x coords={40, 50, 100, 200, 500, 1000, 2000, 3000, 4000, 5024}, enlarge x limits=0.1, xtick=data,
                ymin=20, ymax=50, ymajorgrids=true, ytick distance=5,
                legend style={font=\scriptsize\sffamily\sansmath, legend columns=2, /tikz/every even column/.append style={column sep=0.3cm}},
                legend image code/.code={\draw[#1, draw=black] (0cm,-0.1cm) rectangle (0.6cm,0.1cm);},
                title={},
                ylabel=Performance in PSNR,
                xlabel=Number of training data points, xlabel style={yshift=2pt},
                enlarge x limits=0.1,
            ]
            \addplot+[draw=black, fill=knred!80] coordinates {
                (40, 23.697307009442095)
                (50, 23.312291143483456)
                (100, 24.972936804225018)
                (200, 27.426546623259732)
                (500, 30.015963304213347)
                (1000, 33.412447852785185)
                (2000, 38.0762393367052)
                (3000, 39.45859673073849)
                (4000, 39.751043385465465)
                (5024, 40.37742722010145)
            }; \addlegendentry{Unsupervised}

            \coordinate (A) at (axis cs:40, 41);
            \coordinate (O1) at (rel axis cs:0,0);
            \coordinate (O2) at (rel axis cs:1,0);
            \draw [black, sharp plot,dashed] (A -| O1) -- (A -| O2); \addlegendentry{Supervised}
            \addlegendimage{line legend,black,sharp plot,dashed}
        \end{axis}
    \end{tikzpicture}
    \caption{\textbf{Training set size and unsupervised denoising}. Performance of blind-spot net~\cite{laine2019high} on a held-out set of images, measured in PSNR, when trained on simulated training sets of different sizes. The held-out set and the training data comprise white contrast images. The performance is poor when the training set is small, but improves to the level of supervised approaches (denoted by dashed line) when trained using more training data. Our real dataset described in Section~\ref{sec:real_dataset} contains only 40 images, so it is likely that the limited amount of data explains the poor performance of the blind-spot net in Figure~\ref{fig:unsup_comparison}. }
    \label{fig:unsup_data}
\end{figure}
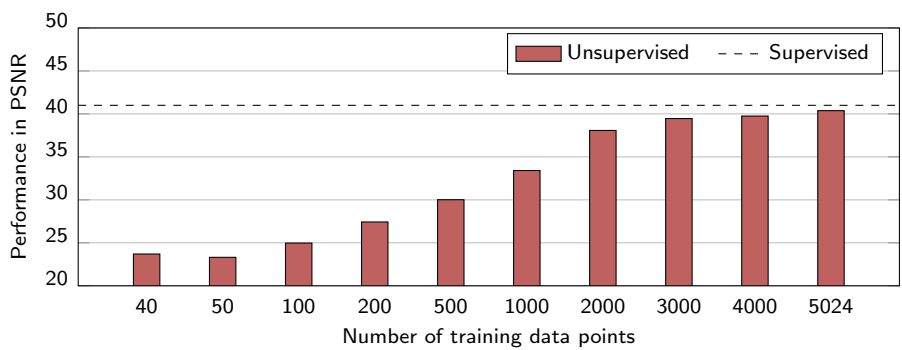

\subsubsection{Comparison to unsupervised deep denoising methods}
\label{sec:unsupervised}

Unsupervised denoising techniques can be used to train a denoising CNN using only noisy images (see Section~\ref{sec:related_work} for a discussion on this methodology). We apply the following unsupervised methods to the real data described in Section~\ref{sec:real_dataset}:
\begin{itemize}
    \item \textbf{Blind-spot net}~\cite{laine2019high} is a CNN which is constrained to predict the intensity of a pixel as a function of the noisy pixels in its neighbourhood, without using the pixel itself. This method is competitive with the current supervised state-of-the-art CNN on photographic images. However, when applied to our real dataset it produces denoised images with visible artefacts (see Figure~\ref{fig:unsup_comparison}(c)). A possible explanation is the limited amount of data ($40$ noisy images) we train on. To validate this hypothesis, we trained a blind-spot net on simulated training sets of different sizes. The performance on  held-out data is indeed poor when the training set is small, but it improves to the level of supervised approaches as we use more training data (see Figure~\ref{fig:unsup_data}). 
    
    \item \textbf{Blind-spot net with early stopping}. 
    In Ref.~\cite{udvd} it is shown that early stopping based on noisy held-out data can boost the performance of blind-spot nets. 
    Here we used $35$ images for training the blind-spot net and the remaining $5$ images as a held-out validation set. We chose the model parameters that minimized the mean squared error between the noisy validation images and the corresponding denoised estimates.
    The results (shown in Figure~\ref{fig:unsup_comparison}(d)) are significantly better than those of the standard blind-spot network. However, there are still noticeable artefacts, which include missing atoms. 
    
    \item \textbf{Unsupervised Deep Video Denoising (UDVD)}~\cite{udvd} is an unsupervised method for denoising video data based on the blind-spot approach. It estimates a denoised frame using $5$ consecutive noisy frames around it. Our real data consists of $40$ frames acquired sequentially. UDVD produces better results than blind-spot net, but still contains visible artefacts, including missing atoms (see Figure~\ref{fig:unsup_comparison}(d)). Note that, UDVD uses $5$ noisy images as input, and thus has more context to perform denoising than the other methods (including SBD).

    \item \textbf{Self2Self}~\cite{self2self} is an unsupervised method specifically designed for denoising based on a single noisy image. This approach achieves near state-of-the-art performance in noisy photographic images corrupted with moderate amounts of noise~\cite{self2self}. However, when applied to our data, Self2Self produces images with clear artefacts; some of the atoms are missing and the shape of atoms are distorted (see Figure~\ref{fig:unsup_comparison}(e)).
\end{itemize}

It is important to note that the backbone architectures of all these methods are UNets with large fields of view, like the one used for SBD. 
In our experiments, we trained the blind-spot nets and UDVD on $600 \times 600$ patches extracted from the real data. We used Adam optimizer~\cite{kingma2014adam} with a starting learning rate of $1\times 10^{-4}$ which was reduced in half for every $2000$ epochs. We trained for a total of $5000$ epochs. When performing early stopping, we picked the checkpoint with the best mean squared error on the validation set. Following Ref.~\cite{self2self}, Self2Self was trained for $150,000$ steps with the Adam optimizer and a starting learning rate of $10^{-4}$.  

As shown in Figure~\ref{fig:unsup_comparison}, the unsupervised denoising methods produce higher-quality reconstructions than those of the baseline methods discussed in Section~\ref{sec:sbd_comparison} (see Figure~\ref{fig:real-denoised-appendix1}). However, they still suffer from visible artefacts, particularly on the surface of the nanoparticle, limiting their practical utility. UDVD is the method that achieves best performance, but it requires multiple noisy frames as input. In contrast, SBD can denoise the image effectively from a single noisy input frame (see Figure~\ref{fig:unsup_comparison}(g)), as long as the simulated training data correspond closely to the real noisy image. Using a single frame is important in some applications, such as our case of interest, where the ultimate goal is to identify dynamic changes in the atomic structure of the nanoparticle.   


\subsubsection{A word of caution: Effect of training data on SBD}
\label{sec:datashift}

\begin{figure*}[t]
    \centering
    \begin{tabular}{c@{\hskip 0.01in}c@{\hskip 0.01in}c@{\hskip 0.01in}c@{\hskip 0.01in}c}
    
    \footnotesize{(a) Estimated ref.} &  \multicolumn{2}{c}{\footnotesize{SBD } } & \multicolumn{2}{c}{\footnotesize{Unsupervised} } \\
    
    \cmidrule(lr){2-3}
    \cmidrule(lr){4-5}
        
     &  \footnotesize{(b) white contrast} & \footnotesize{(c) adapted data} & \footnotesize{(d) Blind-spot} & \footnotesize{(e) Self2Self} \\
    \includegraphics[width=1.1in]{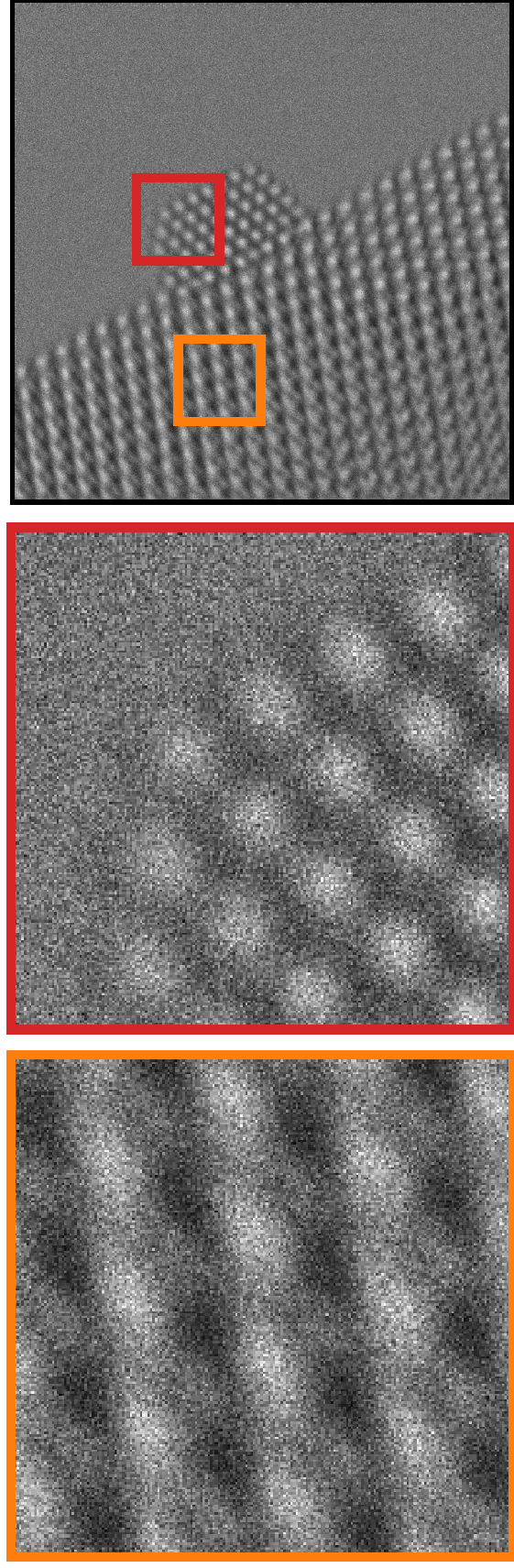}&
    \includegraphics[width=1.1in]{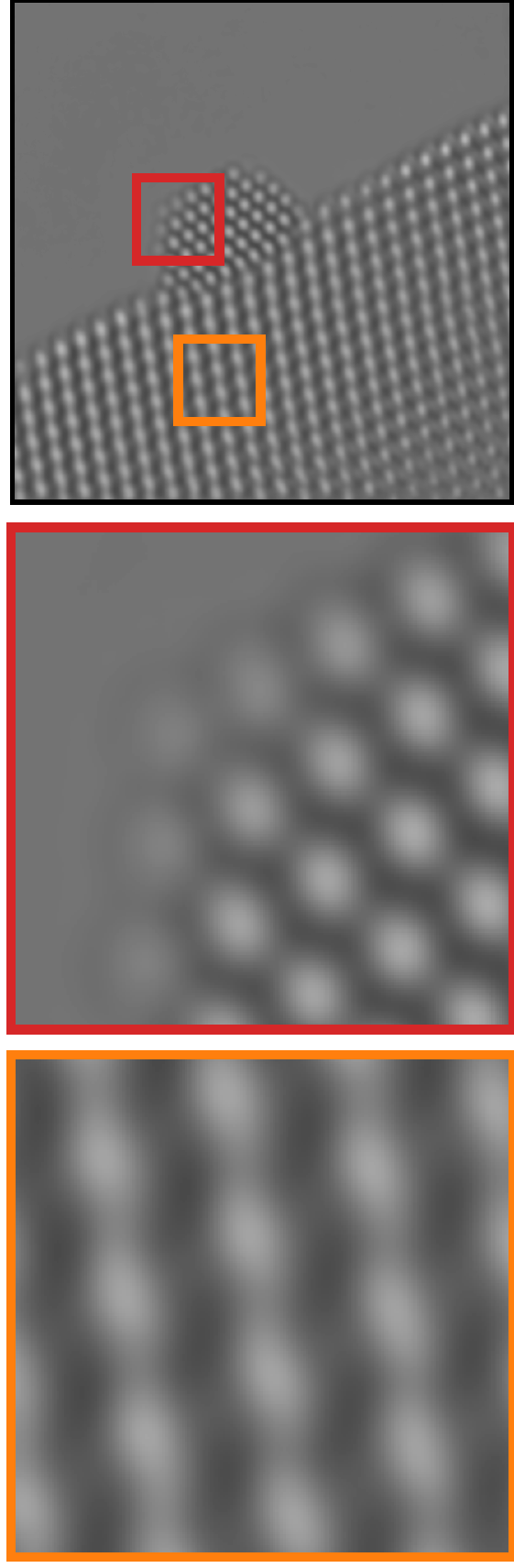}&
    \includegraphics[width=1.1in]{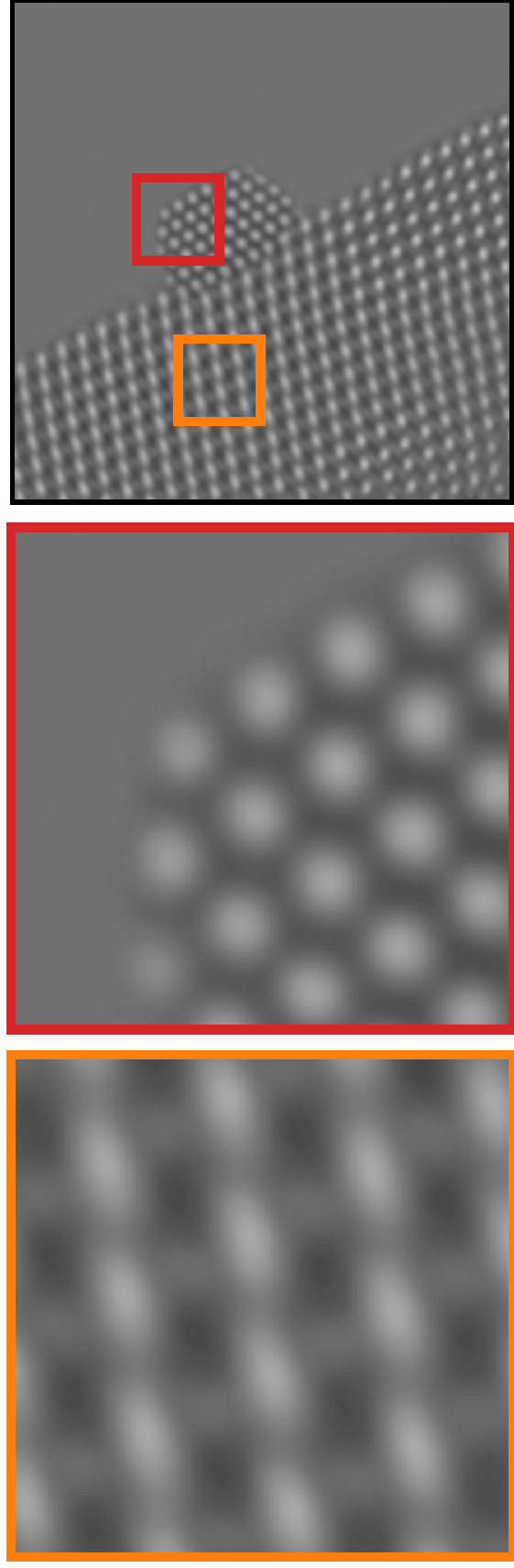}&
    \includegraphics[width=1.1in]{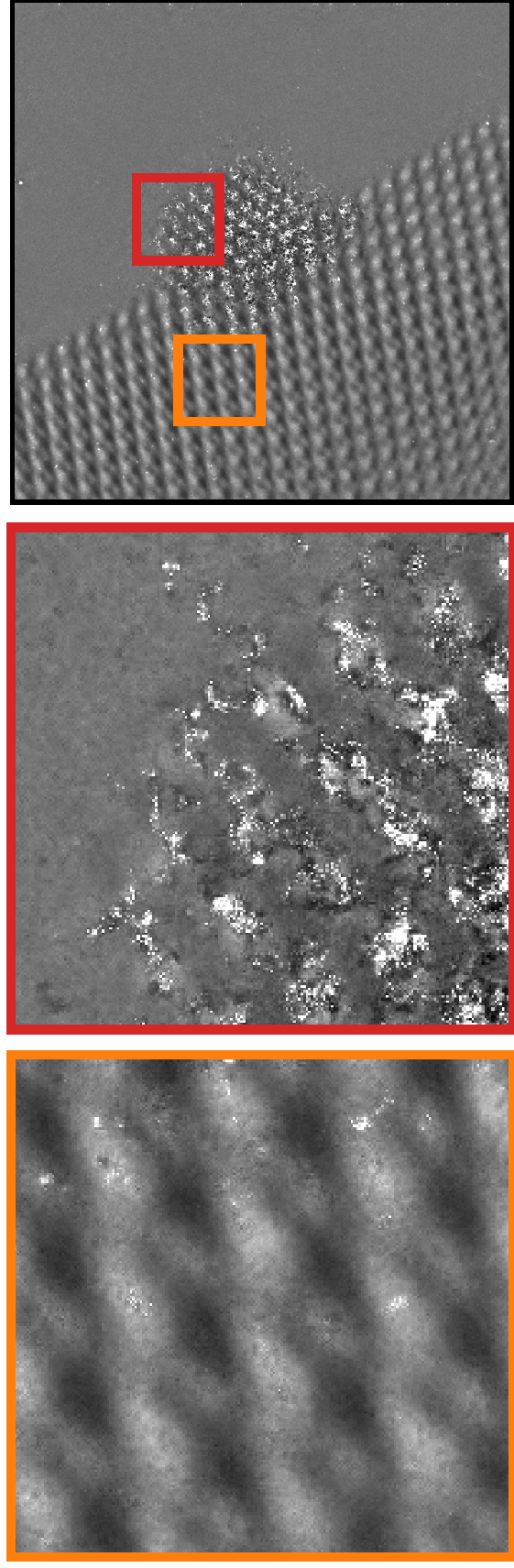}&
    \includegraphics[width=1.1in]{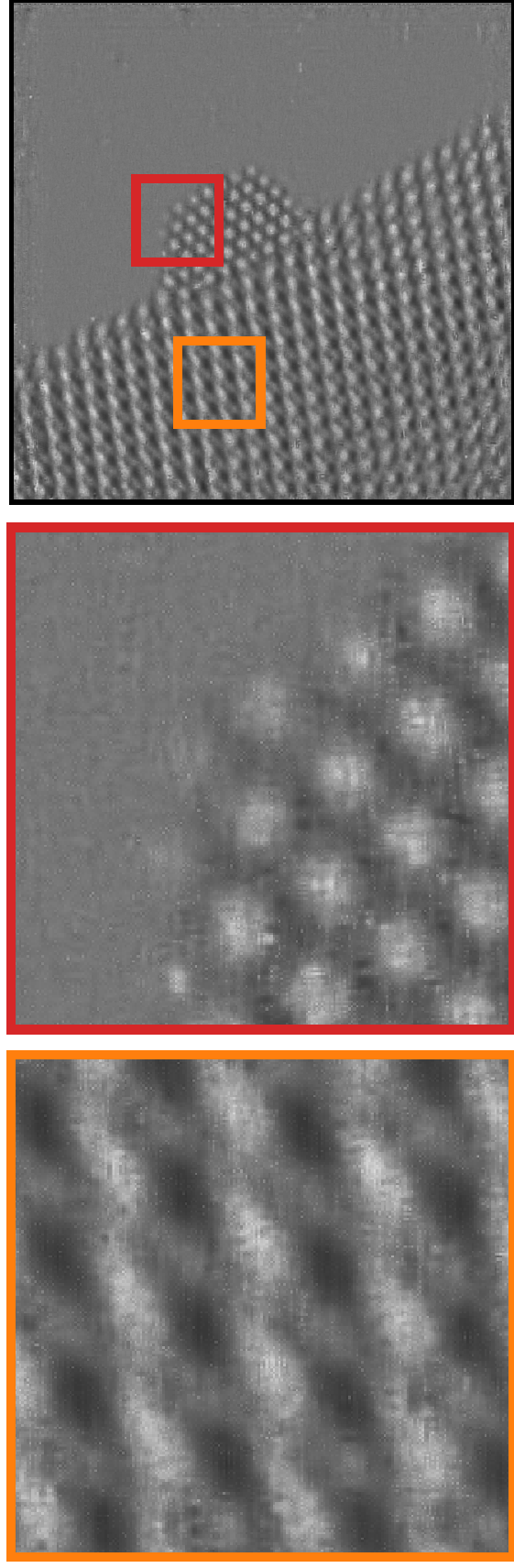}\\
    \end{tabular}
    \caption{\textbf{Effect of training data on SBD}. The real data consists of 40 frames which are approximately stationary and hence their temporal mean (a) can be used as an estimate for the reference image. The second and third row zooms into the red and yellow region respectively. When the network is trained on white contrast (i.e, the training data does not align well with real data), the denoised image of the real data (b) shows an unnatural streak pattern in the bulk (compare row 3 in (b) and (a)). Interestingly, when the training data is more reflective of the real data, the patterns in the bulk are recovered better (row 3 in (c) and (a)). Further, note that unsupervised methods denoise the contrast patterns in the bulk well (row 3 in (d), (e) and (a)) but they suffer from significant artefacts on the nanoparticles (row 2 in (d), (e) and (a)). }
    \label{fig:datashift}
\end{figure*}

Figures~\ref{fig:real-denoised-appendix1},~\ref{fig:real-denoised-appendix2} and~\ref{fig:unsup_comparison} show that SBD achieves impressive results on real data, but it is important to point out that this requires a careful design of the training dataset. 
Our real data broadly corresponds to images in the white contrast category, defined in Section~\ref{sec:generalization}. 
However, when a network trained on white contrast images (Section~\ref{sec:sbd_comparison}) is evaluated on the real data, it produces unnatural streak patterns in the bulk (see third row in Figure~\ref{fig:datashift}). When visually comparing this to the pattern in the bulk of the reference image computed by time averaging, it is evident that this is an artefact of denoising. This can be remedied by training the network on the more restricted subset of images described in Section~\ref{sec:exp_real_data} (see third row in Figure~\ref{fig:datashift}), whose imaging parameters are more suited to the real acquisition conditions. 
Since unsupervised denoising methods directly train on the real data, they do not suffer from this problem of mismatch between training and test data. The patterns recovered by unsupervised methods in the bulk are close to the estimated reference image (see Figure~\ref{fig:datashift}). However, as discussed in Section~\ref{sec:unsupervised}, they show significant artefacts on the surface of nanoparticle.

\section{Discussion and Conclusions}
Our case study is a proof of concept that CNNs trained on simulated data can be remarkably effective when applied to real imaging data. 
It provides several insights and suggests future research directions that are relevant, beyond electron microscopy, to other domains where the images of interest can be simulated, such as medical imaging \cite{kim2019performance, minarik2020denoising}, other types of microscopy~\cite{giannatou2019deep, vasudevan2019deep}, or astronomy~\cite{peterson2015simulation}.
We show that the design of the training dataset is critical, so an important question is how to design simulated training datasets in a principled systematic way. Answering it will require a deeper understanding of the generalization ability of CNNs with respect to variations in the statistics of the input images. We also demonstrate that architectures tailored to photographic imaging can perform poorly when applied to other data. Designing CNNs for other domains requires an understanding of the image features that are exploited for denoising. Gradient visualization is shown to be useful here, but more advanced visualization techniques are needed. In addition, we demonstrate that standard metrics used to quantify performance in photographs may not be sensitive to scientifically relevant features, and propose several new metrics to address this problem. Although SBD outperforms other methods by a large margin, some artefacts such as phantom atoms still appear. Our proposed likelihood maps help to flag such events, but may still fail to do so in regions of unusually low SNR. Developing more sophisticated methods for uncertainty quantification is therefore a key research direction. It would also be of great interest to develop unsupervised or self-supervised denoising approaches that are effective with small amounts of data at low SNRs. Finally, to encourage further development of deep-learning methodologies for scientific imaging, we release a denoising benchmark dataset of TEM images, containing 18,000 examples.

\section*{Acknowledgements}
We gratefully acknowledge financial support from the National Science Foundation (NSF). NSF NRT HDR Award 1922658 partially supported SM who designed, implemented and analyzed the proposed methodology. NSF CBET 1604971 supported JLV and PAC who acquired and processed the experimental data. NSF OAC-1940263 supported RM and PAC who created the training dataset. NSF OAC-1940124 and CCF-1934985 supported BT, who contributed to implement the methodology and conduct the experiments, and DM, who supervised the design of the methodology. NSF OAC-1940097 supported CFG who supervised the design, implementation and analysis of the methodology. All authors participated in writing and reviewing the manuscript. The authors acknowledge ASU Research Computing and NYU HPC for providing high performance computing resources, and the John M. Cowley Center for High Resolution Electron Microscopy at Arizona State University.

\bibliographystyle{siamplain}
\bibliography{ref}

\begin{thebibliography}{10}

\bibitem{Agustsson_2017_CVPR_Workshops}
{\sc E.~Agustsson and R.~Timofte}, {\em Ntire 2017 challenge on single image
  super-resolution: Dataset and study}, in The IEEE Conference on Computer
  Vision and Pattern Recognition (CVPR) Workshops, July 2017.

\bibitem{Cryo2016}
{\sc X.-C. Bai, G.~McMullan, and S.~H. Scheres}, {\em How cryo-em is
  revolutionizing structural biology}, Trends in Biochemical Sciences, 40
  (2015), pp.~49 -- 57,
  \url{https://doi.org/https://doi.org/10.1016/j.tibs.2014.10.005},
  \url{http://www.sciencedirect.com/science/article/pii/S096800041400187X}.

\bibitem{barford1967experimental}
{\sc N.~C. Barford}, {\em Experimental measurements: precision, error and
  truth},  (1967).

\bibitem{barthel2018dr}
{\sc J.~Barthel}, {\em Dr. probe: A software for high-resolution stem image
  simulation}, Ultramicroscopy, 193 (2018), pp.~1--11.

\bibitem{batson2019noise2self}
{\sc J.~Batson and L.~Royer}, {\em Noise2self: Blind denoising by
  self-supervision}, arXiv preprint arXiv:1901.11365,  (2019).

\bibitem{beckouche2013astronomical}
{\sc S.~Beckouche, J.-L. Starck, and J.~Fadili}, {\em Astronomical image
  denoising using dictionary learning}, Astronomy \& Astrophysics, 556 (2013),
  p.~A132.

\bibitem{benjamini2010simultaneous}
{\sc Y.~Benjamini}, {\em Simultaneous and selective inference: Current
  successes and future challenges}, Biometrical Journal, 52 (2010),
  pp.~708--721.

\bibitem{benjamini1995controlling}
{\sc Y.~Benjamini and Y.~Hochberg}, {\em Controlling the false discovery rate:
  A practical and powerful approach to multiple testing}, Journal of the Royal
  Statistical Society. Series B (Methodological), 57 (1995), pp.~289--300,
  \url{https://doi.org/10.2307/2346101},
  \url{http://dx.doi.org/10.2307/2346101}.

\bibitem{bernal1998interpretation}
{\sc S.~Bernal, F.~Botana, J.~Calvino, C.~Lopez-Cartes, J.~Perez-Omil, and
  J.~Rodr{\i}guez-Izquierdo}, {\em The interpretation of hrem images of
  supported metal catalysts using image simulation: profile view images},
  Ultramicroscopy, 72 (1998), pp.~135--164.

\bibitem{buades2005non}
{\sc A.~Buades, B.~Coll, and J.-M. Morel}, {\em A non-local algorithm for image
  denoising}, in 2005 IEEE Computer Society Conference on Computer Vision and
  Pattern Recognition (CVPR'05), vol.~2, IEEE, 2005, pp.~60--65.

\bibitem{buchholz2019cryo}
{\sc T.-O. Buchholz, M.~Jordan, G.~Pigino, and F.~Jug}, {\em Cryo-care:
  content-aware image restoration for cryo-transmission electron microscopy
  data}, in 2019 IEEE 16th International Symposium on Biomedical Imaging (ISBI
  2019), IEEE, 2019, pp.~502--506.

\bibitem{chang2000adaptive}
{\sc S.~G. Chang, B.~Yu, and M.~Vetterli}, {\em Adaptive wavelet thresholding
  for image denoising and compression}, IEEE Trans. Image Processing, 9 (2000),
  pp.~1532--1546.

\bibitem{chen2016trainable}
{\sc Y.~Chen and T.~Pock}, {\em Trainable nonlinear reaction diffusion: A
  flexible framework for fast and effective image restoration}, IEEE
  transactions on pattern analysis and machine intelligence, 39 (2016),
  pp.~1256--1272.

\bibitem{crozier2019dynamics}
{\sc P.~A. Crozier, E.~L. Lawrence, J.~L. Vincent, and B.~D. Levin}, {\em
  Dynamic restructuring during processing: approaches to higher temporal
  resolution}, Microscopy and Microanalysis, 25 (2019), pp.~1464--1465.

\bibitem{ede2020deep}
{\sc J.~M. Ede}, {\em Deep learning in electron microscopy}, Machine Learning:
  Science and Technology,  (2020).

\bibitem{ede2019improving}
{\sc J.~M. Ede and R.~Beanland}, {\em Improving electron micrograph
  signal-to-noise with an atrous convolutional encoder-decoder},
  Ultramicroscopy, 202 (2019), pp.~18--25.

\bibitem{ercius2020}
{\sc P.~Ercius, I.~Johnson, H.~Brown, P.~Pelz, S.-L. Hsu, B.~Draney, E.~Fong,
  A.~Goldschmidt, J.~Joseph, J.~Lee, and et~al.}, {\em The 4d camera – a 87
  khz frame-rate detector for counted 4d-stem experiments}, Microscopy and
  Microanalysis,  (2020), p.~1–3,
  \url{https://doi.org/10.1017/S1431927620019753}.

\bibitem{FARUQI2018}
{\sc A.~Faruqi and G.~McMullan}, {\em Direct imaging detectors for electron
  microscopy}, Nuclear Instruments and Methods in Physics Research Section A:
  Accelerators, Spectrometers, Detectors and Associated Equipment, 878 (2018),
  pp.~180 -- 190,
  \url{https://doi.org/https://doi.org/10.1016/j.nima.2017.07.037},
  \url{http://www.sciencedirect.com/science/article/pii/S0168900217307787}.
\newblock Radiation Imaging Techniques and Applications.

\bibitem{giannatou2019deep}
{\sc E.~Giannatou, G.~Papavieros, V.~Constantoudis, H.~Papageorgiou, and
  E.~Gogolides}, {\em Deep learning denoising of sem images towards
  noise-reduced ler measurements}, Microelectronic Engineering, 216 (2019),
  p.~111051.

\bibitem{gong2018pet}
{\sc K.~Gong, J.~Guan, C.-C. Liu, and J.~Qi}, {\em Pet image denoising using a
  deep neural network through fine tuning}, IEEE Transactions on Radiation and
  Plasma Medical Sciences, 3 (2018), pp.~153--161.

\bibitem{Guo2020}
{\sc H.~Guo, P.~Sautet, and A.~N. Alexandrova}, {\em Reagent-triggered
  isomerization of fluxional cluster catalyst via dynamic coupling}, The
  Journal of Physical Chemistry Letters, 11 (2020), pp.~3089--3094,
  \url{https://doi.org/10.1021/acs.jpclett.0c00548},
  \url{https://doi.org/10.1021/acs.jpclett.0c00548},
  \url{https://arxiv.org/abs/https://doi.org/10.1021/acs.jpclett.0c00548}.
\newblock PMID: 32227852.

\bibitem{helor2008dscrim}
{\sc Y.~Hel-Or and D.~Shaked}, {\em A discriminative approach for wavelet
  denoising}, IEEE Transactions on Image Processing, 17 (2008), pp.~443--457.

\bibitem{horwath2020understanding}
{\sc J.~P. Horwath, D.~N. Zakharov, R.~Megret, and E.~A. Stach}, {\em
  Understanding important features of deep learning models for segmentation of
  high-resolution transmission electron microscopy images}, npj Computational
  Materials, 6 (2020), pp.~1--9.

\bibitem{ioffe2015batch}
{\sc S.~Ioffe and C.~Szegedy}, {\em Batch normalization: Accelerating deep
  network training by reducing internal covariate shift}, arXiv preprint
  arXiv:1502.03167,  (2015).

\bibitem{jiang2003applications}
{\sc W.~Jiang, M.~L. Baker, Q.~Wu, C.~Bajaj, and W.~Chiu}, {\em Applications of
  a bilateral denoising filter in biological electron microscopy}, Journal of
  structural biology, 144 (2003), pp.~114--122.

\bibitem{khademi2020self}
{\sc W.~Khademi, S.~Rao, C.~Minnerath, G.~Hagen, and J.~Ventura}, {\em
  Self-supervised poisson-gaussian denoising}, arXiv preprint arXiv:2002.09558,
   (2020).

\bibitem{kim2019performance}
{\sc B.~Kim, M.~Han, H.~Shim, and J.~Baek}, {\em A performance comparison of
  convolutional neural network-based image denoising methods: The effect of
  loss functions on low-dose ct images}, Medical physics, 46 (2019),
  pp.~3906--3923.

\bibitem{kingma2014adam}
{\sc D.~P. Kingma and J.~Ba}, {\em Adam: A method for stochastic optimization},
  arXiv preprint arXiv:1412.6980,  (2014).

\bibitem{kirkland2006image}
{\sc E.~J. Kirkland et~al.}, {\em Image simulation in transmission electron
  microscopy}, Cornell University, Ithaca,  (2006).

\bibitem{krull2019noise2void}
{\sc A.~Krull, T.-O. Buchholz, and F.~Jug}, {\em Noise2void-learning denoising
  from single noisy images}, in Proceedings of the IEEE Conference on Computer
  Vision and Pattern Recognition, 2019, pp.~2129--2137.

\bibitem{krull2019probabilistic}
{\sc A.~Krull, T.~Vicar, and F.~Jug}, {\em Probabilistic noise2void:
  Unsupervised content-aware denoising}, arXiv preprint arXiv:1906.00651,
  (2019).

\bibitem{laine2019high}
{\sc S.~Laine, T.~Karras, J.~Lehtinen, and T.~Aila}, {\em High-quality
  self-supervised deep image denoising}, in Advances in Neural Information
  Processing Systems, 2019, pp.~6970--6980.

\bibitem{lawrence_levin_miller_crozier_2020}
{\sc E.~L. Lawrence, B.~D. Levin, B.~K. Miller, and P.~A. Crozier}, {\em
  Approaches to exploring spatio-temporal surface dynamics in nanoparticles
  with in situ transmission electron microscopy}, Microscopy and Microanalysis,
  26 (2020), p.~86–94, \url{https://doi.org/10.1017/S1431927619015228}.

\bibitem{lecun2015deep}
{\sc Y.~LeCun, Y.~Bengio, and G.~Hinton}, {\em Deep learning}, nature, 521
  (2015), p.~436.

\bibitem{lehtinen2018noise2noise}
{\sc J.~Lehtinen, J.~Munkberg, J.~Hasselgren, S.~Laine, T.~Karras, M.~Aittala,
  and T.~Aila}, {\em Noise2noise: Learning image restoration without clean
  data}, arXiv preprint arXiv:1803.04189,  (2018).

\bibitem{LEVIN2020}
{\sc B.~D. Levin, E.~L. Lawrence, and P.~A. Crozier}, {\em Tracking the
  picoscale spatial motion of atomic columns during dynamic structural change},
  Ultramicroscopy, 213 (2020), p.~112978,
  \url{https://doi.org/https://doi.org/10.1016/j.ultramic.2020.112978},
  \url{http://www.sciencedirect.com/science/article/pii/S0304399119303122}.

\bibitem{lichtman2005fluorescence}
{\sc J.~W. Lichtman and J.-A. Conchello}, {\em Fluorescence microscopy}, Nature
  methods, 2 (2005), pp.~910--919.

\bibitem{lim1990two}
{\sc J.~S. Lim}, {\em Two-dimensional signal and image processing}, ph,
  (1990).

\bibitem{lindeberg2013scale}
{\sc T.~Lindeberg}, {\em Scale selection properties of generalized scale-space
  interest point detectors}, Journal of Mathematical Imaging and vision, 46
  (2013), pp.~177--210.

\bibitem{lusier2007SURE}
{\sc F.~Luisier, T.~Blu, and M.~Unser}, {\em A new sure approach to image
  denoising: Interscale orthonormal wavelet thresholding}, IEEE Transactions on
  Image Processing, 16 (2007), pp.~593--606.

\bibitem{luisier2010image}
{\sc F.~Luisier, T.~Blu, and M.~Unser}, {\em Image denoising in mixed
  poisson--gaussian noise}, IEEE Transactions on image processing, 20 (2010),
  pp.~696--708.

\bibitem{madsen2018deep}
{\sc J.~Madsen, P.~Liu, J.~Kling, J.~B. Wagner, T.~W. Hansen, O.~Winther, and
  J.~Schi{\o}tz}, {\em A deep learning approach to identify local structures in
  atomic-resolution transmission electron microscopy images}, Advanced Theory
  and Simulations, 1 (2018), p.~1800037.

\bibitem{makitalo2012optimal}
{\sc M.~Makitalo and A.~Foi}, {\em Optimal inversion of the generalized
  anscombe transformation for poisson-gaussian noise}, IEEE transactions on
  image processing, 22 (2012), pp.~91--103.

\bibitem{manifold2019denoising}
{\sc B.~Manifold, E.~Thomas, A.~T. Francis, A.~H. Hill, and D.~Fu}, {\em
  Denoising of stimulated raman scattering microscopy images via deep
  learning}, Biomedical optics express, 10 (2019), pp.~3860--3874.

\bibitem{mclean2008electronic}
{\sc I.~S. McLean}, {\em Electronic imaging in astronomy: detectors and
  instrumentation}, Springer Science \& Business Media, 2008.

\bibitem{meiniel2018denoising}
{\sc W.~Meiniel, J.-C. Olivo-Marin, and E.~D. Angelini}, {\em Denoising of
  microscopy images: a review of the state-of-the-art, and a new sparsity-based
  method}, IEEE Transactions on Image Processing, 27 (2018), pp.~3842--3856.

\bibitem{milanfar2012tour}
{\sc P.~Milanfar}, {\em A tour of modern image filtering: New insights and
  methods, both practical and theoretical}, IEEE signal processing magazine, 30
  (2012), pp.~106--128.

\bibitem{minarik2020denoising}
{\sc D.~Minarik, O.~Enqvist, and E.~Tr{\"a}g{\aa}rdh}, {\em Denoising of
  scintillation camera images using a deep convolutional neural network: a
  monte carlo simulation approach}, Journal of Nuclear Medicine, 61 (2020),
  pp.~298--303.

\bibitem{Mohan2020Robust}
{\sc S.~Mohan, Z.~Kadkhodaie, E.~P. Simoncelli, and C.~Fernandez-Granda}, {\em
  Robust and interpretable blind image denoising via bias-free convolutional
  neural networks}, in International Conference on Learning Representations,
  2020, \url{https://openreview.net/forum?id=HJlSmC4FPS}.

\bibitem{montavon2017explaining}
{\sc G.~Montavon, S.~Lapuschkin, A.~Binder, W.~Samek, and K.-R. M{\"u}ller},
  {\em Explaining nonlinear classification decisions with deep taylor
  decomposition}, Pattern Rec., 65 (2017), pp.~211--222.

\bibitem{montini2016fundamentals}
{\sc T.~Montini, M.~Melchionna, M.~Monai, and P.~Fornasiero}, {\em Fundamentals
  and catalytic applications of ceo2-based materials}, Chemical reviews, 116
  (2016), pp.~5987--6041.

\bibitem{nellist1998accurate}
{\sc P.~Nellist and S.~Pennycook}, {\em Accurate structure determination from
  image reconstruction in adf stem}, Journal of Microscopy, 190 (1998),
  pp.~159--170.

\bibitem{nie2015recent}
{\sc Y.~Nie, L.~Li, and Z.~Wei}, {\em Recent advancements in pt and pt-free
  catalysts for oxygen reduction reaction}, Chemical Society Reviews, 44
  (2015), pp.~2168--2201.

\bibitem{peterson2015simulation}
{\sc J.~Peterson, J.~Jernigan, S.~Kahn, A.~Rasmussen, E.~Peng, Z.~Ahmad,
  J.~Bankert, C.~Chang, C.~Claver, D.~Gilmore, et~al.}, {\em Simulation of
  astronomical images from optical survey telescopes using a comprehensive
  photon monte carlo approach}, The Astrophysical Journal Supplement Series,
  218 (2015), p.~14.

\bibitem{portilla2003image}
{\sc J.~Portilla, V.~Strela, M.~J. Wainwright, and E.~P. Simoncelli}, {\em
  Image denoising using scale mixtures of gaussians in the wavelet domain},
  IEEE Trans. Image Processing, 12 (2003).

\bibitem{prakash2020fully}
{\sc M.~Prakash, M.~Lalit, P.~Tomancak, A.~Krul, and F.~Jug}, {\em Fully
  unsupervised probabilistic noise2void}, in 2020 IEEE 17th International
  Symposium on Biomedical Imaging (ISBI), IEEE, 2020, pp.~154--158.

\bibitem{preparata1985convex}
{\sc F.~P. Preparata and M.~I. Shamos}, {\em Convex hulls: Basic algorithms},
  in Computational geometry, Springer, 1985, pp.~95--149.

\bibitem{self2self}
{\sc Y.~Quan, M.~Chen, T.~Pang, and H.~Ji}, {\em Self2self with dropout:
  Learning self-supervised denoising from single image}, in Proceedings of the
  IEEE/CVF Conference on Computer Vision and Pattern Recognition, 2020,
  pp.~1890--1898.

\bibitem{ragone2020atomic}
{\sc M.~Ragone, V.~Yurkiv, B.~Song, A.~Ramsubramanian, R.~Shahbazian-Yassar,
  and F.~Mashayek}, {\em Atomic column heights detection in metallic
  nanoparticles using deep convolutional learning}, Computational Materials
  Science, 180 (2020), p.~109722.

\bibitem{raphan2008optimal}
{\sc M.~Raphan and E.~P. Simoncelli}, {\em Optimal denoising in redundant
  representations}, IEEE Transactions on image processing, 17 (2008),
  pp.~1342--1352.

\bibitem{ronneberger2015u}
{\sc O.~Ronneberger, P.~Fischer, and T.~Brox}, {\em U-net: Convolutional
  networks for biomedical image segmentation}, in International Conference on
  Medical image computing and computer-assisted intervention, Springer, 2015,
  pp.~234--241.

\bibitem{udvd}
{\sc D.~Y. Sheth, S.~Mohan, J.~L. Vincent, R.~Manzorro, P.~A. Crozier, M.~M.
  Khapra, E.~P. Simoncelli, and C.~Fernandez-Granda}, {\em Unsupervised deep
  video denoising}, arXiv preprint arXiv:2011.15045,  (2020).

\bibitem{simonyan2013deep}
{\sc K.~Simonyan, A.~Vedaldi, and A.~Zisserman}, {\em Deep inside convolutional
  networks: Visualising image classification models and saliency maps}, arXiv
  preprint arXiv:1312.6034,  (2013).

\bibitem{Smith2015}
{\sc D.~Smith}, {\em CHAPTER 1: Characterization of nanomaterials using
  transmission electron microscopy}, no.~37 in RSC Nanoscience and
  Nanotechnology, Royal Society of Chemistry, 37~ed., Jan. 2015, pp.~1--29,
  \url{https://doi.org/10.1039/9781782621867-00001}.

\bibitem{Sun2020}
{\sc G.~Sun, A.~N. Alexandrova, and P.~Sautet}, {\em Structural rearrangements
  of subnanometer cu oxide clusters govern catalytic oxidation}, ACS Catalysis,
  10 (2020), pp.~5309--5317, \url{https://doi.org/10.1021/acscatal.0c00824},
  \url{https://doi.org/10.1021/acscatal.0c00824},
  \url{https://arxiv.org/abs/https://doi.org/10.1021/acscatal.0c00824}.

\bibitem{suveer2019super}
{\sc A.~Suveer, A.~Gupta, G.~Kylberg, and I.-M. Sintorn}, {\em Super-resolution
  reconstruction of transmission electron microscopy images using deep
  learning}, in 2019 IEEE 16th International Symposium on Biomedical Imaging
  (ISBI 2019), IEEE, 2019, pp.~548--551.

\bibitem{Crozier-insitu2016}
{\sc F.~Tao and P.~Crozier}, {\em Atomic-scale observations of catalyst
  structures under reaction conditions and during catalysis}, Chemical Reviews,
  116 (2016), pp.~3487--3539, \url{https://doi.org/10.1021/cr5002657}.

\bibitem{tomasi1998bilateral}
{\sc C.~Tomasi and R.~Manduchi}, {\em Bilateral filtering for gray and color
  images.}, in ICCV, vol.~98, 1998.

\bibitem{vasudevan2019deep}
{\sc R.~K. Vasudevan and S.~Jesse}, {\em Deep learning as a tool for image
  denoising and drift correction}, Microscopy and Microanalysis, 25 (2019),
  pp.~190--191.

\bibitem{vincent2021developing}
{\sc J.~L. Vincent, R.~Manzorro, S.~Mohan, B.~Tang, D.~Y. Sheth, E.~P.
  Simoncelli, D.~S. Matteson, C.~Fernandez-Granda, and P.~A. Crozier}, {\em
  Developing and evaluating deep neural network-based denoising for
  nanoparticle tem images with ultra-low signal-to-noise},  (2021),
  \url{https://arxiv.org/abs/2101.07770}.

\bibitem{wang2004image}
{\sc Z.~Wang, A.~C. Bovik, H.~R. Sheikh, and E.~P. Simoncelli}, {\em Image
  quality assessment: from error visibility to structural similarity}, IEEE
  transactions on image processing, 13 (2004), pp.~600--612.

\bibitem{wang2003new}
{\sc Z.~L. Wang}, {\em New developments in transmission electron microscopy for
  nanotechnology}, Advanced Materials, 15 (2003), pp.~1497--1514.

\bibitem{wiener1950extrapolation}
{\sc N.~Wiener}, {\em Extrapolation, interpolation, and smoothing of stationary
  time series: with engineering applications}, Technology Press, 1950.

\bibitem{yu2012review}
{\sc W.~Yu, M.~D. Porosoff, and J.~G. Chen}, {\em Review of pt-based bimetallic
  catalysis: from model surfaces to supported catalysts}, Chemical reviews, 112
  (2012), pp.~5780--5817.

\bibitem{zhang2017beyond}
{\sc K.~Zhang, W.~Zuo, Y.~Chen, D.~Meng, and L.~Zhang}, {\em Beyond a
  {Gaussian} denoiser: Residual learning of deep {CNN} for image denoising},
  IEEE Transactions on Image Processing, 26 (2017), pp.~3142--3155.

\bibitem{zhang2018dynamically}
{\sc X.~Zhang, Y.~Lu, J.~Liu, and B.~Dong}, {\em Dynamically unfolding
  recurrent restorer: A moving endpoint control method for image restoration},
  arXiv preprint arXiv:1805.07709,  (2018).

\bibitem{zhang2019poisson}
{\sc Y.~Zhang, Y.~Zhu, E.~Nichols, Q.~Wang, S.~Zhang, C.~Smith, and S.~Howard},
  {\em A poisson-gaussian denoising dataset with real fluorescence microscopy
  images}, in Proceedings of the IEEE Conference on Computer Vision and Pattern
  Recognition, 2019, pp.~11710--11718.

\bibitem{zhu2015survey}
{\sc H.~J. Zhu, B.~C. Han, and B.~Qiu}, {\em Survey of astronomical image
  processing methods}, in International Conference on Image and Graphics,
  Springer, 2015, pp.~420--429.

\bibitem{book}
{\sc J.-M. Zuo and J.~Spence}, {\em Advanced Transmission Electron Microscopy,
  Imaging and Diffraction in Nanoscience}, 01 2017.

\end{thebibliography}

\appendix

\section{Data simulation}

\subsection{Simulation process}
\label{sec:data_simulation}
The simulated TEM image dataset was generated using the multi-slice image simulation method, as implemented in the Dr. Probe software package~\cite{barthel2018dr}. In the multi-slice approach, the modeled specimen is sectioned into many thin slices (here they are 0.167 Angstroms thick), and quantum mechanical calculations are performed to simulate the incident electron wave function propagating through and interacting with each slice of the material \cite{kirkland2006image}. The resultant wave function exiting the last slice is then convolved with a point spread function that emulates the effect of imaging it in the electron microscope. All of the image simulations were performed using an accelerating voltage of 300 kV with a beam convergence angle of 0.2 mrad and a focal spread of 4 nm. The third-order spherical aberration coefficient ($C_s$) was set to be -13 $\mu$m. The fifth-order spherical aberration coefficient ($C$5) was set as 5 mm. All other aberrations (e.g., 2-fold and 3-fold astigmatism, coma, star aberration, etc.) were approximated to be negligible. The defocus ($C$1) was varied systematically between 0 nm and 20 nm, as discussed below. Image calculations were computed using a non-linear model including partial temporal coherence by explicit averaging and partial spatial coherence, which is treated by a quasi-coherent approach with a dampening envelope applied to the wave function. An isotropic vibration envelope of 50 pm was applied during the image calculation. Images were simulated with 1024 x 1024 pixels and then later binned to desired sizes to match the pixel size of the experimentally acquired image series. Finally, to equate the intensity range of the simulated images with those acquired experimentally, the intensities of the simulated images were scaled by a factor which equalized the vacuum intensity in a single simulation to the average intensity measured over a large area of the vacuum in a single 0.025 second experimental frame (i.e., 0.45 counts per pixel in the vacuum region).

\subsection{Experimental parameters}
\label{sec:experimental_parameters}
\begin{figure}
    \centering
    \includegraphics[width=\linewidth]{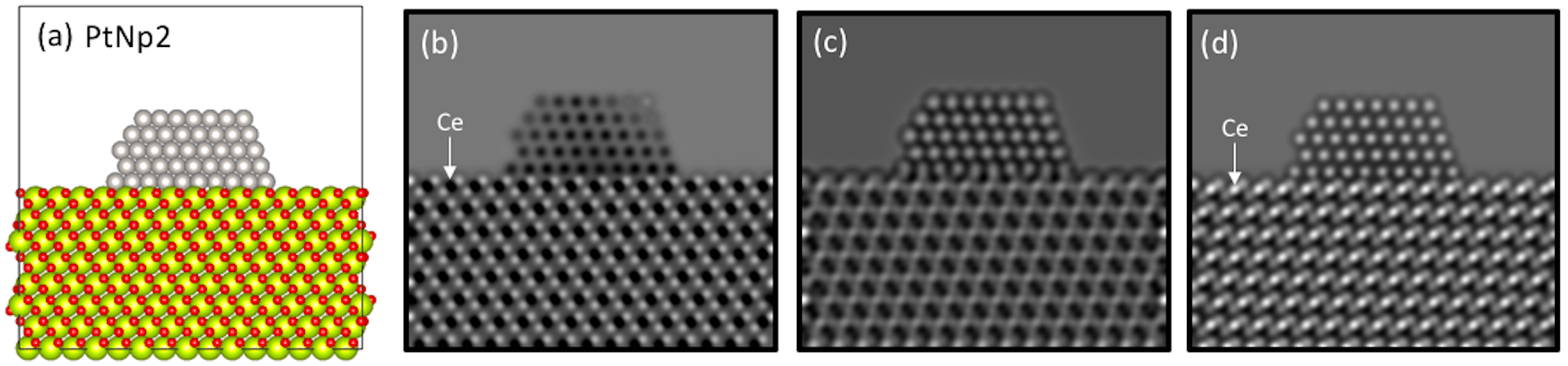}
    \caption{\textbf{Demonstration of contrast reversal with changes in defocus.} (a) Image of the Pt/CeO$_2$ atomic structural model.(b) to (d) Simulated images  under different electron-optical focusing conditions, emphasizing variations on the Ce and Pt column contrast. In (b), the image shows black contrast for both Ce and Pt columns. In (c), the Pt columns reverse contrast and now appear white, while Ce columns become challenging to discriminate. Finally, in (d) all of the atomic columns appear with white contrast.}
    \label{fig:schematic_contrast}
\end{figure}

\begin{figure}
    \centering
    \includegraphics[width=\linewidth]{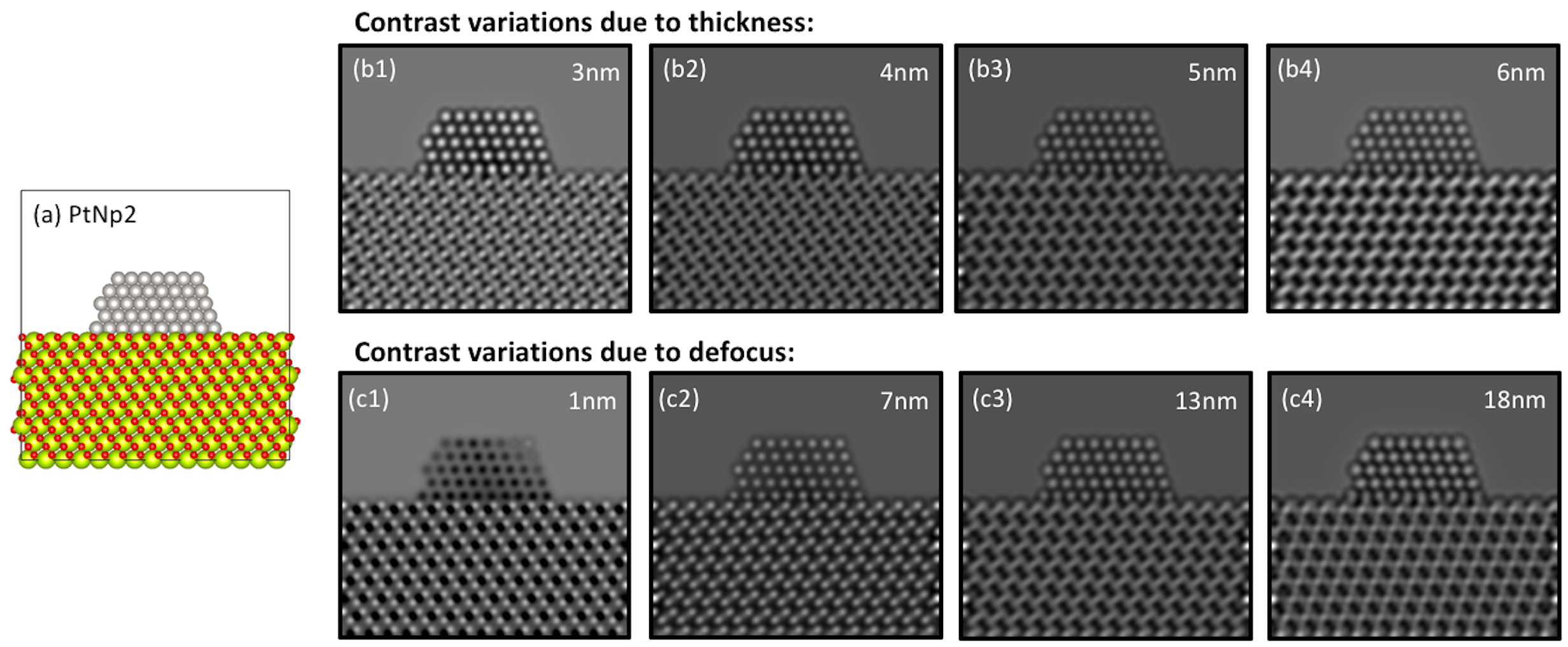}
    \caption{\textbf{Image contrast variations due to thickness and defocus.} (a) Image of the Pt/CeO$_2$ atomic structural model. (b1) to (b4) Simulated images at a defocus value of 13 nm, where the variations of the contrast are due to the thickness of the model, increased from 3 nm to 6nm. (c1) to (c4) Contrast variations on simulated images due to defocus: the thickness of the model has been kept constant at 5 nm and the defocus has been tuned to 1 nm, 7 nm, 13 nm, and 18 nm.}
    \label{fig:schematic_thickness_defocus}
\end{figure}

In phase-contrast TEM imaging (the technique employed here), multiple electron-optical and specimen parameters can give rise to complex, non-linear modulations of the image contrast. These parameters can include the objective lens defocus, the specimen thickness, the orientation of the specimen, and its crystallographic shape/structure. Due to the complex image formation mechanisms, atomic columns of the same material imaged may appear black or white (or somewhere in between, i.e., intermediate) depending on the exact combination of these various factors. Examples of the type of contrast reversal that may occur for a static structure imaged at constant thickness and tilt are given in Figure \ref{fig:schematic_contrast}. Additionally, images showing the type of contrast variations that may occur when the support thickness is changed, and how these compare to those which arise from changes in defocus are given in Figure \ref{fig:schematic_thickness_defocus}. 

\begin{figure}
\centering
\includegraphics[width=1\linewidth]{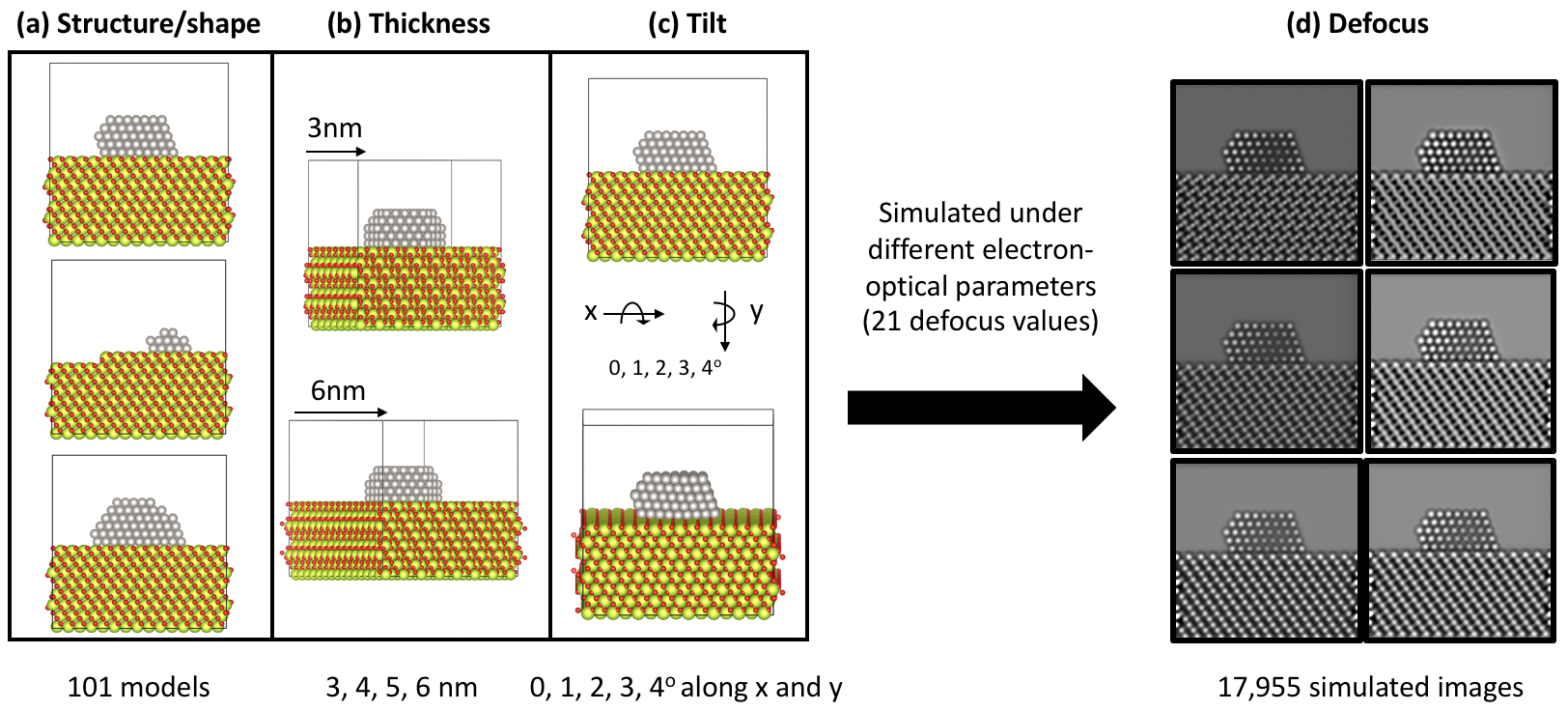}
\caption{\textbf{Summary of parameters considered during the modelling and image simulation processes.} Subset of Pt/CeO$_2$ atomic models presenting variations on the (a) structure and shape of the nanoparticle and the support, (b) the thickness of the CeO$_2$ slab and (c) the tilt of the atomic models. Color code for the models matches Pt, Ce and O with grey, yellow and red atoms respectively. (d) Simulated images under different defocus values.}
\label{fig:simulation}
\end{figure}

Here, we systematically varied these parameters to generate a large number of cases (approximately 18,000), corresponding to potential combinations that may arise during a real experiment (see Figure \ref{fig:simulation}). First, around 100 atomic-scale structural models of CeO$_2$-supported Pt nanoparticles were generated. Each model represents Pt nanoparticles of various size, shape and atomic structure (e.g., small, medium, or large size, with either faceted or defected surfaces, or some combination of both), supported on CeO$_2$, which itself may present either a faceted surface or one characterized by surface defects. Secondly, the thickness of the CeO$_2$ support was varied from 3 nm to 6 nm along 1 nm increments. One aspect to note is that the thickness variation is not equally applied to each of the aforementioned models. Third, each resultant model was tilted from 0$^{\circ}$ to 4$^{\circ}$ about the \textit{x} and \textit{y} axes independently in increments of 1$^{\circ}$. Thus, variations from 0$^{\circ}$ in \textit{x} and 0$^{\circ}$ in \textit{y}, to 4$^{\circ}$ in \textit{x} and 0$^{\circ}$ in \textit{y}, or 0$^{\circ}$ in \textit{x} and 4$^{\circ}$ in \textit{y} were considered. The final parameter systematically varied in the simulated image dataset was the electron optical defocus. Every model containing a unique shape/structure, thickness, and tilt (855 total) was imaged under a range of defocus values which often arise experimentally. Namely, the defocus was varied from 0 nm to 20 nm, along increments of 1 nm. Considering all combinations of the varied parameters, a total of 17,955 simulated images were generated for training and testing the neural network.

\subsection{Description of nanoparticle structures}
\label{sec:nanoparticle_structures}
The 3D atomic structural models utilized in this work consist of faceted Pt nanoparticles that oriented in a [110] zone axis and that are supported on a CeO$_2$ (111) surface which is itself oriented in the [110] zone axis. This crystallographic configuration corresponds to that which is often observed experimentally and is thus the focus of the current work. All of the models have been constructed with the freely available Rhodius software \cite{bernal1998interpretation}. Each model consists of a supercell having \textit{x} and \textit{y} dimensions of 5 nm x 5 nm. As discussed above, the support thickness was systematically varied for each model, and so the supercell's \textit{z} dimension varies between 3 nm and 6 nm, depending on the thickness of the particular model.

\begin{figure}
    \centering
    \includegraphics[width=\linewidth]{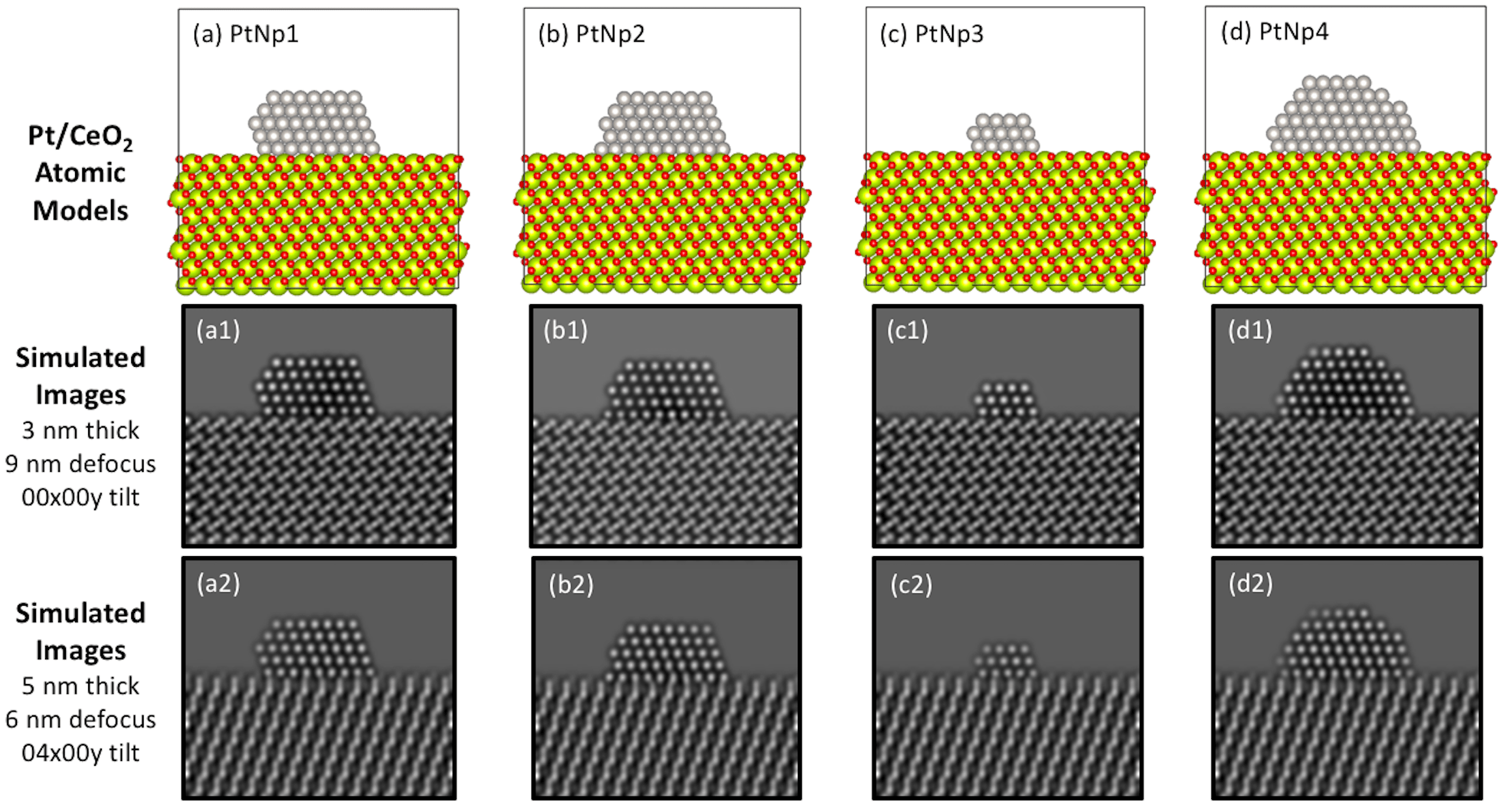}
    \caption{\textbf{Variations in the structure/size of the supported Pt nanoparticle.}(a) to (d) Atomic models of Pt nanoparticles (grey atoms) with different shapes and supported over a CeO$_2$ slab (yellow and red atoms respectively). (a1) to (d1) Simulated images depicting the described atomic models, considering a thickness of 3 nm, 9 nm of defocus and no tilt on the model, whereas (a2) to (d2) illustrate the same model under different conditions: 5 nm thickness, 6 nm defocus and 4 degrees tilted along x axis. All the simulated images present a $Cs$ value of -13 $\mu$m.}
    \label{fig:schematic_structure}
\end{figure}

\begin{figure}
    \centering
    \includegraphics[width=\linewidth]{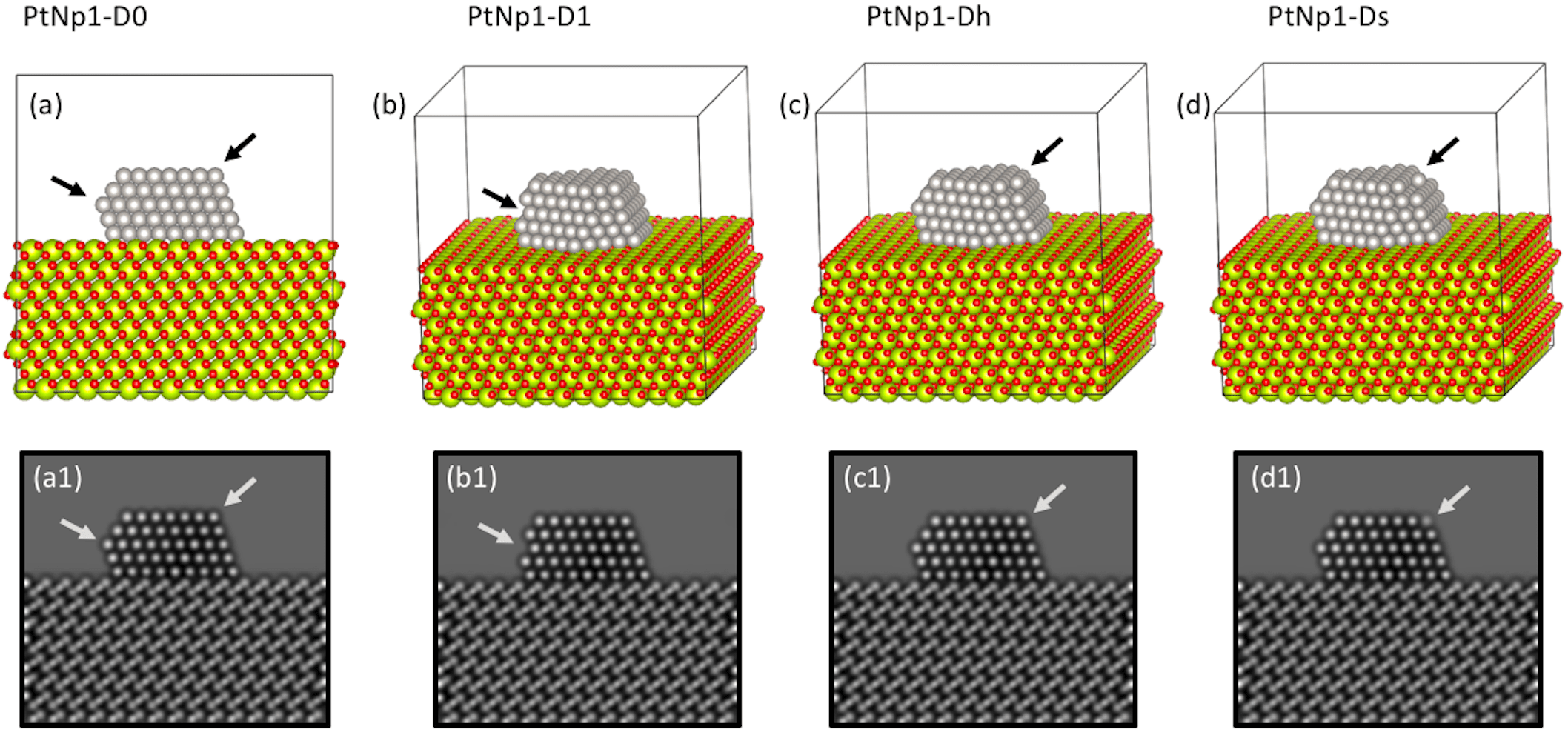}
    \caption{\textbf{Variations in the defects of the Pt surface structure.} (a) Atomic model of a CeO$_2$-supported Pt nanoparticle without any defects. The surface has been modified by (b) removing a full atomic column, (c) removing half of the occupancy and (d) keeping a single atom. Black arrows point the sites where these defects are taking place. Models (b), (c) and (d) have been slightly tilted to observe these modifications. (a1) to (d1) Simulated images of the presented atomic columns considering a 3nm thickness, 9 nm defocus and no tilt.}
    \label{fig:schematic_defect}
\end{figure}

While imaging these materials systems, experimentalists often aspire to visualize atomic-level structural rearrangements that can occur at the surfaces of the supported nanoparticles. Additionally, there are many millions of nanoparticles on a typical TEM sample, and the specific atomic-scale structural features comprising the surfaces of those imaged during an experiment may vary slightly from nanoparticle to nanoparticle. In order to encompass such complexity in the training dataset, a variety of Pt nanoparticles of multiple sizes/shape and surface defect character were incorporated into the 3D models. For example, four such models of CeO$_2$-supported Pt nanoparticles having various size and shape are shown in parts (a) to (d) of Figure \ref{fig:schematic_structure}. The multi-slice TEM image simulations generated from the models are shown below each for two different conditions, namely in parts (a1) to (d1), images are shown for a case in which the support is 3 nm thick, the defocus is 9 nm, and the tilt is 0$^{\circ}$ in \textit{x} and 0$^{\circ}$ in \textit{y}; in parts (a2) to (d2), images are shown for the case in which the support is 5 nm thick, the defocus is 6 nm, and the tilt is 4$^{\circ}$ in \textit{x} and 0$^{\circ}$ in \textit{y}. Furthermore, the surface character of the Pt nanoparticles was varied by altering the defect structure at different surface sites. A few examples are depicted in Figure \ref{fig:schematic_defect}. Here, in part (a), a CeO$_2$-supported Pt nanoparticle with faceted surfaces is shown; directly beneath it in (a1) is an image simulated under conditions in which the support is 3 nm thick, the defocus is 9nm, and there is no tilt. The arrowed sites designate locations on the Pt surface that have been subsequently altered. In part (b), the surface has been modified by removing a full atomic column from the arrowed location. In part (c), the occupancy of the arrowed corner site has been reduced by half. And in part (d), the occupancy of the arrowed corner site has been further reduced to a single atom. Parts (b1) - (d1) show the images simulated from these respective structures under the same imaging condition. Note that the surface sites altered in the structure correspond to high-energy sites (e.g., corners and edges) which are more likely to dynamically rearrange or show variation than, say, a low-energy terrace site located in the middle of the surface.

\section{Description of denoising CNNs}
\label{sec:architectures}
In this section we describe the CNN architectures and training procedure used for our computational experiments in more detail. 

\subsection{Proposed Architecture: UNet with large field of view}
\label{sec:unet}
We propose to use a modified version of UNet \cite{ronneberger2015u} with $n=4,6$ scales to achieve a large field of view. The network consisting of $n$ \texttt{down-block}s and $n$ \texttt{up-block}s. A \texttt{down-block} consists of a max-pooling layer, which reduces the spatial-dimension by half, followed by a \texttt{conv-block}. Similarly, an \texttt{up-block} consists of bilinear upsampling, which enlarges the size of the feature-map by a factor of two, followed by \texttt{conv-block}. A \texttt{conv-block} consists of conv-BN-ReLU-conv-BN-ReLU, where conv represents a convolutional layer and BN stands for batch normalization \cite{ioffe2015batch}. In our final model, we use $128$ channels in each layer of \texttt{conv-block} and $n=6$ scales. 

\subsection{DnCNN}
\label{sec:dncnn}
DnCNN~\cite{zhang2017beyond} consists of $20$ convolutional layers, each consisting of $3 \times 3$ filters and $64$ channels, batch normalization~\cite{ioffe2015batch}, and a ReLU nonlinearity. It has a skip connection from the initial layer to the final layer, which has no nonlinear units.

\subsection{Small UNet from DURR}
\label{sec:Small UNet}

We use the UNet proposed for the restoration module in DURR ~\cite{zhang2018dynamically}. The architecture consists of the following:
\begin{enumerate}
    \item \emph{conv1} - Takes in input image and maps to $32$ channels with $5 \times 5$ convolutional kernels.
    \item \emph{conv2} - Input: $32$ channels. Output: $32$ channels. $3 \times 3$ convolutional kernels. 
    \item \emph{conv3} -  Input: $32$ channels. Output: $64$ channels. $3 \times 3$ convolutional kernels with stride 2.
    \item \emph{conv4}-  Input: $64$ channels. Output: $64$ channels. $3 \times 3$ convolutional kernels.
    \item \emph{conv5}-  Input: $64$ channels. Output: $64$ channels. $3 \times 3$ convolutional kernels with dilation factor of 2.
    \item \emph{conv6}-  Input: $64$ channels. Output: $64$ channels. $3 \times 3$ convolutional kernels with dilation factor of 4.
    \item \emph{conv7}-  Transpose Convolution layer. Input: $64$ channels. Output: $64$ channels. $4 \times 4$ filters with stride $2$.
    \item \emph{conv8}-  Input: $96$ channels. Output: $64$ channels. $3 \times 3$ convolutional kernels. The input to this layer is the concatenation of the outputs of layer \emph{conv7} and \emph{conv2}.
    \item \emph{conv9}-  Input: $32$ channels. Output: $1$ channels. $5 \times 5$ convolutional kernels.
\end{enumerate}

 This configuration of UNet assumes even width and height, so we remove one row or column from images in with odd height or width.

  \section{Additional Results}
\begin{figure*}
    \centering
    \begin{tabular}{c@{\hskip 0.01in}c@{\hskip 0.01in}c@{\hskip 0.01in}c@{\hskip 0.01in}c}
    \footnotesize{Noisy} & \footnotesize{WF} & \footnotesize{LPF} & \footnotesize{VST+NLM} & \footnotesize{VST+BM3D} \\
    \includegraphics[width=1.05in]{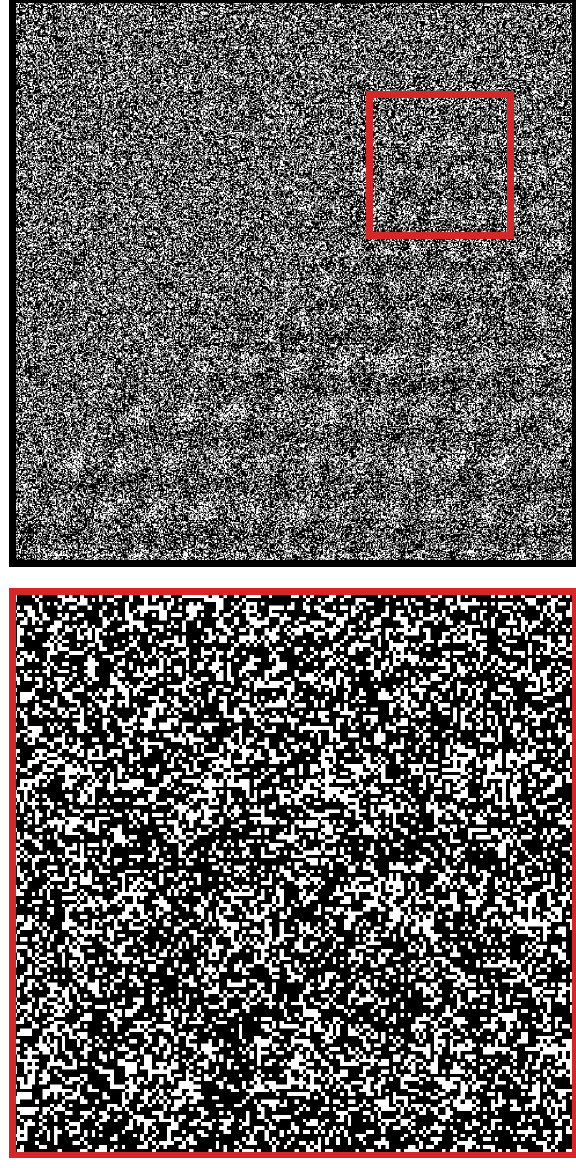}&
    \includegraphics[width=1.05in]{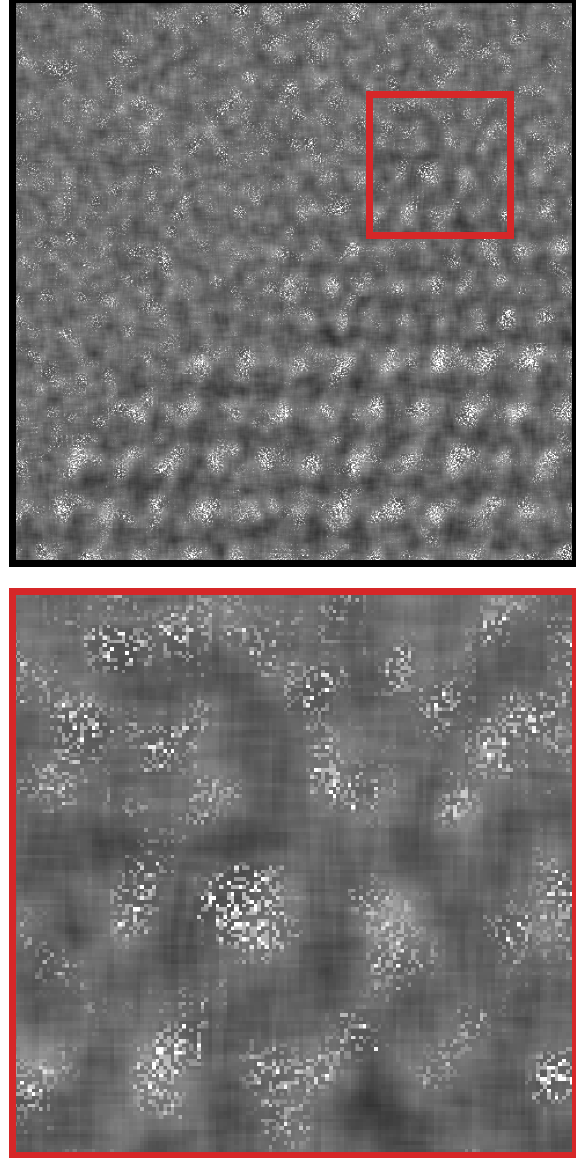}&
    \includegraphics[width=1.05in]{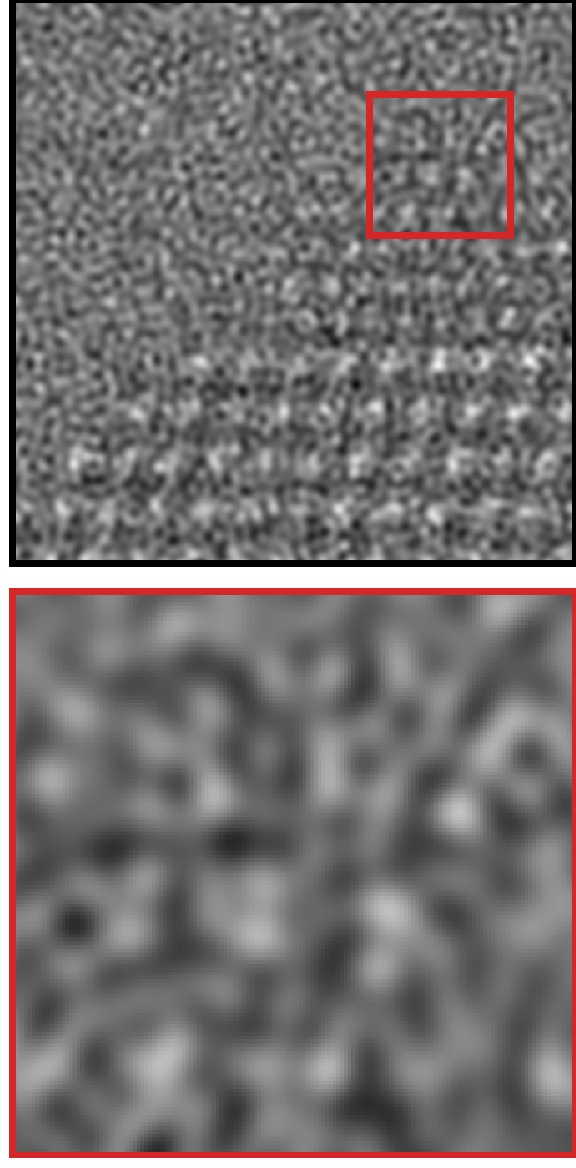}&
    \includegraphics[width=1.05in]{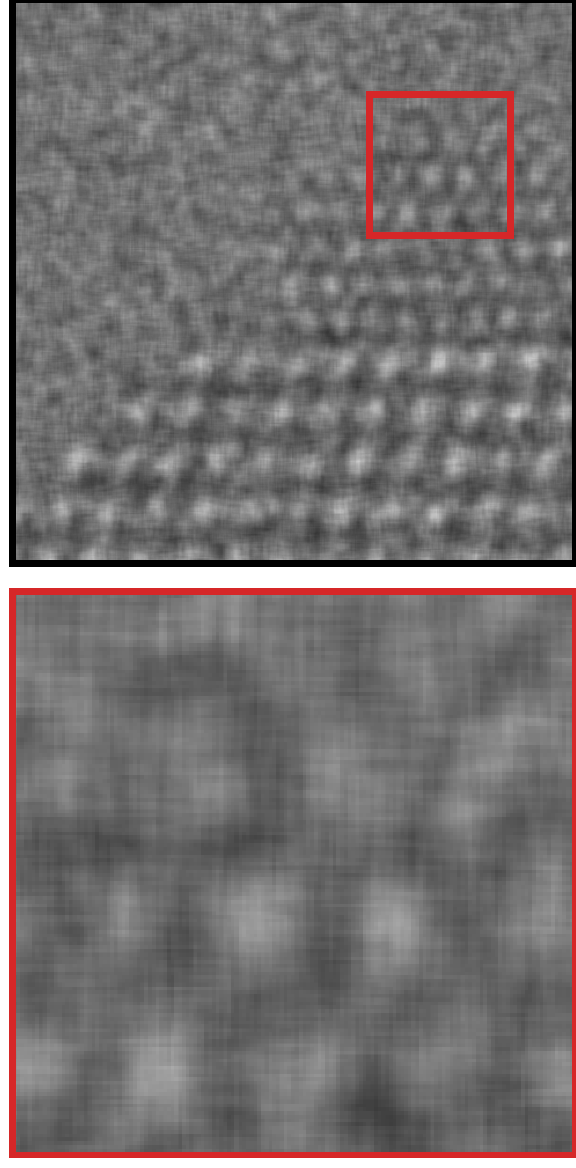}&
    \includegraphics[width=1.05in]{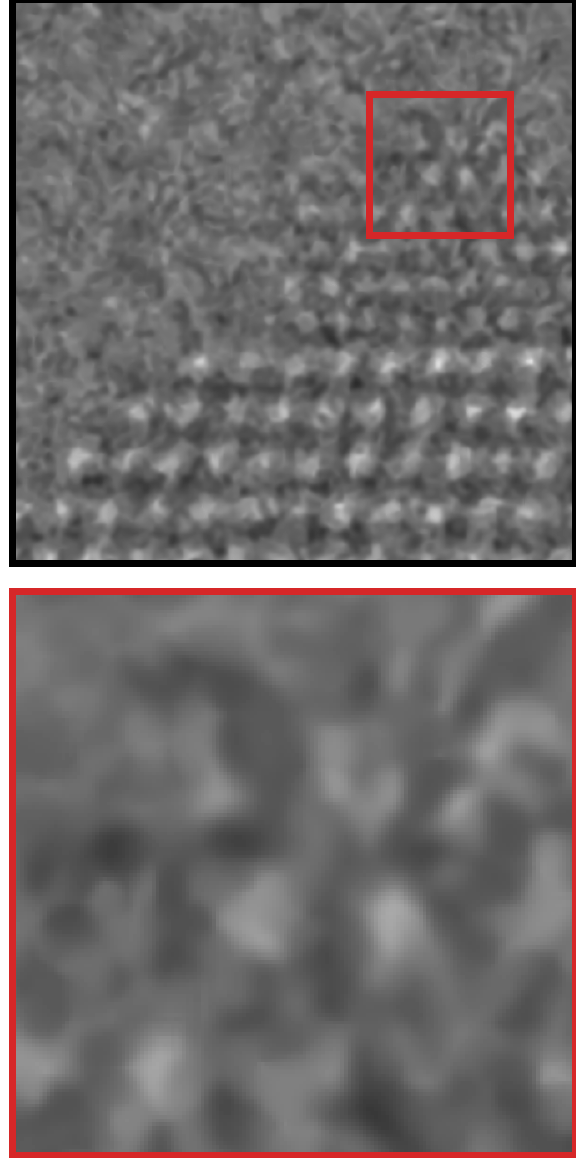}\\

    \footnotesize{PURE-LET} & \footnotesize{SBD+DnCNN} & \footnotesize{SBD+Small UNet} & \footnotesize{Ours} & \footnotesize{Ground Truth} \\
    \includegraphics[width=1.05in]{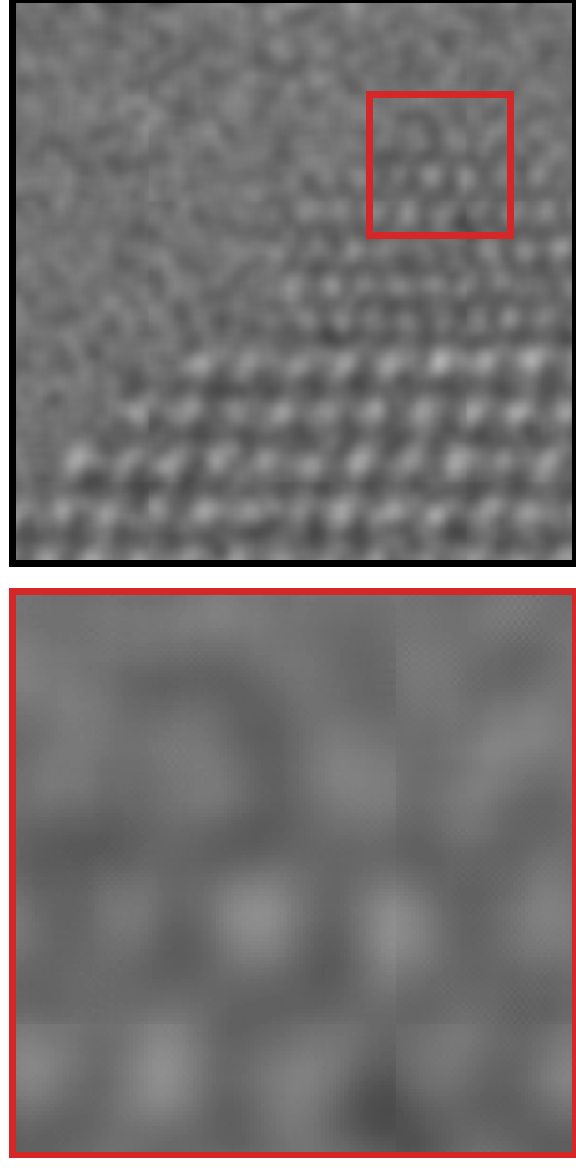}& \includegraphics[width=1.05in]{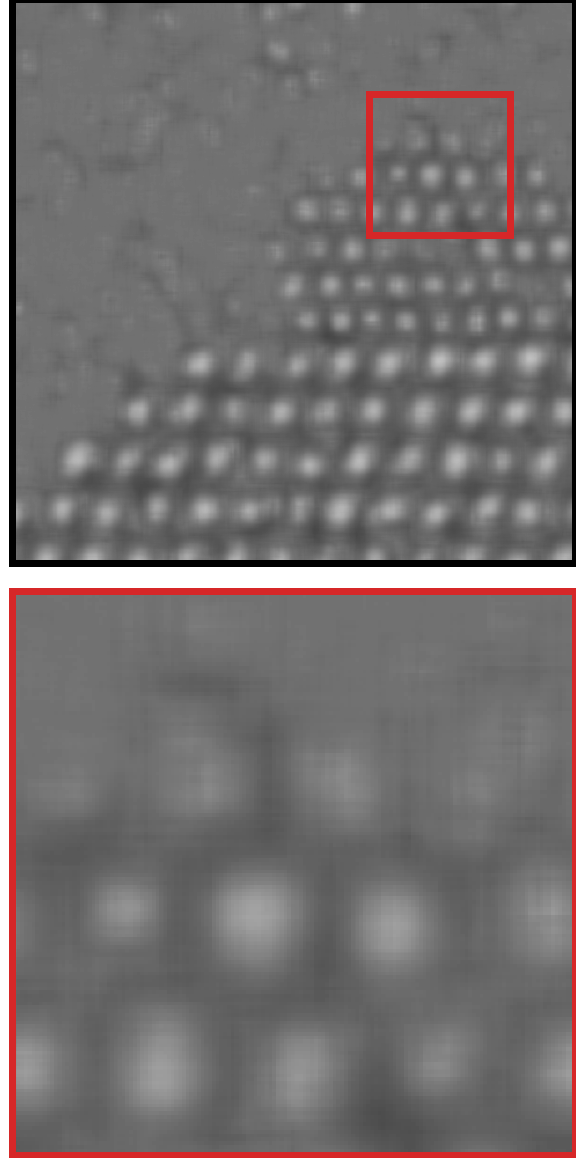}&
    \includegraphics[width=1.05in]{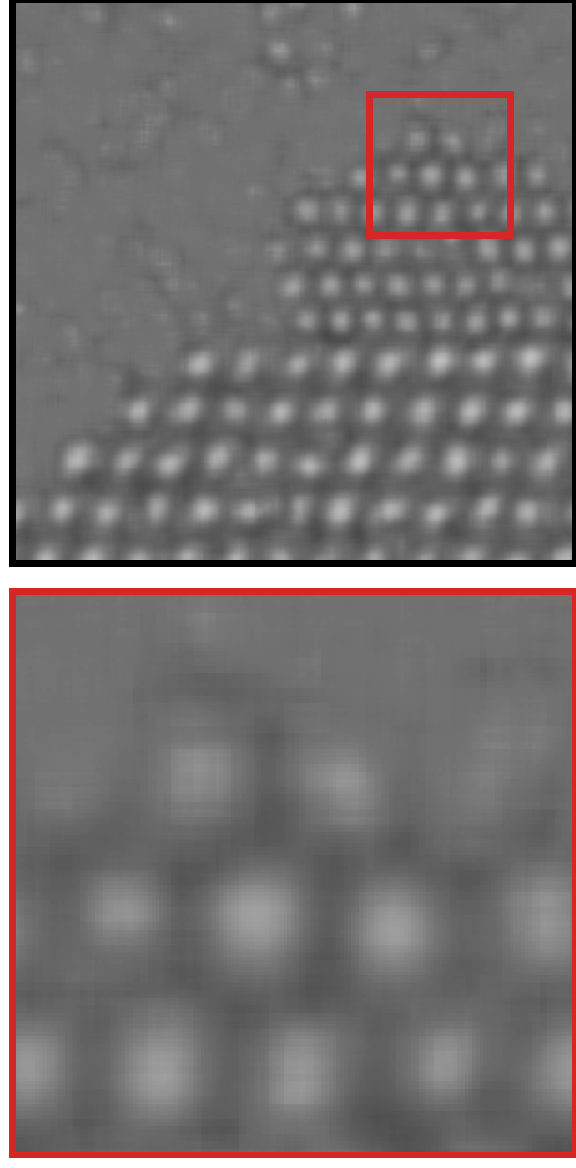}&
    \includegraphics[width=1.05in]{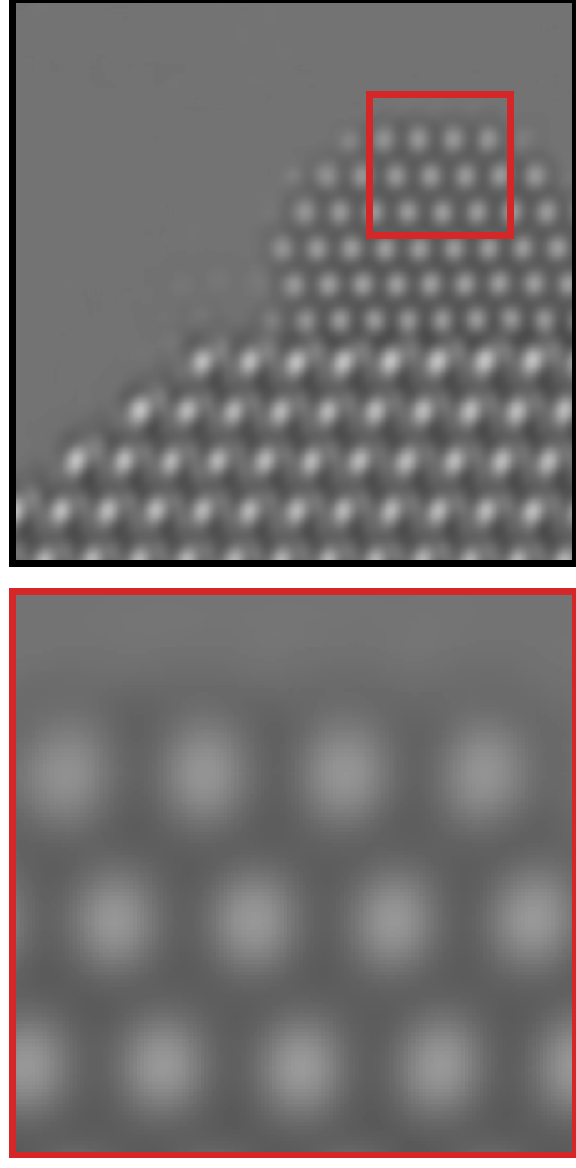}&
    \includegraphics[width=1.05in]{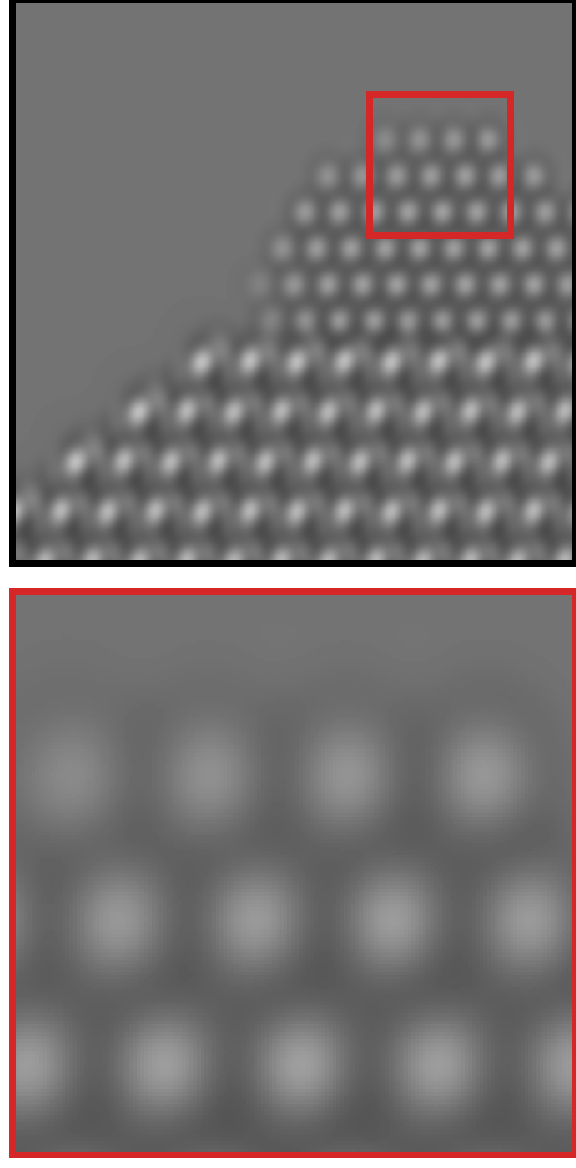}\\
    \end{tabular}
    \caption{\textbf{Denoising results for simulated data}. An additional example comparing SBD and the baseline methods described in Sections~\ref{sec:sbd_comparison} and~\ref{sec:architectures}. The second row zooms in on the region in red box. Our proposed approach produces images of much higher quality than the other approaches, and is able to accurately recover the atomic structure of the nanoparticle. For example, the vacuum region in images denoised by several of the baselines contain visible artefacts, including missing atoms.}
    \label{fig:simulation-denoised-2}
\end{figure*}

\begin{figure}
\centering
\includegraphics[width=1\linewidth]{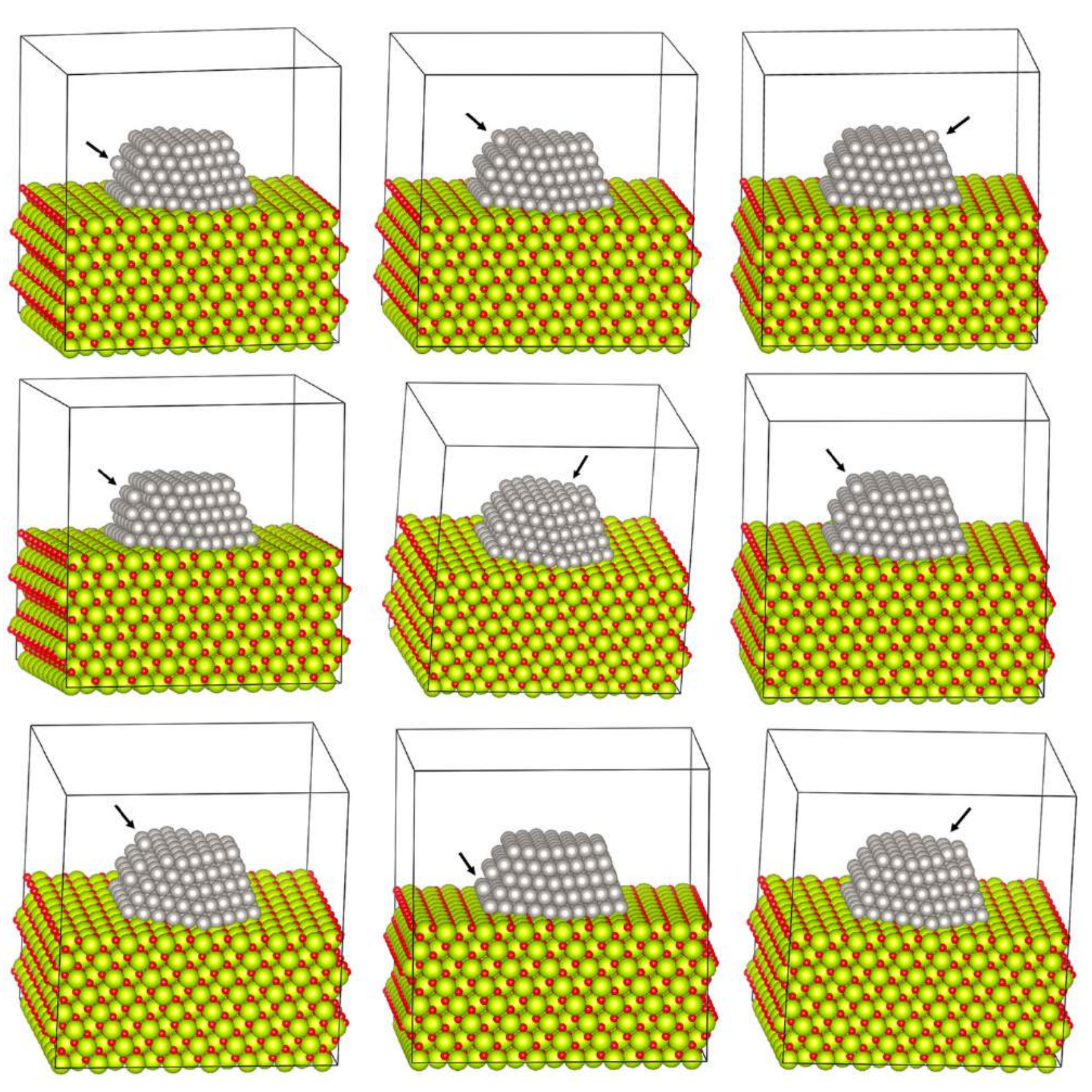}
\caption{\textbf{Example of nanoparticle structures used for surface dataset in Section~\ref{sec:metrics}} A Subset of Pt/CeO$_2$ structual models with atomic-level surface defects like removal of an atom from a column, removal of two atoms, removal of all but one atom and the addition of a new atom at a site. See Section~\ref{sec:metrics} for more details.}
\label{fig:metrics_dataset_fig}
\end{figure}

\begin{figure*}
    \centering
    \begin{tabular}{c@{\hskip 0.01in}c@{\hskip 0.01in}c@{\hskip 0.01in}c@{\hskip 0.01in}c@{\hskip 0.01in}c}
    \footnotesize{Noisy} & \footnotesize{WF} & \footnotesize{Spot Filter} & \footnotesize{VST+NLM} & \footnotesize{VST+BM3D}  &  \\
    \includegraphics[width=1.0in]{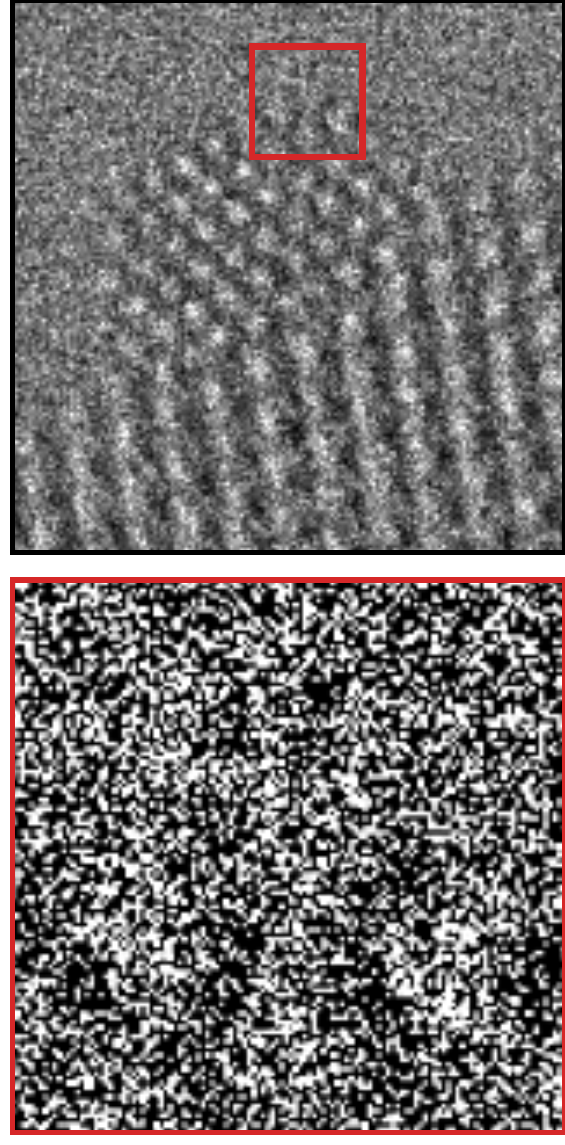}&
    \includegraphics[width=1.0in]{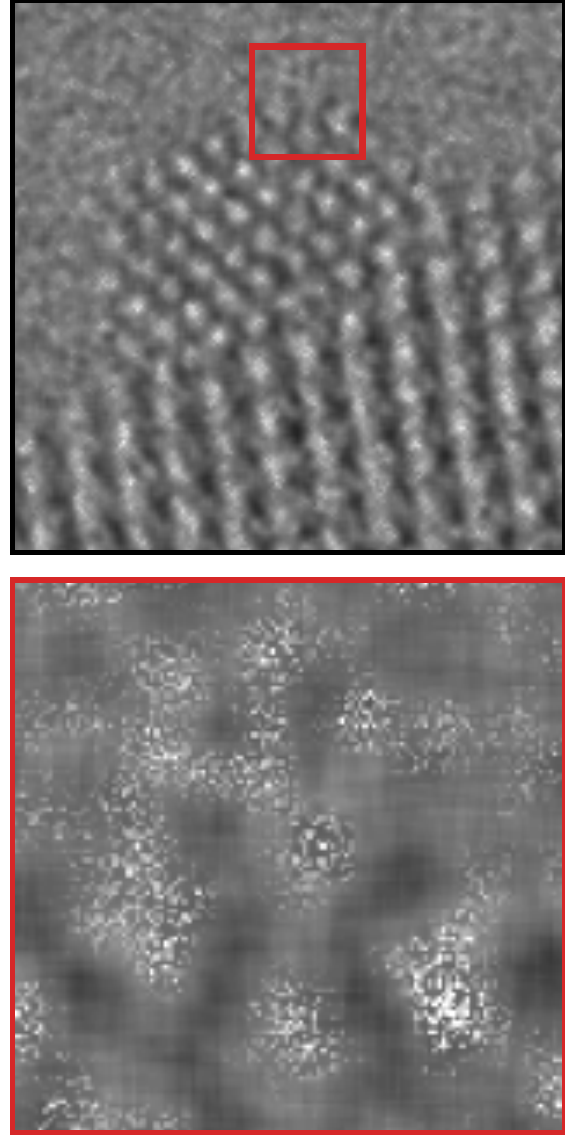}&
    \includegraphics[width=1.0in]{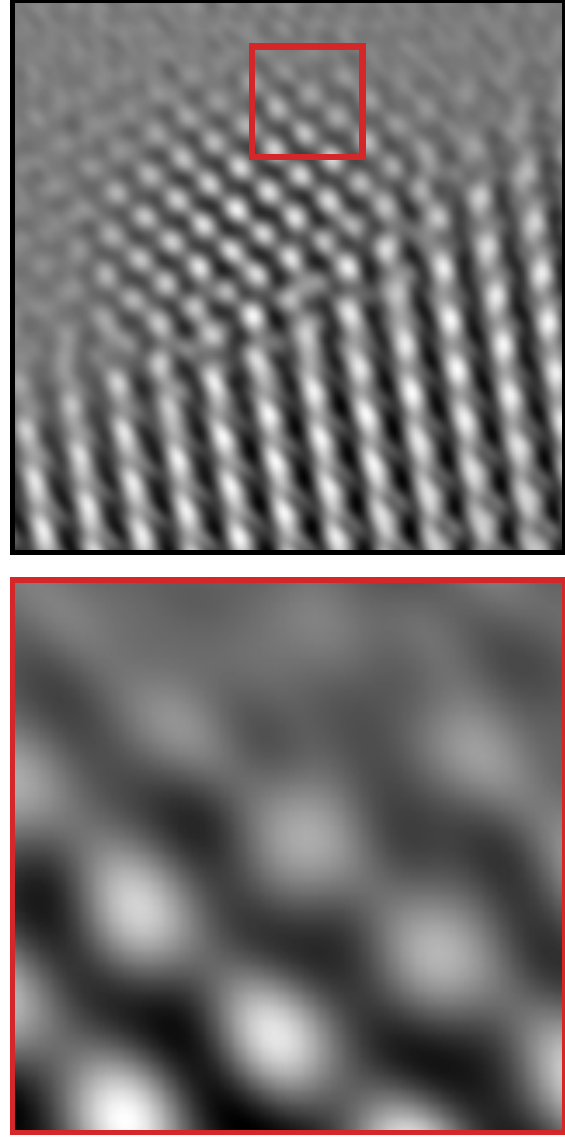}&
    \includegraphics[width=1.0in]{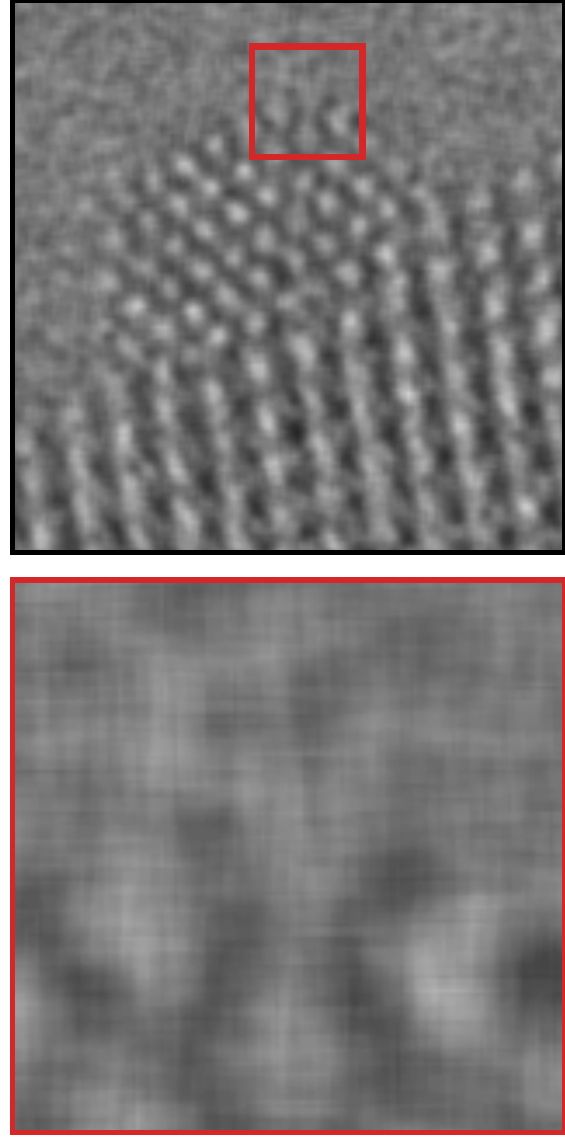}&
    \includegraphics[width=1.0in]{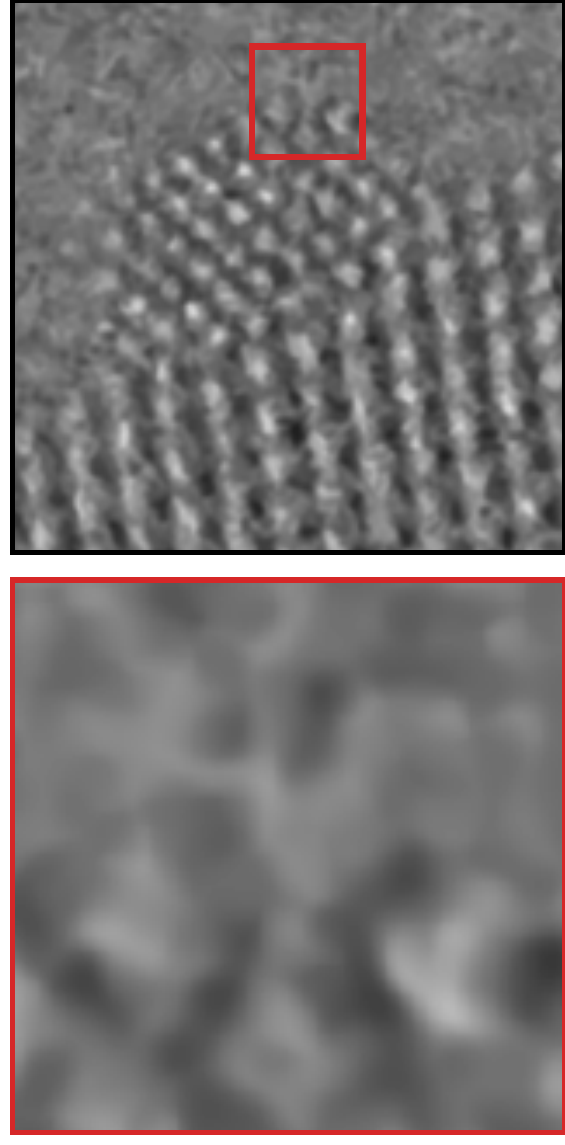}& \\

    \footnotesize{PURE-LET} & \footnotesize{SBD+DnCNN} & \footnotesize{SBD+Small UNet} & \footnotesize{Ours} & \footnotesize{Likelihood Map} &  \\
    \includegraphics[width=1.0in]{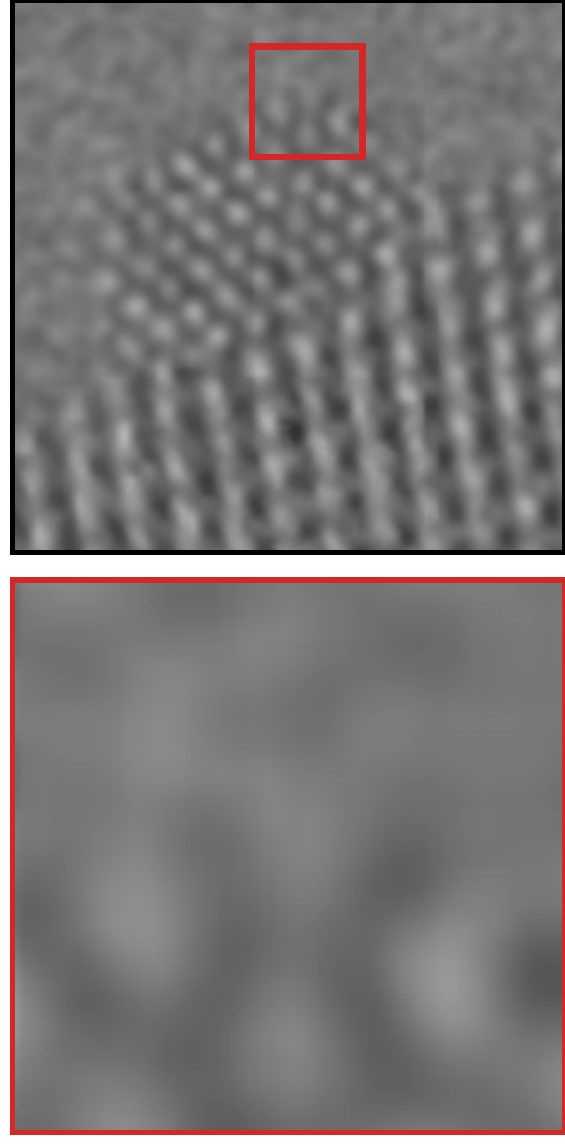}& \includegraphics[width=1.0in]{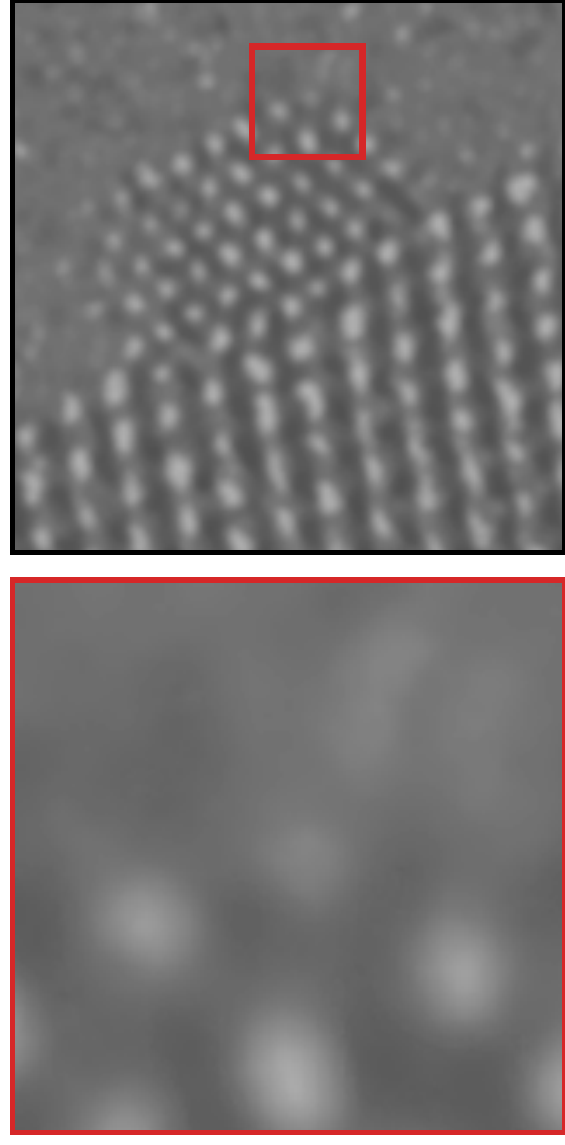}&
    \includegraphics[width=1.0in]{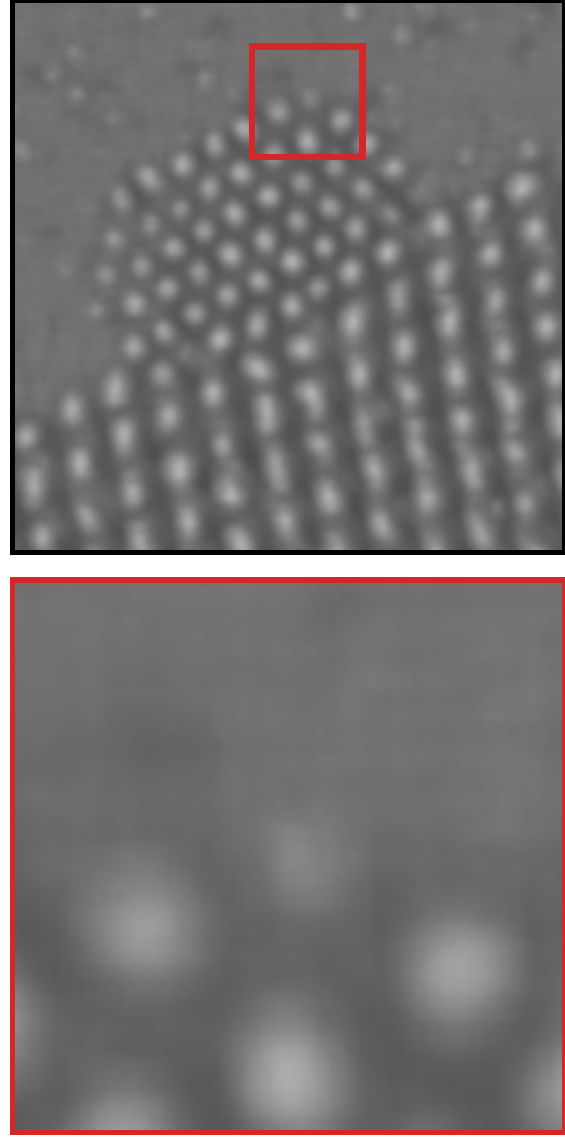}&
    \includegraphics[width=1.0in]{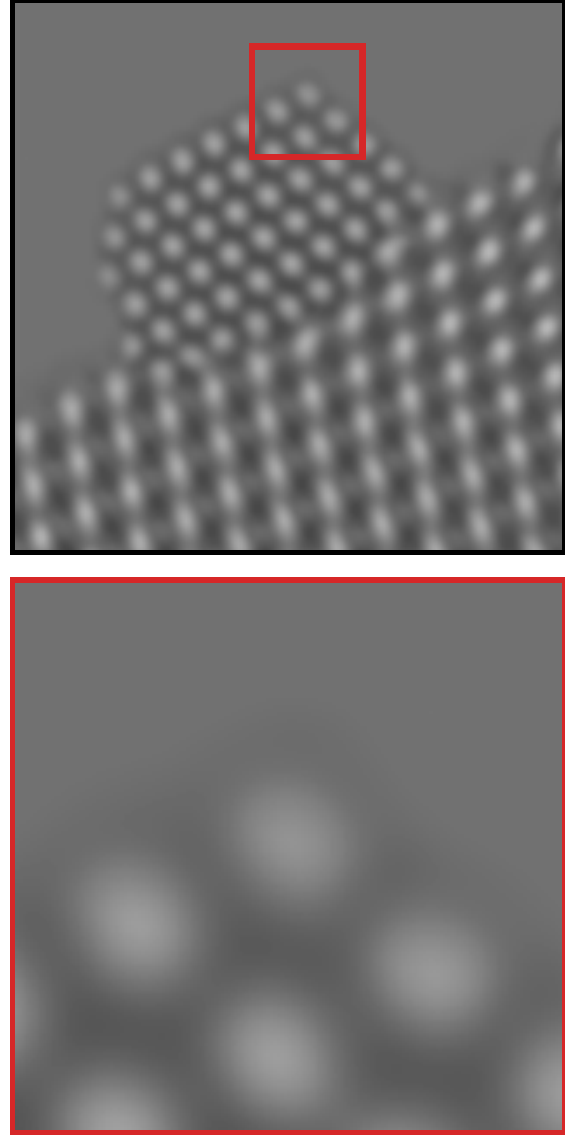}&
    \includegraphics[width=1.0in]{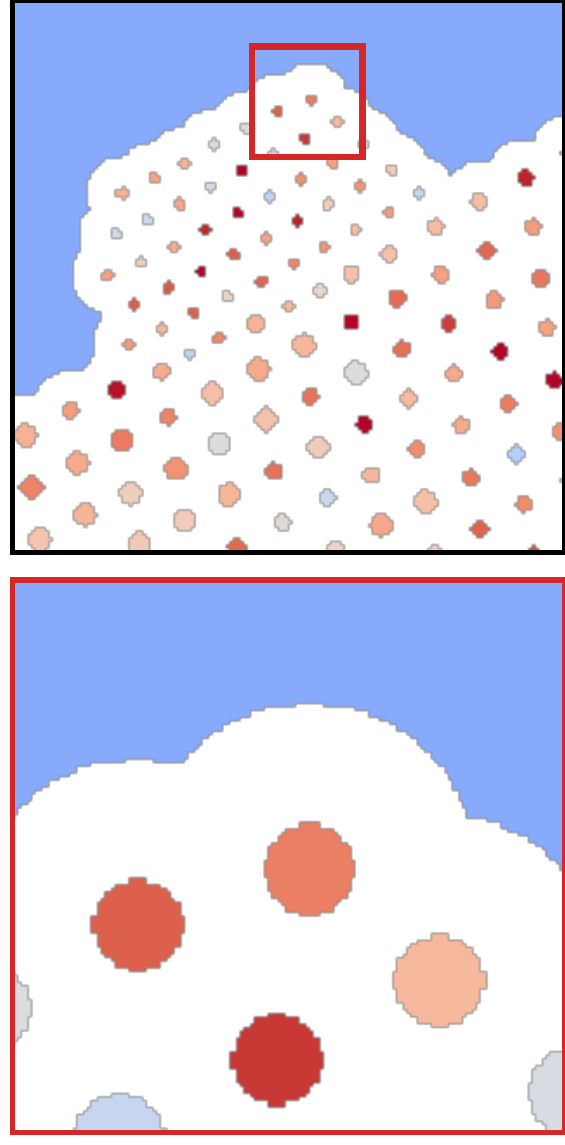}&
    \includegraphics[width=0.335in]{images/appendix_experiment_denoised-U6/colorbar.pdf}\\
    \end{tabular}
    \caption{\textbf{Denoising results for real data}. An additional example comparing SBD and the baseline methods described in Sections~\ref{sec:sbd_comparison} and~\ref{sec:architectures} when applied on the real data described in Section~\ref{sec:real_dataset}. The second row zooms in on the region in red box. In contrast to the other methods, SBD combined with the proposed architecture is able to precisely recover the structure of the nanoparticle and has very few artefacts, particularly in the vacuum region. The likelihood map quantifies the agreement between recovered structures in the denoised images, such as atomic columns and the vacuum, and the observed data (see Section~\ref{sec:likelihood} for more details).}
    \label{fig:real-denoised-appendix2}
\end{figure*}

 In this section we include the following additional results:
 \begin{itemize}[leftmargin=*]
     \item Figure \ref{fig:simulation-denoised-2} show an additional example of simulated image denoised using the proposed approach and the  methods described in Sections~\ref{sec:architectures} and~\ref{sec:sbd_comparison}. 
     \item Figure~\ref{fig:metrics_dataset_fig} visualizes nine nanoparticle structures used for creating the simulated dataset with surface defects described in Section~\ref{sec:metrics}.
     \item Figure~\ref{fig:real-denoised-appendix2} show an additional example of real images denoised using the proposed approach and the  methods described in Sections~\ref{sec:architectures} and~\ref{sec:sbd_comparison}.
     
 \end{itemize}

\end{document}